\documentclass[smallextended]{svjour3}
\usepackage[utf8]{inputenc}
\usepackage{natbib}
\usepackage{hyperref}
\usepackage{bm}
\usepackage{graphicx}
\usepackage{algorithm}
\usepackage[noend]{algpseudocode}
\usepackage{amsmath}
\graphicspath{ {images/} }

\usepackage{listings}
\usepackage{xcolor}
\PassOptionsToPackage{hyphens}{url}\usepackage{hyperref}

\definecolor{codegreen}{rgb}{0,0.6,0}
\definecolor{codegray}{rgb}{0.5,0.5,0.5}
\definecolor{codepurple}{rgb}{0.58,0,0.82}
\definecolor{backcolour}{rgb}{0.97,0.97,0.97}

\lstdefinestyle{python_jay}{
    backgroundcolor=\color{backcolour},
    commentstyle=\color{codegreen},
    keywordstyle=\color{blue},
    numberstyle=\tiny\color{codegray},
    stringstyle=\color{codepurple},
    basicstyle=\ttfamily\footnotesize,
    breakatwhitespace=false,
    breaklines=true,
    captionpos=b,
    keepspaces=true,
    numbers=left,
    numbersep=5pt,
    showspaces=false,
    showstringspaces=false,
    showtabs=false,
    tabsize=2
}

\begin{document}

\title{A Review and Evaluation of Elastic Distance Functions for Time Series Clustering.}
\titlerunning{Elastic Distance Functions for Time Series Clustering}
\author{Christopher Holder \and Matthew Middlehurst \and Anthony Bagnall}
\authorrunning{Holder et al.}

\institute{
Christopher Holder, c.holder@uea.ac.uk, https://orcid.org/0000-0001-9571-3764\\
Matthew Middlehurst, m.middlehurst@uea.ac.uk, https://orcid.org/0000-0002-3293-8779 \\
Anthony Bagnall, ajb@uea.ac.uk, https://orcid.org/0000-0003-2360-8994 \\
School of Computing Sciences, University of East Anglia, Norwich, UK
}

\maketitle

\begin{abstract}
    Time series clustering is the act of grouping time series data without recourse to a label. Algorithms that cluster time series can be classified into two groups: those that employ a time series specific distance measure; and those that derive features from time series. Both approaches usually rely on traditional clustering algorithms such as $k$-means. Our focus is on distance based time series that employ elastic distance measures, i.e. distances that perform some kind of realignment whilst measuring distance. We describe nine commonly used elastic distance measures and compare their performance with $k$-means and $k$-medoids clustering. Our findings are surprising. The most popular technique, dynamic time warping (DTW), performs worse than Euclidean distance with $k$-means, and even when tuned, is no better.
Using $k$-medoids rather than $k$-means improved the clusterings for all nine distance measures.
DTW is not significantly better than Euclidean distance with $k$-medoids. Generally, distance measures that employ editing in conjunction with warping perform better, and one distance measure, the move-split-merge (MSM) method, is the best performing measure of this study. We also compare to clustering with DTW using barycentre averaging (DBA). We find that DBA does improve DTW $k$-means, but that the standard DBA is still worse than using MSM. Our conclusion is to recommend MSM with $k$-medoids as the benchmark algorithm for clustering time series with elastic distance measures. We provide implementations, results and guidance on reproducing results on the associated GitHub repository.
    \keywords{time series clustering; dynamic time warping; barycenter averaging; move-split-merge; edit distance with real penalty; time warp edit distance. }
\end{abstract}

\section{Introduction}
Clustering is an unsupervised analysis technique where a set of cases, defined as a vector of continuous or discrete variables, are grouped to create clusters which contain cases considered to be homogeneous, whereas cases in different clusters are considered heterogeneous~\citep{bonner64clustering}.

Time series clustering is the act of grouping ordered, time series data without recourse to a label. We use the acronym TSCL for time series clustering, to make a distinction between TSCL and time series classification, which is commonly referred to as TSC. There have been a wide range of algorithms proposed for TSCL. Our focus is specifically on clustering using elastic distance measures, i.e. distance measures that use some form of realignment of the series being compared. Our aim to perform a comparative study of these algorithms that  follows the basic structure of bake offs in distance based TSC~\citep{lines15elastic}, univariate TSC~\citep{bagnall17bakeoff} and multivariate TSC~\citep{ruiz21mtsc}. A huge number of alternative transformation based e.g.~\citep{li21linear}, deep learning based clustering algorithms e.g.~\citep{lafabregue22deep} and statistical model based approaches~\citep{caiado16clustering} have been proposed for TSCL. These approaches are not the focus of this research. Our aim is to provide a detailed description of nine commonly used elastic distance measures and conduct an extensive experimentation to compare their utility for TSCL.

Clustering is often the starting point for exploratory data analysis, and is widely performed in all fields of data science~\citep{zolhavarieh14subsequence}. However, clustering is harder to define than, for example, classification. There is debate about what clustering means~\citep{jain88clustering} and no accepted standard definition of what constitutes a good clustering or what it means for cases to be homogeneous or heterogeneous. For example, homogeneous could mean generated by some common underlying process, or mean it has some common hidden variable in common. A clustering of patients based on some medical data might group all male patients in one cluster and all female patients in another. The clusters are from one view homogeneous, but that does not mean it is necessarily a good clustering. The interpretation of the usefulness of a clustering of a particular dataset requires domain knowledge. This makes comparing algorithms over a range of problems more difficult than performing a bake off of TSC algorithms. Nevertheless, there have been numerous comparative studies (for example~\citep{aghabozorgi15time} and~\citep{javed20benchmark}) which take TSC problems, then evaluate clusterings based on how well they recreate class labels. We aim to add to this body of knowledge with a detailed description and a reproducible evaluation of clustering with elastic distance measures (described in Section~\ref{sec:distance}) using the UCR datasets~\citep{dau19ucr}. We do this in the context of partitional clustering algorithms (see Section~\ref{sec:background}). Our first null hypothesis is that elastic distance measures do not perform any better than Euclidean distance with $k$-means clustering. Elastic distance measures are significantly more accurate on average for TSC when used with a one nearest neighbour classifier~\citep{lines15elastic}. We evaluate whether this improvement translates to centroid based clustering. We conclude that, somewhat surprisingly, this is not the case with $k$-means clustering. Only one elastic distance measure, move-split-merge~\citep{stefan13msm}, is significantly better than Euclidean distance, and five of those considered are significantly worse, including the most popular approach, dynamic time warping (DTW). We believe we have detected this difference where others have not because of two factors: firstly, we normalise all series, thus removing the confounding effect of scale; secondly, we evaluate on separate test sets, rather than compare algorithms on train data performance. We explore using dynamic time warping with barycentre averaging (DBA) to improve $k$-means~\citep{petitjean11dba}. DBA finds centroids by aligning cluster members and averaging over values warped to each location. We reproduce the reported improvement that DBA brings to DTW with $k$-means, but we observe that the improvement is not enough to make it better than Euclidean distance, and it comes with a heavy computational overhead. Our first conclusion is that $k$-means clustering with elastic distances is generally not a useful benchmark for TSCL, and if it is used as such, it should employ MSM distance. We then experiment with $k$-medoids TSCL. Using medoids, data points in the training data, rather than averaged cluster members, overcomes the mismatch between simple averaging and elastic distances that barycentre averaging is designed to mitigate against. We found that $k$-medoids improved the clustering of every elastic distance measure, and that again MSM was the top performing algorithm. Our concluding recommendation is that  $k$-medoids with MSM should be the standard benchmark clusterer against which new algorithms should be assessed. We have provided optimised Python implementations for the distances and clusterers in the aeon toolkit\footnote{https://github.com/aeon-toolkit} which provides notebooks on using distances and clusterers and contains all our results with guidance on how to reproduce them.

The remainder of this paper is structured as follows: Section~\ref{sec:background} describes the general TSCL problem and gives a detailed description of nine elastic distance measures that have been proposed in the literature that are used in our experiments.  For more general background on clustering, we direct the reader to~\citep{jain99data,saxena17review}. Section~\ref{sec:evaluation} gives details of the data we use in the analysis, and reviews performance metrics for clustering algorithms. Section~\ref{sec:results} present the results of our experiments and describes our analysis of performance on UCR datasets. Finally, Section~\ref{sec:conclusions} concludes and signposts the future direction of our research.

\section{Time Series Clustering Background}
\label{sec:background}
A time series $\bm{x}$ is a sequence of $m$ observations, $(x_1,\ldots, x_m)$. We assume all series are equal length. For univariate time series, $x_i$ are scalars and for multivariate time series, \textbf{$x_i$} are vectors. A time series data set, $D = \{\bm{x_1}, \bm{x_2}, ..., \bm{x_n}\}$, is a set of $n$ time series cases. A clustering is some form of grouping of cases. Broadly speaking, clustering algorithms are either partitional or hierarchical. Partitional clustering algorithms assign (possibly probabilistic) cluster membership to each time series, usually through an iterative heuristic process of optimising some objective function that measures homogeneity.  Given a dataset of $n$ time series, $D$, the partitional time series clustering problem is to partition $D$ into $k$ clusters, $C = \{C_1, C_2, ..., C_k\}$ where $k$ is the number of clusters.
It is usually assumed that the number of clusters is set prior to the optimisation heuristic. If $k$ is not set, it is commonly found through some wrapper method. We assume $k$ is fixed in advance for all our experiments.

Hierarchical clustering algorithms form a hierarchy of clusters in the form of a dendrogram, a tree where all elements are in a single cluster at the root and in a unique cluster at the leaves of the tree. This presents the analyst with a range of clusterings with different number of clusters. Time series hierarchical clustering is beyond the scope of this study.

As with classification, clustering algorithms can be split into those that work directly with the time series, and those that employ a transformation to derive features prior to clustering. The focus of this study is on non-probabilistic partitional clustering algorithms that work directly with time series.

\subsection{Partitional Time Series Clustering Algorithms}
\label{sec:clustering}

Partitional clustering algorithms have the same basic components that involve example cases (which we call exemplars) that characterise clusters: {\em initialise}  the cluster time exemplars; {\em assign} cases to clusters based on their distance to exemplars.  {\em update} the exemplars based on the revised cluster membership. Iterations of {\em assign} and  {\em update} are repeated until some convergence condition is met.

$k$-means~\citep{macqueen67multivariate}, also known as Lloyd's algorithm~\citep{lloyd82Leastsq}, is the most well known and popular partitional clustering algorithm, in both standard clustering and time series clustering. The algorithm uses $k$ centroids as exemplars for each cluster. A centroid is a summary of the members of a cluster found through the {\em update} operation, which for standard $k$-means involves averaging each time point over cluster members. Each case is assigned to the cluster of its closest centroid, as defined by a distance measure.




Many enhancements of the core algorithm have been proposed. One of the most effective refinements is changing how the exemplars are initialised.  By default, the initial centroids for \textit{k}-means are chosen randomly, either by randomly assigning cluster membership then averaging or by choosing random instances from the training data as initial clusters. However, this risks choosing centroids that are in fact in the same cluster, making it harder to split the clusters. To address this problem, forms of restart and selection are often employed. For example,~\citep{bradley98refining} propose restarting \textit{k}-means over  subsamples of the data and using the resulting centroids to seed the full run. Another solution is to run the algorithm multiple times and keep the model that yields the best result according to some unsupervised performance measure.




$k$-means assumes the number of clusters, $k$ is set a priori. There are a range of methods of finding $k$. These often involve iteratively increasing the number of clusters until some stopping condition is met. This can involve some form of elbow finding or unsupervised quality measure, such as the silhouette value~\citep{lleti04selecting}.  Time series specific enhancements concern the distance metric used and the averaging technique to recalculate centroids. Averaging series matched with an elastic distance measure will mean that, often, the characteristics that made the series similar are lost in the centroid. \citep{petitjean11dba} describe an alternative averaging method based on pairwise DTW. This is described in detail in Section \ref{sec:averaging}.



\textit{k}-medoids is an alternative clustering algorithm which uses cases, or {\em medoids}, as cluster exemplars. One benefit of using instances as cluster centres is that the pairwise distance matrix of the training data is sufficient to fit the model, and this can be calculated before the iterative partitioning. This is particularly important when performing the {\em update} operation, which is the main difference between  $k$-means and $k$-medoids; $k$-medoid chooses a cluster member as the exemplar rather than average. The medoid is the case with the minimum total distance to the other members of the cluster. Refinements such as Partition Around Medoids (PAM)~\citep{kaufman90pam} avoid the complete enumeration of total distances through a swapping heuristic. However, with a precalculated distance matrix and access to reasonable computational resources, this approximation is not necessary.



\section{Time Series Distance Measures}
\label{sec:distance}

Suppose we want to measure the distance between two equal length, univariate time series, $\mathbf{a}=\{a_1,a_2,\ldots,a_m\}$ and $\mathbf{b}=\{b_1,b_2,\ldots,b_m\}$. The Euclidean distance $d_{ed}$ is the L2 norm between series,
    $$d_{ed}(\mathbf{a}, \mathbf{b}) = \sqrt{\sum_{i=1}^m(a_i-b_i)^2}.$$

$d_{ed}$ is a standard starting point for distance based analysis. It puts no priority on the ordering of the series and, when used in TSC, is a poor benchmark for distance based algorithms and a very long way from state of the art~\citep{middlehurst21hc2}. Elastic distance measures allow for possible error in offset by attempting to optimally align two series based on distorting indices or editing series. These have been shown to significantly improve k-nearest neighbour classifiers in comparison to $d_{ed}$~\citep{lines15elastic}. Our aim is to see if we observe the same improvement with clustering.

\subsection{Dynamic Time Warping}

Dynamic Time Warping (DTW)~\citep{ratanamahatana05threemyths} is the most widely researched and used elastic distance measure. It mitigates distortions in the time axis by realigning (warping) the series to best match each other. Let $M(\mathbf{a},\mathbf{b})$ be the $m \times
m$ pointwise distance matrix between series $\mathbf{a}$ and $\mathbf{b}$, where
$M_{i,j}=   (a_i-b_j)^2$. A warping path $$P=<(e_1,f_1),(e_2,f_2),\ldots,(e_s,f_s)>$$ is a set of pairs of indices that define a traversal of matrix $M$. A valid warping path must start at location $(1,1)$, end at point $(m,m)$ and not backtrack, i.e. $0 \leq e_{i+1}-e_{i} \leq 1$ and $0 \leq f_{i+1}- f_i \leq 1$
for all $1< i < m$. The DTW distance between series is the path through $M$ that minimizes the total distance. The distance for any path $P$ of length $s$ is
\[ D_P(\mathbf{a},\mathbf{b}, M) =\sum_{i=1}^s M_{e_i,f_i}.\]
If $\mathcal{P}$ is the space of all possible paths, the DTW path $P^*$ is the path that has the minimum distance, hence the DTW distance between series is
\begin{equation*}
d_{dtw}(\mathbf{a}, \mathbf{b}) =D_{P*}(\mathbf{a},\mathbf{b}, M).
\end{equation*}

\begin{figure}[h]
    \centering
    \includegraphics[width=1.0\textwidth,trim={0cm 0.3cm 1cm 1cm},clip]{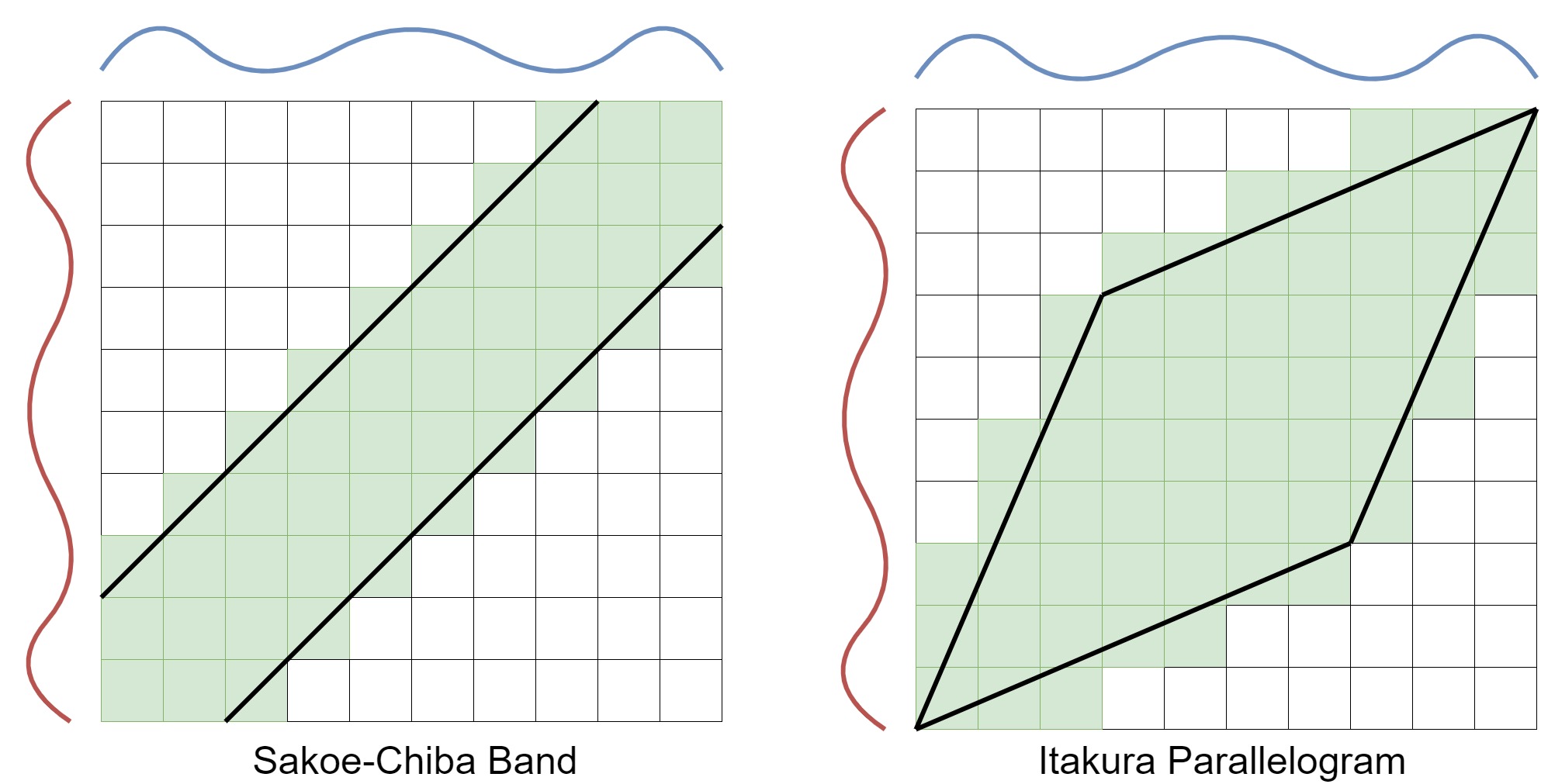}
    \caption{Two most common bounding techniques~\citep{everything2004knowdtw}.}
    \label{fig:bounding-matrix}
\end{figure}

The optimal warping path $P^*$ can be found exactly through a dynamic programming formulation described in Algorithm~\ref{algo:dtw}. This can be a time-consuming operation, and it is common to put a restriction on the amount of warping allowed. Figure~\ref{fig:bounding-matrix} describes the two most frequently used bounding techniques, the Sakoe-Chiba band~\citep{sakoe78band} and Itakura parallelogram~\citep{itakura75band}. In Figure~\ref{fig:bounding-matrix} each individual square represents an element of matrix $M$. Applying a bounding constraint (represented by the darker squares in figure~\ref{fig:bounding-matrix}) reduces the required computation.  The Sakoe-Chiba band creates a warping path window that has the same width along the diagonal of $M$. The Itakura paralleogram allows for less warping at the start or end of the series than in the middle. Algorithm~\ref{algo:dtw} assumes a Sakoe-Chiba band.

\begin{figure}[htb]
    \centering
        \begin{tabular}{c c}
\includegraphics[width=0.5\textwidth]{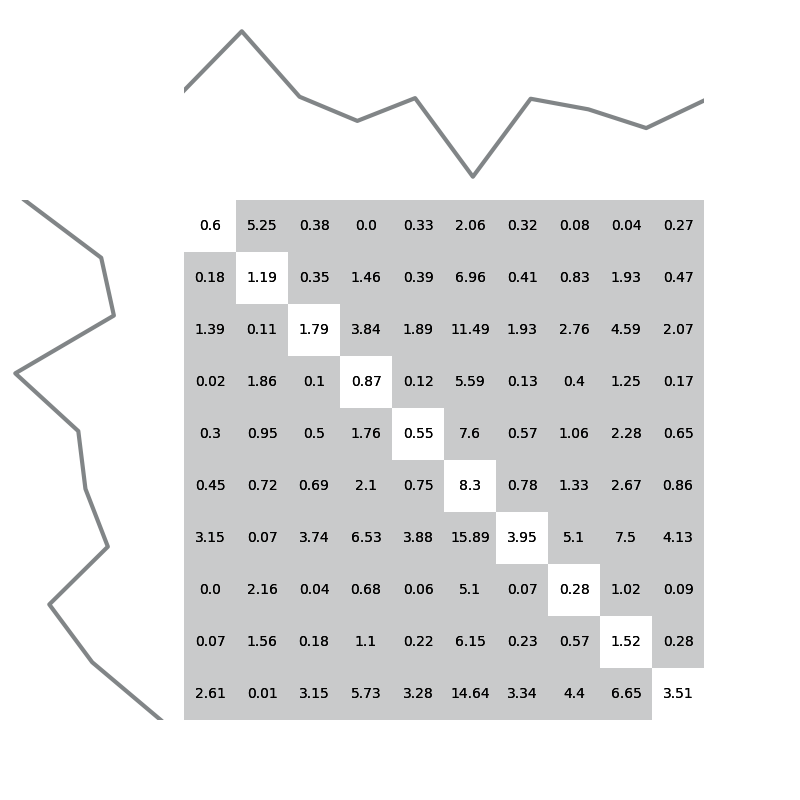} &
    \includegraphics[width=0.5\textwidth]{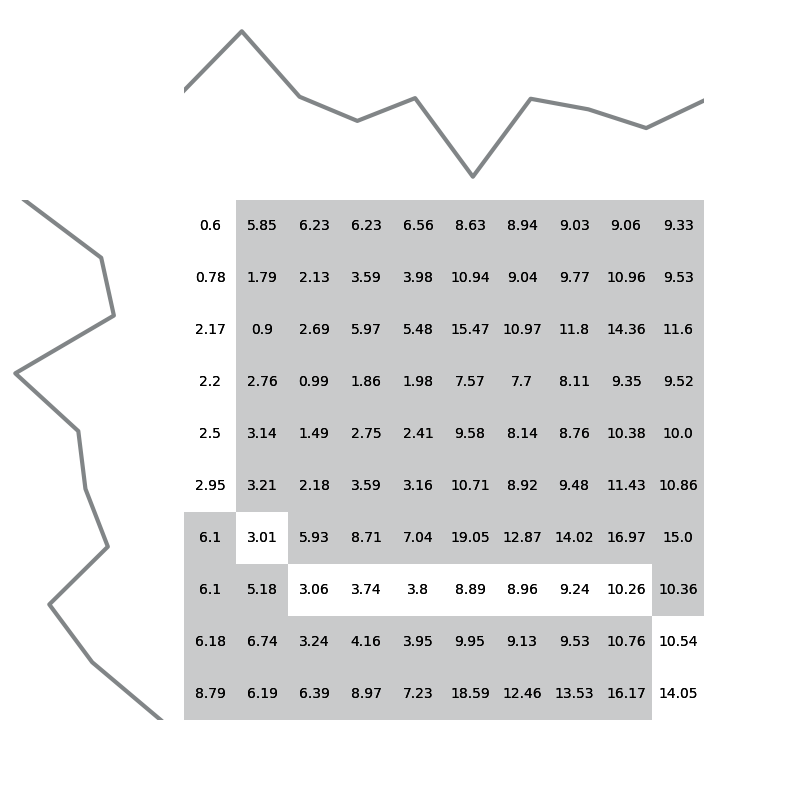} \\
    (a) Pointwise distances matrix $M$& (b) Full window DTW in Matrix $C$
    \end{tabular}
    \caption{An example of Euclidean and DTW distance functions for two series. The left hand matrix, Figure~\ref{fig:example-distances}(a), shows the pointwise distance between the series (matrix $M$ in Equation~\ref{eqn:dtw}). The Euclidean distance is the sum of the diagonal path. The right hand matrix, Figure~\ref{fig:example-distances}(b), shows the DTW distances (matrix $C$ in Equation~\ref{eqn:dtw}) and the resulting warping path when the window size is unconstrained.}
    \label{fig:example-distances}
\end{figure}

\begin{figure}[htb]
    \centering
        \begin{tabular}{c c}
\includegraphics[width=0.5\textwidth]{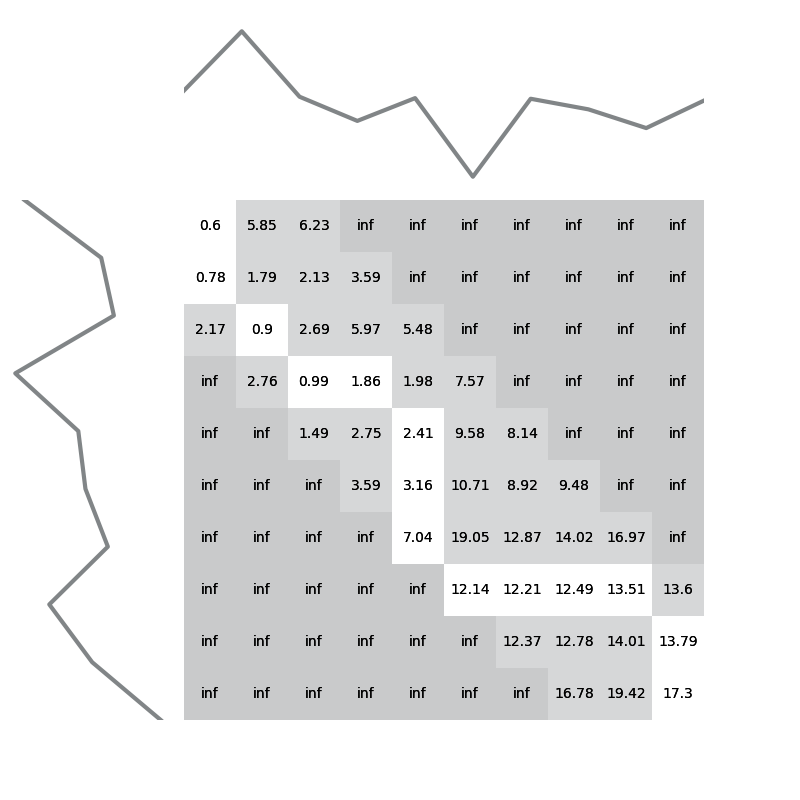} &
    \includegraphics[width=0.5\textwidth]{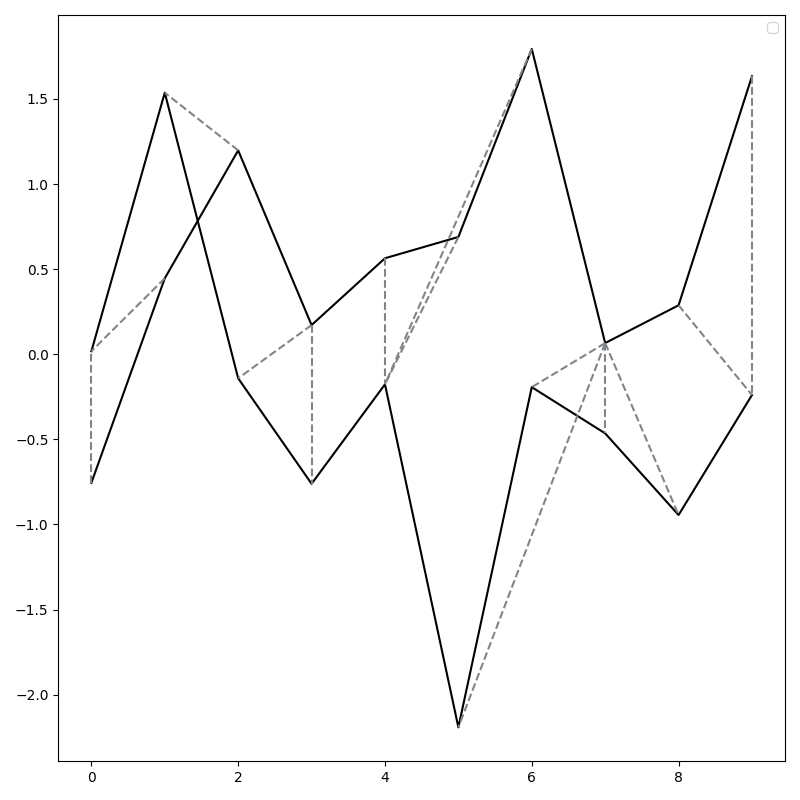} \\
    (a) Constrained DTW with 20\% window & (b) Constrained DTW alignment.
    \end{tabular}
    \caption{An example of constraining the DTW window. The left hand matrix, Figure~\ref{fig:example-window}(a), show the DTW distances for the same series used in Figure~\ref{fig:example-distances} when the window is contrained to 20\% of the series length using a Sakoe-Chiba band. Figure~\ref{fig:example-window}(a) shows the resulting alignment.}
    \label{fig:example-window}
\end{figure}

\begin{algorithm}[htbp]
	\caption{DTW (${\bf a},{\bf b}$, (\textit{both series of length $m$}), $w$ (\textit{window proportion}, default value $ w \leftarrow 1$), $M$ (\textit{pointwise distance matrix}))}
\label{algo:dtw}
	\begin{algorithmic}[1]
\State Let $C$ be an $(m+1)\times(m+1)$ matrix initialised to zero, indexed from zero.
\For{$i \leftarrow  1$ to $m$}
    \For{$j \leftarrow  1$ to $m$}
        \If{$|i-j| < w\cdot m$}
            \State $C_{i,j} \leftarrow M_{i,j} + \min( C_{i-1,j-1}, C_{i-1, j}, C_{i,j-1})$
        \EndIf
   \EndFor
\EndFor
\Return $C_{m,m}$
	\end{algorithmic}
\end{algorithm}
The DTW distance with Sakoe-Chiba window $w$ can be expressed as Equation~\ref{eqn:dtw}.
\begin{equation}
d_{dtw}(\mathbf{a}, \mathbf{b}) =D_{P*}(\mathbf{a},\mathbf{b}, M) = DTW(\mathbf{a}, \mathbf{b}, w, M).
\label{eqn:dtw}
\end{equation}
More general bands can be imposed in an implementation by setting values outside the band in the matrix, $M$, to infinity. Figure~\ref{fig:example-distances} helps explain how the DTW calculations are arrived at. Euclidean distance is simply the sum of the diagonals of the Matrix $M$, in Figure~\ref{fig:example-distances}(a). DTW constructs $C$ using $M$ and previously found values. For example, $C_{1,1} = M_{1,1} = 0.6$ and $C_{1,2}$ is the minimum of $C_{1,1}$, $C_{0,1}$ and $C_{0,2}$ plus $M_{1,2}$. $C_{0,2}$ and $C_{0,1}$ are initialised to infinity, so the best path to get to  $C_{1,2}$ has distance  $C_{1,1}+ M_{1,2}$ which equals 0.6+5.25 = 5.85. Similarly, cell $C_{10,10}$ is the minimum of the three cells $C_{9,9}, C_{10,9}, C_{9,10}$ plus the pointwise distance $M_{10,10}$. The optimal path is the trace back through the minimum values (shown in white in Figure~\ref{fig:example-distances}(b)).
Figure~\ref{fig:example-window} gives a demonstration of the effect of constraining the warping path on DTW using the same two series from Figure~\ref{fig:example-distances}. The relatively extreme warping from point 0 to point 5 evident in Figure~\ref{fig:example-distances}(a) is constrained when the maximum warping allowed is 2 places ($w=0.2$) in Figure~\ref{fig:example-window}.

\subsection{Derivative Dynamic Time Warping}

\cite{keogh01derivative} proposed a modification of DTW called Derivative Dynamic Time Warping (DDTW) that first transforms the series into a differential series. The difference series of $\mathbf{a}$, $\mathbf{a'}=\{a'_2,a'_3,\ldots,a'_{m-1}\}$ where $a'_i$ is defined as the average of the slopes between $a_{i-1}$ and $a_i$ and $a_i$ and $a_{i+1}$, i.e.

$$a'_i = \frac{(a_i-a_{i-1})+(a_{i+1}-a_{i-1})/2}{2},$$
for $1<i<m$. The DDTW is then simply the DTW of the differenced series,
\begin{equation}
d_{ddtw}(\mathbf{a},\mathbf{b})= d_{dtw}(\mathbf{a'},\mathbf{b'}).
\label{eqn:ddtw}
\end{equation}

\subsection{Weighted Dynamic Time Warping}

\begin{figure}[htb]
    \centering
        \begin{tabular}{c c}
\includegraphics[width=0.5\textwidth]{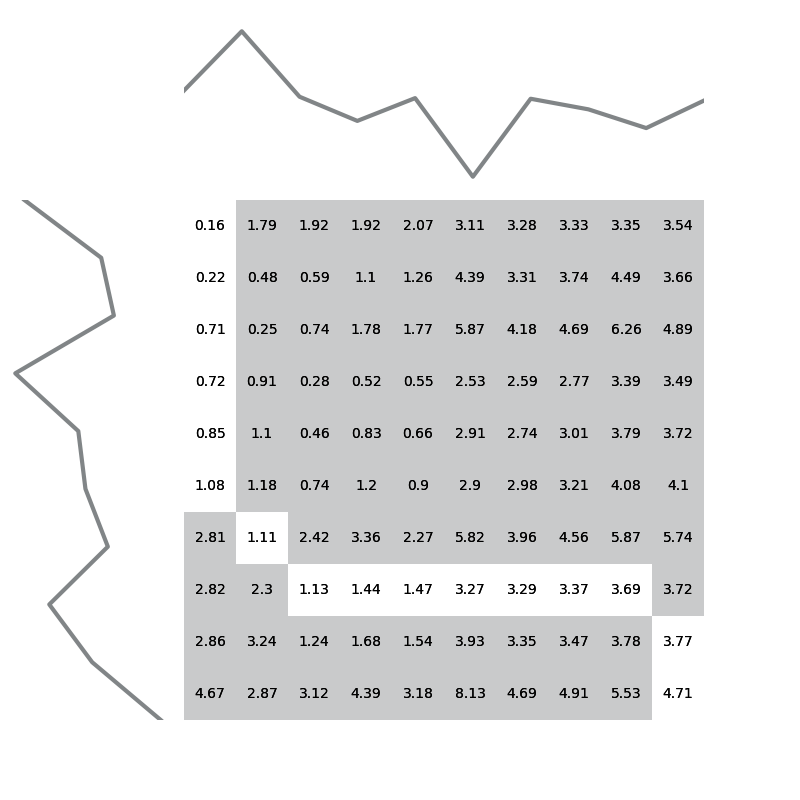} &
    \includegraphics[width=0.5\textwidth]{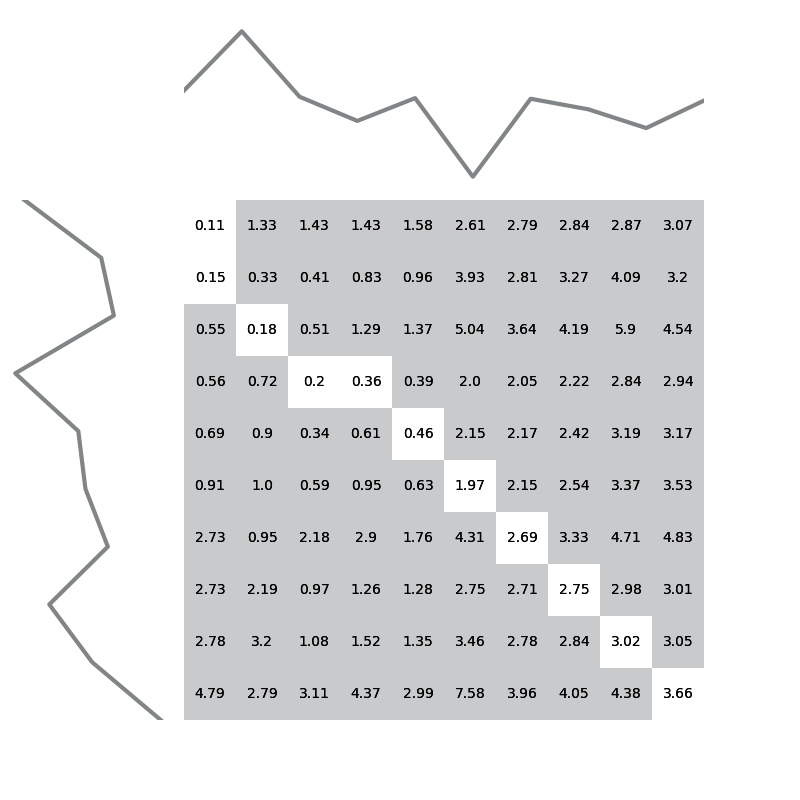} \\
    (a) Weighted DTW with weight g = 0.2 & (b) Weighted DTW with a weight g = 0.3.
    \end{tabular}
    \caption{Examples of the weighted DTW cost matrix $C$ and resulting alignment for two weight parameters.}
    \label{fig:example-wdtw}
\end{figure}

A weighted form of DTW (WDTW) was proposed by~\cite{jeong11weighted}. WDTW adds a multiplicative weight penalty based on the warping distance between points in the warping path. It is a smooth alternative to the cutoff point approach of using a warping window. When creating the distance matrix $M$, a weight penalty  $w(|i-j|)$ for a warping distance of  $|i-j|$ is applied, so that
$$M^w_{i,j}=  w(|i-j|) \cdot (a_i-b_j)^2.$$
A logistic weight function is proposed in~\cite{jeong11weighted}, so that a warping of $a$ places imposes a weighting of
$$w(a)=\frac{w_{max}}{1+e^{-g\cdot(a-m/2)}},$$
where $w_{max}$ is an upper bound on the weight (set to 1), $m$ is the series length and $g$ is a parameter that controls the penalty level for large warpings. The larger $g$ is, the greater the penalty for warping. Note that WDTW does not benefit from the reduced search space a window induces. The WDTW distance is then
\begin{equation} d_{wdtw}(\mathbf{a}, \mathbf{b}) =D_{P*}(\mathbf{a},\mathbf{b}, M^w) = DTW(\mathbf{a},\mathbf{b}, M^w).
\label{eqn:wdtw}
\end{equation}
Figure~\ref{fig:example-wdtw} shows the warping resulting from two parameter values. For this example, $g =0.2$ gives the same warping as full window DTW, but $g=0.3$ is more constrained.

We also investigate the derivative weighted distance (WDDTW),
\begin{equation} d_{wddtw}(\mathbf{a}, \mathbf{b}) = d_{wdtw}(\mathbf{a'}, \mathbf{b'}).
\label{eqn:dwdtw}
\end{equation}

\subsection{Longest Common Subsequence}

DTW is usually expressed as a mechanism of finding an alignment, so that points are warped onto each other to form a path. An alternative way of looking at the process is as a mechanism of forming a common series between the two input series. With this view, the warping operation can be seen as inserting a gap in one series, or removing an element from another series. This way of thinking derives from approaches for aligning sequences of discrete variables, such as strings or DNA. The Longest Common Subsequence (LCSS) distance for time series is derived from a solution to the problem of finding the longest common subsequence between two discrete series through edit operations. For example, if two discrete series are $abaacb$ and $bcacab$, the LCSS is $baab$. Unlike DTW, The LCSS does not provide a path from $(1,1)$ to $(m,m)$. Instead, it describes edit operations to form a final sequence, and these operations are given a certain cost. So, for example, to edit $abaacb$ into the LCSS $baab$ requires two deletion operations. For DTW, you can then think of the choice between the three warping paths in line 5 of Equation~\ref{eqn:dtw} as $C_{i-1,j}$ being a deletion in series $\mathbf{b}$, $C_{i, j-1}$ as a deletion in series $\mathbf{a}$ and $C_{i-1, j-1}$ as a match. The warping path shown in Figure~\ref{fig:example-window} is a sequence of pairs

$$<(1,1), (2,1), (3,2), (4,3), (4,4), (5,5), (6,5),$$
$$(7,5), (8,6), (8,7), (8,8), (8,9), (9,10), (10,10)>$$ can instead be expressed as an edited series $$<(1,1), (3,2), (4,3), (5,5), (8,6), (9,10)>.$$ With this representation the warping operations are in fact deletions (or gaps) in series $\mathbf{a}$ in positions 2, 6, 7 and 10 and in $\mathbf{b}$ in positions 4, 7, 8 and 9. With discrete series, the matches in the common subsequence have the same value. Thus, each pair in the subsequence would be, for example, a letter in common for the two series. Obviously, actual matches in real valued series will be rare. The discrete LCSS algorithm can be extended to consider real-valued time series by using a distance threshold $\epsilon$, which defines the maximum difference between a pair of values that is allowed for them to be considered a match. The length of the LCSS between two series ${\bf a}$ and ${\bf b}$ can be found using Algorithm~\ref{algo:lcss}. If two cells are considered the same (line 4), the previously considered LCSS is increased by one. If not, then the LCSS seen so far is carried forward.
\begin{algorithm}[ht]
	\caption{LCSS (${\bf a},{\bf b}$ , (\textit{both series of length $m$}),  $\epsilon$ (\textit{equality threshold}))}
\label{algo:lcss}
	\begin{algorithmic}[1]
\State Let $L$ be a $(m+1)\times(m+1)$ matrix initialised to zero, indexed from zero.
\For{$i \leftarrow  1$ to $m$}
    \For{$j \leftarrow  1$ to $m$}
            \If{$|a_i - b_j| < \epsilon$}
                \State $L_{i,j} \leftarrow L_{i-1,j-1}+1$
            \Else
                \State $L_{i,j} \leftarrow \max(L_{i-1,j}, L_{i,j-1})$
            \EndIf
   \EndFor
\EndFor
\Return $L_{m,m}$
	\end{algorithmic}
\end{algorithm}
The LCSS distance between ${\bf a}$ and ${\bf b}$ is
\begin{equation}
d_{LCSS}({\bf a,b}) = 1- \frac{LCSS({\bf a,b})}{m}.
\label{eqn:lcss}
\end{equation}

\begin{figure}[htb]
    \centering
    \includegraphics[width=1\textwidth]{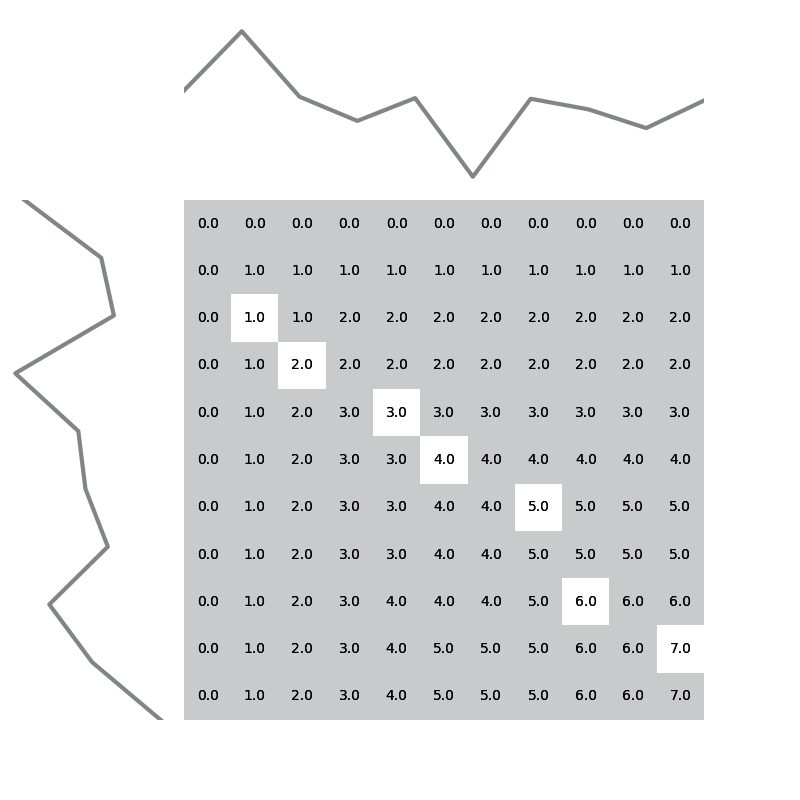}
    \caption{An example of the LCSS cost matrix with $\epsilon = 1$, where the white cells are members of the LCSS.}
    \label{fig:example-lcss}
\end{figure}

Figure~\ref{fig:example-lcss} shows the cost match between our two example series. The longest sequence of matches is seven. These are the pairs $$<(2,1), (3,2), (4,4), (5,5), (6,8), (8,8), (9,10)>$$.


\subsection{Edit Distance on Real Sequences (EDR)}
\begin{algorithm}[htbp]
	\caption{EDR (${\bf a},{\bf b}$ , (\textit{both series of length $m$}), $\epsilon$ (\textit{equality threshold}) }
\label{algo:edr}
	\begin{algorithmic}[1]
\State Let $E$ be an $(m+1)\times(m+1)$ matrix initialised to zero, indexed from zero.
\For{$i \leftarrow  1$ to $m$}
    \For{$j \leftarrow  1$ to $m$}
            \If{$i = 0 \lor j = 0$}
                \State $E_{i,j} \leftarrow m$
                \If{$|a_i - b_j| < \epsilon$}
                    \State $c \leftarrow 0$
                \Else
                    \State  $c \leftarrow 1$
                \EndIf
                \State $match \leftarrow E_{i-1,j-1}+ c$
                \State $insert \leftarrow E_{i-1,j}+ 1$
                \State $delete \leftarrow  E_{i,j-1}+ 1$
                \State $E_{i,j} \leftarrow \min(match, insert, delete)$
            \EndIf
   \EndFor
\EndFor
\Return $E_{m,m}$
	\end{algorithmic}
\end{algorithm}
\begin{figure}[htb]
    \centering
    \includegraphics[width=1\textwidth]{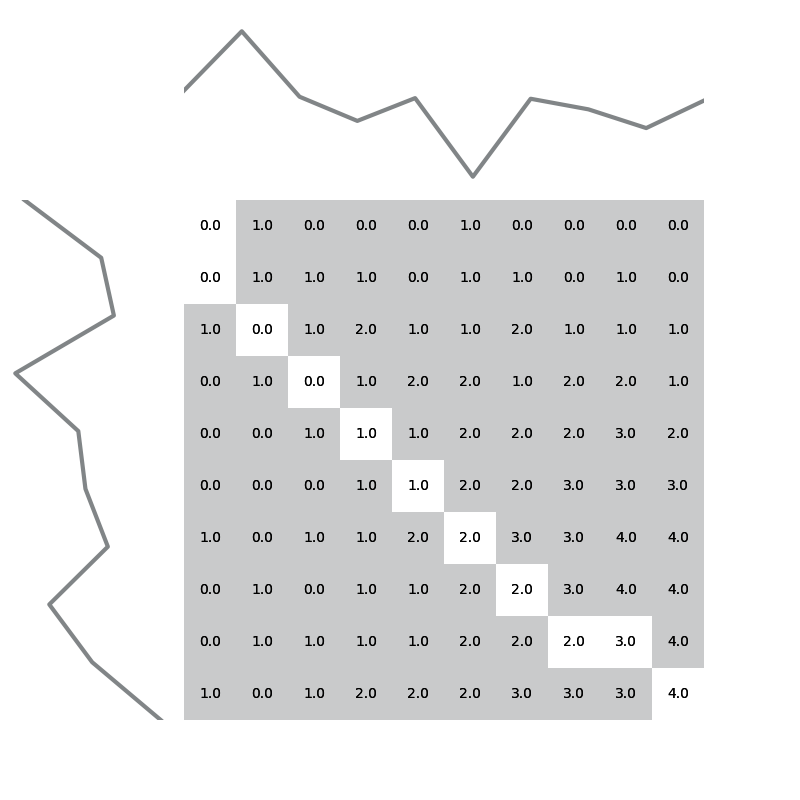}
    \caption{An example of the EDR cost matrix with $\epsilon = 1$, where the white cells are members of the least cost alignment path.}
    \label{fig:example-edr}
\end{figure}

LCSS was adapted in~\cite{chen05erd}, where the edit distance on real sequences (EDR) is described. Like LCSS, EDR uses a distance threshold to define when two elements of a series match. However, rather than count matches and look for the longest sequence, ERP applies a (constant) penalty for non-matching elements where deletions, or gaps, are created in the alignment. Given a distance threshold, $\epsilon$, the EDR distance between two points in series ${\bf a}$ and ${\bf b}$ is given by Algorithm~\ref{algo:edr}. The EDR distance between ${\bf a}$ and ${\bf b}$ is then defined by Equation~\ref{eqn:edr}.
\begin{equation}
d_{EDR}({\bf a,b}) = EDR({\bf a, b}, w, \epsilon)
\label{eqn:edr}
\end{equation}
At any step, elastic distances can use one of three costs: diagonal, horizontal or vertical, in forming an alignment. In terms of forming a subsequence from series $\mathbf{a}$, we can characterise these as operations as match (diagonal), deletion (horizontal) and insertion (vertical). Insertion to $\mathbf{a}$ can also be characterised as deletion from $\mathbf{b}$, but we retain the match/delete/insert terminology for consistence and clarity.
EDR does not satisfy triangular inequality, as equality is relaxed by assuming elements are equal when the distance between them is less than or equal to $\epsilon$. EDR is very similar to LCSS, but it directly finds a distance rather than simply counting matches, thus removing the need for the subtraction in Equation~\ref{eqn:lcss}. The resulting cost matrix shown in Figure~\ref{fig:example-edr} can easily be used to find either an alignment path or a common subsequence.
EDR then characterises the three operations in DTW

\subsection{Edit Distance with Real Penalty (ERP)}
\begin{algorithm}[htbp]
	\caption{ERP (${\bf a},{\bf b}$ (\textit{both series of length $m$}),  $g$, (\textit{penalty value}), $d$, (\textit{pointwise distance function}))}
\label{algo:erp}
	\begin{algorithmic}[1]
\State Let $E$ be an $(m+1)\times(m+1)$ matrix initialised to zero, indexed from zero.
\For{$i \leftarrow  1$ to $m$}
    \For{$j \leftarrow  1$ to $m$}
            \If{i = 0}
                \State $E_{i,j} \leftarrow \sum_{k=1}^m d(b_k,g)$
            \ElsIf{j = 0}
                \State $E_{i,j}\leftarrow \sum_{k=1}^m d(a_k,g)$
            \Else
                \State $match \leftarrow E_{i-1,j-1}+ d(a_i, b_j)$
                \State $insert \leftarrow E_{i-1,j}+ d(a_i, g)$
                \State $delete \leftarrow  E_{i,j-1}+ d(g, b_j)$
                \State $E_{i,j} \leftarrow \min(match, insert, delete)$
            \EndIf
   \EndFor
\EndFor
\Return $E_{m,m}$
	\end{algorithmic}
\end{algorithm}
\begin{figure}[htb]
    \centering
\includegraphics[width=1\textwidth]{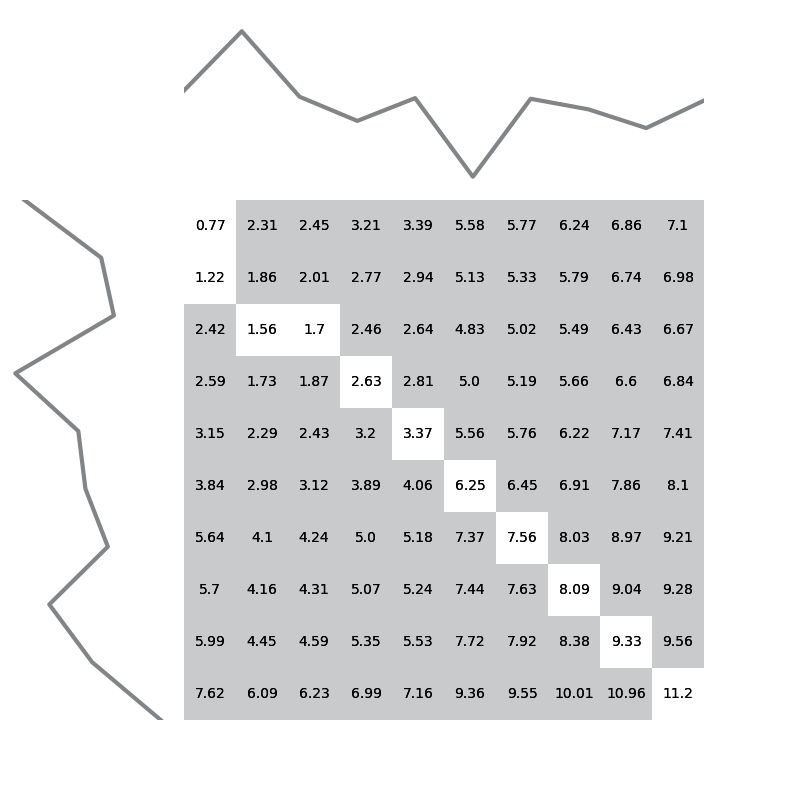}]
    \caption{An example of ERP distance calculation using our example series, with $g=0$.}
    \label{fig:example-erp}
\end{figure}

An alternative to EDR was proposed in~\cite{chen04erp}, where edit distance with real penalty (ERP) was introduced. LCSS produces a subsequence that can have gaps between elements. For example, there is a gap between  (3,2) and (4,4) in the subsequence shown in Figure~\ref{fig:example-lcss}. ERP imposes a penalty for when gaps occur based on distance to a constant parameter $g$. ERP uses $d(a,b) = \sqrt((a-b)^2)$ for the pointwise distance rather than $d(a,b) = (a-b)^2$ used to find $M$ for DTW. ERP calculates a cost matrix $E$ that is more like DTW than LCSS: it describes a path alignment of two series based on edits rather than warping. The ERP distance between two series is described in Algorithm~\ref{algo:erp}. The edge terms are initialised to a large constant value (lines 5 and 7).
The cost matrix $E$ is then either the cost of matching, $E_{i-1,j-1}+ d(a_i, b_j)$, where $d(a_i, b_j)$ is the distance between two points or the cost of a inserting/deleting a term on either axis ($E_{i-1,j}+ d(a_i, g)$ or  $E_{i,j-1}+ d(g, b_j)$). The ERP distance between ${\bf a}$ and ${\bf b}$ is given by Equation~\ref{eqn:erp}.
\begin{equation}
d_{ERP}({\bf a,b}) = ERP({\bf a, b}, g,d)
\label{eqn:erp}
\end{equation}
 ERP satisfies triangular inequality and is a metric. Figure~\ref{fig:example-erp} shows the cost matrix and resulting alignment for our example series with $g=0$. $E_{1,1}$ is 0.774, which is simply the square root of $M_{1,1}$. $E_{2,1}$ is the minimum of the two edge cases to the left (large constants) and $E_{1,1}+d(b_1,g)$, where $b_1=0.42$, so $d(0.17,1)=0.97$. Hence, $E_{2,1}= 0.774+0.424 = 1.2.$

\subsection{Move-Split-Merge (MSM)}

\begin{equation}
C(x,y,z) = \left\{ \begin{array}{l}
 	\mbox{$c$ \textbf{if} $y \leq x \leq z $ \textbf{or} $y \geq x \geq z$} \\
  	\mbox{$c+min(|x-y|,|x-z|)$ \textbf{otherwise}.}
       \end{array} \right.
\label{eqn:cost}
\end{equation}
\begin{figure}[htb]
    \centering
\includegraphics[width=1\textwidth]{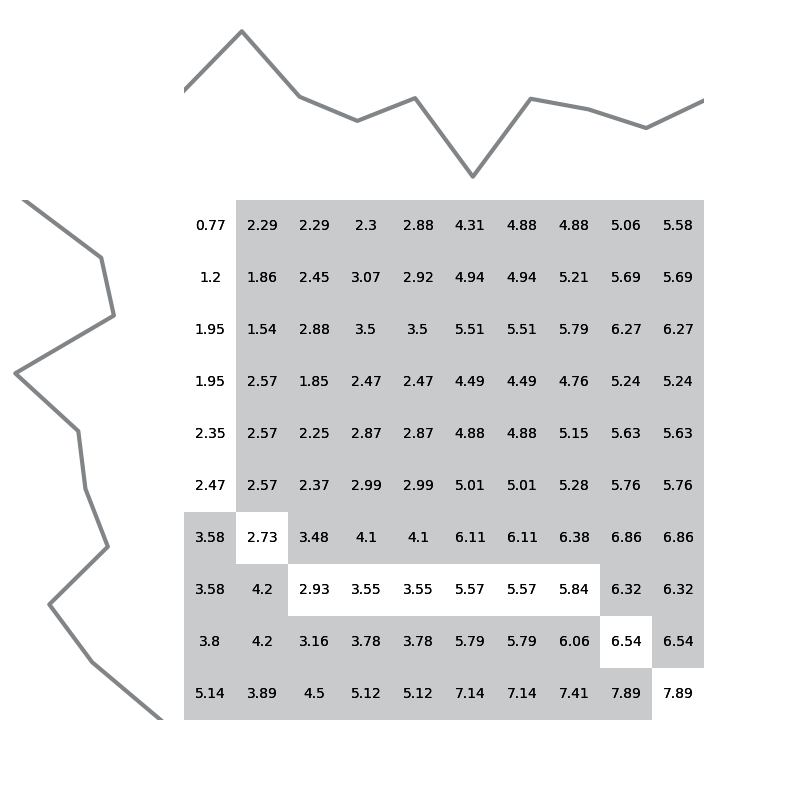}]
    \caption{An example of MSM distance calculation using our example series.}
    \label{fig:example-msm}
\end{figure}

\begin{algorithm}[htbp]
	\caption{MSM(${\bf a},{\bf b}$ (\textit{both series of length $m$}),  $c$ (\textit{minimum cost}), $d$, (\textit{pointwise distance function}))}
\label{algo:msm}
	\begin{algorithmic}[1]
		\State Let $D$ be an $m\times m$ matrix initialised to zero.
		\State $D_{1,1} = d(a_1,b_1)$
		\For{$i \leftarrow  2$ to $m$}
			\State $D_{i,1} = D_{i-1,1}+C(a_i,a_{i-1},b_1)$
		\EndFor
		\For{$i \leftarrow  2$ to $m$}
			\State $D_{1,i} = D_{1,i-1}+C(b_i,a_1,b+{i-1})$
		\EndFor
		
		\For{$i \leftarrow  2$ to $m$}
			\For{$j \leftarrow  2$ to $n$}
                \State $match \leftarrow D_{i-1,j-1}+d(a_i,b_j)$
                \State $insert \leftarrow D_{i-1,j}+C(a_i,a_{i-1},b_j)$
                \State $delete \leftarrow  D_{i,j-1}+C(b_j,b_{j-1},a_i)$
                \State $D_{i,j} \leftarrow \min(match, insert, delete)$
		   \EndFor
		\EndFor
		\Return $D_{m,m}$
	\end{algorithmic}
\end{algorithm}
Move-Split-Merge (MSM)~\citep{stefan13msm} is a distance measure that is conceptually similar to ERP. The core motivation for MSM is that the cost of insertion/deletion in ERP are based on the distance of the value from some constant value, and thus it prefers
inserting and deleting values close to $g$ compared to other values. Algorithm~\ref{algo:msm} shows that the major difference is in the deletion/insertion operations on lines 10 and 11. The move operation in MSM uses the absolute difference rather than the squared distance for matching in ERP. Insert cost in ERP $d(a_i, g)$ is replaced by split operation $C(a_i,a_{i-1},b_j)$, where $C$ is the cost function given in Equation~\ref{eqn:cost}.  If the value being inserted, $b_j$, is between the two values $a_i$ and $a_{i-1}$ being split, the cost is a constant value $c$. If not, the cost is $c$ plus the minimum deviation from the furthest point $a_i$ and the previous point $a_{i-1}$ or $b_{j}$. The delete cost of ERP $d(g, b_j)$ is replaced by the merge cost  $C(b_j,b_{j-1},a_{i})$, which is simply the same operation on the second series. Thus, the cost of splitting and merging values depends on the value itself and adjacent values, rather than treating all insertions and deletions equally as with ERP. The MSM distance between ${\bf a}$ and ${\bf b}$ is given by Equation~\ref{eqn:msm} and illustrated in Figure~\ref{fig:example-msm}.  MSM satisfies triangular inequality and is a metric.
\begin{equation}
d_{MSM}({\bf a,b}) = MSM({\bf a, b},c)
\label{eqn:msm}
\end{equation}

\subsubsection{Time Warp Edit (TWE)}
\begin{figure}[htb]
    \centering
\includegraphics[width=1\textwidth]{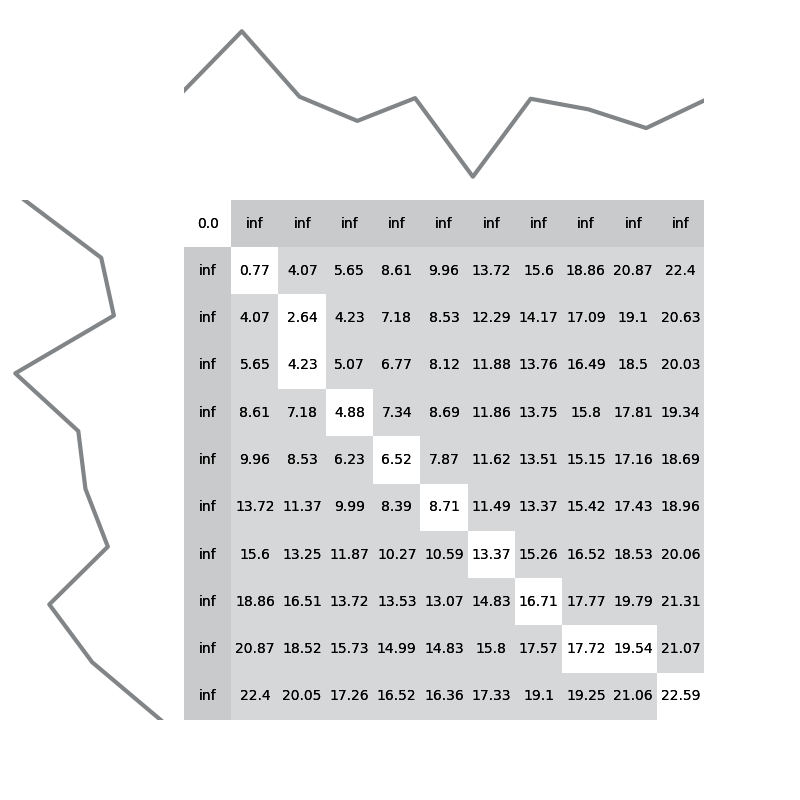}]
    \caption{An example of TWE distance calculation using our example series.}
    \label{fig:example-twe}
\end{figure}

Introduced in~\cite{marteau09stiffness}, Time Warp Edit (TWE) distance is an elastic distance measure described in Algorithm~\ref{algo:twe}. It encompasses characteristics from both warping and editing approaches. The warping, called {\em stiffness}, is controlled by a parameter $\nu$. Stiffness enforces a multiplicative penalty on the distance between matched points in a way that is similar to WDTW, where $\nu = 0$ gives no warping penalty. The stiffness is only weighted this way when considering the match option (line 11). For the delete and insert operations (lines 12 and 13), a constant stiffness ($\nu$) and edit ($\lambda$) penalty are applied since the warping is considered from consecutive points in the same series. Ane exampe is shown in Figure~\ref{fig:example-twe}.

\begin{algorithm}[htbp]
 	\caption{TWE(${\bf a},{\bf b}$ (\textit{both series of length $m$}),  $\lambda$ (\textit{edit cost}), $\nu$ (\textit{warping penalty factor}), $d$, (\textit{pointwise distance function}))}
 \label{algo:twe}
 	\begin{algorithmic}[1]
 		\State Let $D$ be an $m+1\times n+1$ matrix initialised to 0
 		\State $D_{0,0} = 0$
 		\For{$i \leftarrow 1$ to $m$}
 			\State $D_{i,0} = \infty$
 			\State $D_{0,i} = \infty$
        \EndFor
 		\For{$i \leftarrow 1$ to $m$}
 			\For{$j \leftarrow 1$ to $n$}
 				\State $match = D(i-1,j-1)+ d({a_{i},b_{j}})+d({a_{i-1},b_{j-1}}) +2\nu(|i-j|)$
 				\State $delete = D(i-1,j)+d(a_{i},a_{i-1}) + \lambda+\nu$			
 				\State $insert = D(i,j-1)+d(b_{j},b_{j-1}) + \lambda+\nu$	
 		    	\State $D(i,j) = $min$(match,insert, delete)$
 			\EndFor
 		\EndFor
 		\Return $D(m,n)$
 	\end{algorithmic}
 \end{algorithm}
\begin{equation}
d_{TWE}({\bf a,b}) = TWE({\bf a, b},\nu, \lambda)
\label{eqn:twe}
\end{equation}
\begin{table}[htb]
    \centering
    \begin{tabular}{l|c|l|l}
Acronymn                & Metric    & Parameters & Default \\\hline
DTW (Equation~\ref{eqn:dtw})    & No        &   window $w \in [0,1]$ & $w =0.2$ \\
DDTW (Equation~\ref{eqn:ddtw})   & No        &   window $w \in [0,1]$ & $w =0.2$\\
WDTW (Equation~\ref{eqn:wdtw})   & No        & $g \in [0,\infty)$ & $g = 0.05$ \\
DWDTW (Equation~\ref{eqn:dwdtw}) & No        & $g \in [0,\infty)$ & $g = 0.05$ \\
LCSS (Equation~\ref{eqn:lcss}) & No        & x$\epsilon \in [0,\infty)$ & $\epsilon = 0.05$ \\
ERP (Equation~\ref{eqn:erp}) & Yes        &  $g \in [0\ldots 1]$, & $g = 0.05$   \\
EDR (Equation~\ref{eqn:edr}) & No        & $\epsilon \in [0,\infty)$ & $\epsilon = 0.05$\\
MSM (Equation~\ref{eqn:msm}) & Yes        &  $c \in [0,\infty)$ & $c = 1$ \\
TWE (Equation~\ref{eqn:twe}) & Yes        &  $\nu, \lambda \in [0,\infty)$ &  $\nu = 0.05, \lambda = 1$ \\

    \end{tabular}
    \caption{Summary of distance functions, their parameters and the default values}
    \label{tab:parameters}
\end{table}

A summary of distance function parameters and their default values in our implementations is given in Table~\ref{tab:parameters}. DTW is sensitive to the window parameter~\citep{dau18optimizing} and large windows can cause pathological warpings. Based on experimental results~\citep{ratanamahatana05threemyths}, we set the default warping window to 0.2. WDTW uses a default scale parameter value for $g$ of 0.05, based on results reported in~\cite{jeong11weighted}.
LCSS and EDR both have an $\epsilon$ parameter that is a threshold for similarity. If the difference in two random variables is below $\epsilon$, the observations are considered a match. The variability in parameter effects depending on the values of the series is one of the arguments for normalising all series. We set the default $\epsilon$ to 0.05. MSM has a single parameter, $c$, to represent the cost of the move-split-merge operation. This is set to $1$ based on the original paper.  TWE $\lambda$ is analogous to the $c$ parameter in MSM, so we also set it to one, whereas $\nu$ is related to the weighting parameter of $WDTW$, so we set it to 0.05.


\subsection{Averaging Time Series}
\label{sec:averaging}
$k$-means clustering requires a form of characterising a set of time series to form an exemplar. The standard approach for $k$-means is simply to find the mean of the current members of a cluster over time points. This is appropriate if the distance function Euclidean distance, since the average centroid is the series with the minimal Euclidean distance to members of the cluster. However, if cluster membership is being assigned based on an elastic distance measure, the average centroid may misrepresent the elements of a cluster; it is unlikely to be the series that minimizes the elastic distance to cluster members. Dynamic time warping with Barycentre averaging (DBA)~\citep{petitjean11dba} (see Algorithm~\ref{algo:ba}) was proposed to overcome this limitation in the context of DTW.
\begin{algorithm}[htbp]
\caption{DTW Barycentre Averaging(${\bf c}$, {\em the initial average sequence}, ${\bf X_p}$, {\em $p$ time series to average}.}
    \label{algo:ba}
	\begin{algorithmic}[1]
	\State Let dtw\_path be a function that returns the a list of tuples that contain the indexes of the warping path between two time series.
	\State Let $W$ be a list of empty lists, where $W_i$ stores the values in ${\bf X_p}$ of points warped onto centre point $c_i$.
    \For{$x \in {\bf X_p}$}
	    \State $P \leftarrow dtw\_path({\bf x}, {\bf c})$
	    \For{$(i,j) \in P$}
		    \State $W_{i} \leftarrow W_{i} U x_j$
	    \EndFor
    \EndFor
	\For{$i \leftarrow  1$ to $m$}
	    \State $c_i \leftarrow mean(W_i)$
    \EndFor
	\Return $c$
	\end{algorithmic}
\end{algorithm}

 DBA is a heuristic to find a series that minimizes the elastic distance to cluster members rather than the Euclidean distance. Starting with some initial centre, DBA works by first finding the warping path of series in the cluster onto the centre. It then updates the centre, by finding which points were warped onto each element of the centre for all elements of the cluster, then recalculating the centre as the average of these warped points.

Suppose function $f({\bf a}, {\bf b})$ returns the warping path of indexes $$P=<(e_1,f_1),(e_2,f_2),\ldots,(e_s,f_s)>$$ generated by dynamic time warping. Given an initial centre ${\bf c}=<c_1, \ldots, c_m>$, DBA is described by Algorithm~\ref{algo:ba}. It warps each series onto $c$ (line 4), then from the warping path associates the value in the series with the value in the barycentre (line 5 and 6). Once finished for all series, the average of values warped to each index of the centroid is taken. This is meant to better characterise the cluster members in the iterative partitioning of algorithms.


\section{Implementation}

We have implemented the nine distance functions used in the evaluation in the Python scikit learn compatible package aeon~\footnote{https://github.com/aeon-toolkit/aeon}. The distance functions are all implemented using Numba tool~\footnote{https://numba.pydata.org/}, which uses just in time compilation to dynamically convert Python to C. There are of course numerous open source packages that offer some form of elastic distance functions, most commonly variants of DTW. Table~\ref{tab:packages} lists some of the most popular packages with DTW implementations and summarises which other elastic distance measures they contain. We have confirmed equivalence of DTW results with these packages.  
\begin{table}[htb]
\centering
    \begin{tabular}{c|c|c|c|c|c|c|c|c|c}
package  & dtw & wdtw & ddtw & wddtw & edr & erp & msm & lcss & twe  \\
\hline
 & \multicolumn{9}{c} {Univariate} \\ \hline
aeon & Y & Y & Y & Y& Y & Y& Y & Y& Y  \\
tslearn & Y & Y & N & N & N & N & N & N& N \\
dtw-python & Y & N & N & N & N & N & N & N& N  \\
rust-dtw & Y & N & N & N & N & N & N & N& N \\ \hline
 & \multicolumn{9}{c} {Multivariate} \\ \hline
aeon & Y & Y & Y & Y& Y & Y& N & Y& Y  \\
tslearn & N & N & N & N & N & N & N & N& N \\
dtw-python & N & N & N & N & N & N & N & N& N  \\
rust-dtw & N & N & N & N & N & N & N & N& N \\
    \end{tabular}
    \caption{Summary of elastic distance function availability in five Python packages.}
    \label{tab:packages}
\end{table}
\begin{table}[htb]
\begin{center}

\begin{tabular}{c|c|c|c|c}
length &    aeon &   tslearn &  dtw-python &  rust-dtw \\
\hline
        1000 & 1.89     &  2.10 &  8.09 &    1.55 \\
        2000 &  6.50    &   6.25 &  31.38 &    6.00 \\
        3000 &  14.65 &   13.54 &  69.31 &    13.44 \\
        4000 &  25.07 &   25.97 &  121.33 &    24.00 \\
        5000 &  38.96 &   37.39 &  191.91 &    37.42 \\
        6000 &  57.26 &   54.44 &  272.14 &    55.24 \\
        7000 &  73.10 &  73.59 &  342.45 &    71.57 \\
        8000 &   92.25  &  91.94 &  451.83 &    93.49 \\
        9000 &   123.11      &  117.81       &   583.75       &  119.54  \\
        10000 &   175.20     &  163.55       & 780.52     &    167.74 \\

\end{tabular}
    \caption{Time (in seconds) to perform 200 full window DTW distance calculations with random series of length 1000 to 10000.}
    \label{tab:uni-timing}
\end{center}
\end{table}
The time taken to perform 200 DTW distance calculations on random series of length 1000 to 10,000 on a desktop PC are shown in Table~\ref{tab:uni-timing}. aeon run time is very similar to tslearn and rust. dtw-python is approximately five times slower than the other packages. aeon offers the widest range of elastic distance measures with equivalent run time to the other packages.

Distance functions can be used directly, either by explicitly importing them or by using the distance factory provided. A Jupyter notebook with example usages is available on the associated repository. Listing~\ref{distances} shows how to calculate the DTW distance and alignment path for the series used in Figure~\ref{fig:example-distances}. It also shows how to use the factory to get and call a distance function. The distance factory avoids the repeated numba compilation of distance functions that can occur with different parameters, so it is our recommended method for creating a distance function.
There is also the option to find the pairwise distance matrix.
\lstset{style=python_jay}
\begin{lstlisting}[language=Python, caption=A simple example of finding distances and alignments in \texttt{aeon}, label=distances]
from aeon.distances import dtw_distance, dtw_alignment_path
from aeon.distances import distance_factory, distance, distance_alignment_path
from aeon.distances import pairwise_distance
import numpy as np
a = np.array([0.018, 1.537, -0.141, -0.761, -0.177, -2.192, -0.193, -0.465, -0.944, -0.240])
b = np.array([-0.755, 0.446, 1.198, 0.171, 0.564, 0.689, 1.794, 0.066, 0.288, 1.634])
d1 = dtw_distance(a, b)
d2 = dtw_distance(a, b, window=0.2)
d3 = distance(a, b, metric='dtw', window=0.2)
p1 = dtw_alignment_path(a, b)
p2 = dtw_alignment_path(a, b, window=0.2)

# For repeated use create a numba compiled callable
# FIX THE BELOW
p3 = distance_alignment_path(a, b, metric='msm')
dtw_numba = distance_factory(a, b, metric='dtw')
twe_numba = distance_factory(a, b, metric="twe")
d3 = dtw_numba(a, b)
dist_paras = {"mu":0.5, "lmda":0.5}
d4 = twe_numba(a, b, dist_paras)

# To find a distance matrix between one or two 2D matrix, use pairwise_distance
pair = np.array([a, b])
dist = pairwise_distance(pair,metric="twe")
# Equivalent to this
dist = pairwise_distance(pair, pair, metric="twe")

\end{lstlisting}

We have implemented the $k$-means and $k$-medoids clusterers in aeon. These can be easily used and configured, as shown in Listing~\ref{clustering}.
\lstset{style=python_jay}
\begin{lstlisting}[language=Python, caption=An example of using kmeans and kmedoids in \texttt{aeon}, label=clustering]
from aeon.clustering.k_means import TimeSeriesKMeans
from aeon.clustering.k_medoids import TimeSeriesKMedoids
from aeon.datasets import load_unit_test
trainX, trainY = load_unit_test(split="test",return_type="np2D")
testX, testY = load_unit_test(split="train",return_type="np2D")
clst1 = TimeSeriesKMeans()
clst2 = TimeSeriesKMeans(
            averaging_method="dba",
            metric="dtw",
            distance_params={"window":0.1},
            n_clusters=2,
            random_state=1,
        )
clst3 =  TimeSeriesKMedoids()
clst4 =  TimeSeriesKMedoids(
            metric="dtw",
            distance_params={"window": 0.2},
            n_clusters=len(set(trainY)),
            random_state= 1,
        )
clst1.fit(trainX)
pred = clst1.predict(testX)
print(pred)
\end{lstlisting}
To conduct an experiment, we generate results in a standard format.
\lstset{style=python_jay}
\begin{lstlisting}[language=Python, caption=An example of running an experiment in \texttt{aeon}. Random state is set to 1 throughout our experiments to faciliate reproducibility. label=clustering]
from aeon.benchmarking.experiments import run_clustering_experiment

run_clustering_experiment(
        trainX,
        clst1,
        results_path="_contrib/temp/",
        trainY=trainY,
        testX=testX,
        testY=testY,
        cls_name="kmeans",
        dataset_name="UnitTest",
        resample_id=0,
        overwrite=False,
    )


\end{lstlisting}
Currently, we collate these results files using the associated Java package tsml~\footnote{https://github.com/uea-machine-learning/tsml}. An example of this is available on this paper's repository.

\section{Methodology}
\label{sec:methods}
Our experiments involve two steps that are not universally performed in the clustering literature: we normalise all series to zero mean and unit variance; and we train and test on separate folds.  The issue of whether to always normalise is an open question for TSC, since some discriminatory features may be in the scale or variance. Whether to normalise is often treated as a parameter for TSC. However, for clustering we believe it makes sense to always normalise when comparing over multiple datasets; scale is much more likely to dominate an unsupervised algorithm and confound the comparison. Conversely, evaluating on unseen test data is essential for any TSC comparison. The issue is less clear cut with clustering, which is used more as an exploratory tool than a predictive model. We believe the problem of overestimating performance through assessing on training data applies to clustering data. By evaluating algorithms using training and testing data we get an idea of a clustering algorithms ability to generalise and we can simply place the clustering comparison in the context of classification research. We perform all training on the default train split provided, but present the results on both the train set and test set.

\subsection{Data}
\label{sec:data}

We experiment with time series data in the University of California, Riverside (UCR) archive~\citep{dau19ucr}\footnote{https://timeseriesclassification.com/}. We restrict our attention to univariate time series, and in all experiments we use 112 of the 128 datasets from the UCR time series archive. We exclude datasets containing series of unequal length or missing values. We also remove the Fungi data, which only provides a single train case for each class label. We report results using the six performance measures described in Section~\ref{sec:evaluation} on the both the train sample and the test set.

Using the class labels to assessing performance on some of these data may be unfair. For some problems, clustering algorithms naturally find clusters that are independent of the class labels, but are nevertheless valid. For example, the GunPoint data set was created by a man and a woman drawing a gun from their belt, or pretending to draw a gun. The class labels are Gun/No Gun. However, many clusterers finds the Man/Woman clusters rather than Gun/No Gun. Without supervision, this is a perfectly valid clustering, but it will score approximately 50\% accuracy, since the man and woman cases are split evenly between Gun/No Gun. This is an inherent problem with attempting to evaluate exploratory, unsupervised algorithms by comparing them with what we know to be true a priori: if a clustering simply finds what we already know, its utility is limited. Furthermore, as observed in~\citep{lafabregue22deep}, some of the datasets have the same time series but with different labels. We aim to mitigate against these problems by using a large number of problems, but also explore the effect of removing problems where the algorithms perform little better than forming a single cluster.

\subsection{Clustering Metrics}
\label{sec:evaluation}

\textbf{Clustering accuracy (CL-ACC)}, like classification accuracy, is the number of correct predictions divided by the total number of cases. To determine whether a cluster prediction is correct, each cluster has to be assigned to its best matching class value. This can be done naively, taking the maximum accuracy from every permutation of cluster and class value assignment $S_k$.
$$
    ACC(y,\hat{y}) = \max_{s \in S_k} \frac{1}{|y|} \sum_{i=1}^{|y|}
    \begin{cases}
        1, & y_i = s(\hat{y}_i) \\
        0, & \text{otherwise}
    \end{cases}
    \label{eqn:accuracy}
$$

Checking every permutation like this is prohibitively expensive, however, and can be done more efficiently using combinatorial optimization algorithms for the assignment problem. A contingency matrix of cluster assignments and class values is created, and turned into a cost matrix by subtracting each value of the contingency matrix from the maximum value. Clusters are then assigned using an algorithm such as the Hungarian algorithm~\citep{kuhn55hungarian} on the cost matrix. If the class value of a case matches the assigned class value of its cluster, the prediction is deemed correct, else it is incorrect. As classes can only have one assigned cluster each, all cases in unassigned clusters due to a difference in a number of clusters and class values are counted as incorrect.

The \textbf{rand index (RI)} works by measuring the similarity between two sets of labels. This could be between the labels produced by different clustering models (thereby allowing direct comparison) or between the ground truth labels and those the model produced. The rand index is the number of pairs that agree on a label divided by the total number of pairs.



One of the limiting factors of RI is that the score is inflated, especially when the number of clusters is increased. The \textbf{adjusted rand index (ARI)} compensates for this by adjusting the RI based on the expected scores on a purely random model.



The \textbf{mutual information (MI)}, is a function that measures the agreement of the two clusterings or a clustering and a true labelling, based on entropy. \textbf{Normalised mutual information (NMI)} rescales MI onto $[0,1]$, and \textbf{adjusted mutual information (AMI)} adjusts the MI to account for the class distribution.



The \textbf{Davies-Bouldin index} is an unsupervised measure we employ for tuning clusterers. It compares the between cluster variation with the inter cluster variation, with a higher score awarded to a clustering where there is good separation between clusters.

To compare multiple clusterers on multiple datasets, we use the rank ordering of the algorithms we use an adaptation of the critical difference diagram~\citep{demsar06comparisons}, replacing the post-hoc Nemenyi test with a comparison of all classifiers using pairwise Wilcoxon signed-rank tests, and cliques formed using the Holm correction recommended by~\citep{garcia08pairwise,benavoli16pairwise}. We use $\alpha=0.05$ for all hypothesis tests. Critical difference diagrams such as those shown in Figure~\ref{fig:test-kmeans} display the algorithms ordered by average rank of the statistic in question and the groups of algorithms between which there is no significant difference (cliques). So, for example, in Figure~\ref{fig:test-kmeans}(c), MSM has the lowest average rank of 3.8973, and is not in a clique, so has significantly better ARI than the other algorithms. TWE, ERP, WDTW and ED are all grouped into a clique, which means there is no pairwise significant difference between them. LCSS, EDR, WDDTW and DTW form another clique, and DDTW is significantly worse than all other algorithms.



\section{Results}
\label{sec:results}

We report a sequence of three sets of experiments. Firstly, in Section~\ref{sec:kmeans} we compare $k$-means clustering using 10 different distance functions and put these in the context of classification. Secondly, in Section~\ref{sec:kmedoids} we report the $k$-medoids results. We address the questions of whether the same pattern of performance seen in $k$-means is observable in $k$-medoids and whether $k$-medoids is generally more effective than $k$-means. Thirdly, in Section~\ref{sec:dba} we evaluate whether we can recreate the reported improvement to DTW of DBA~\ref{sec:dba} and assess how this improvement compares to our previous experiments. 

\subsection{Elastic distances with $k$-means}
\label{sec:kmeans}
\begin{figure}[htb]
        \centering
        \begin{tabular}{c c}
        \includegraphics[width=0.5\linewidth,trim={3cm 0cm 3cm 0cm},clip] {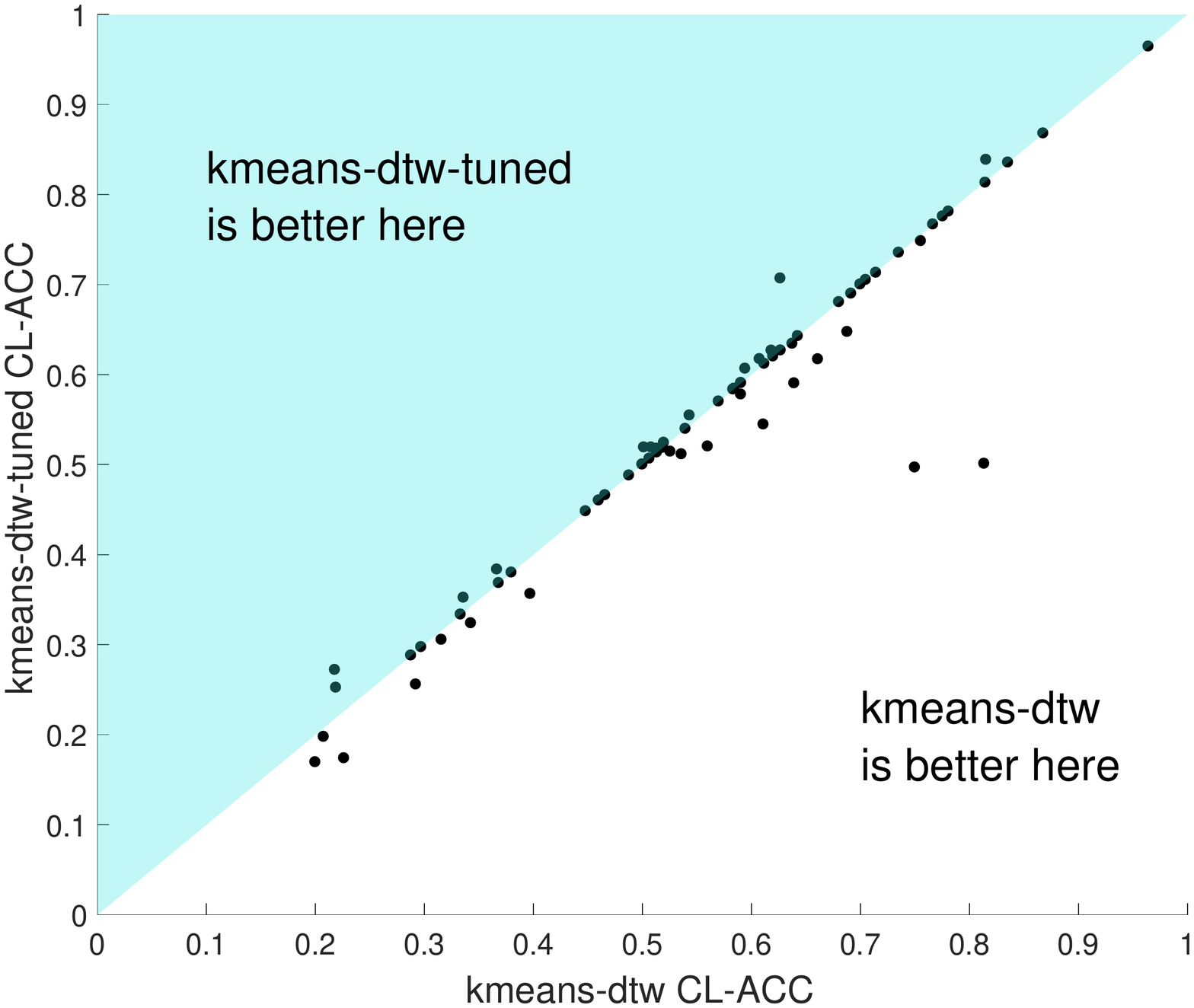} &
        \includegraphics[width=0.5\linewidth,trim={3cm 0cm 3cm 0cm},clip] {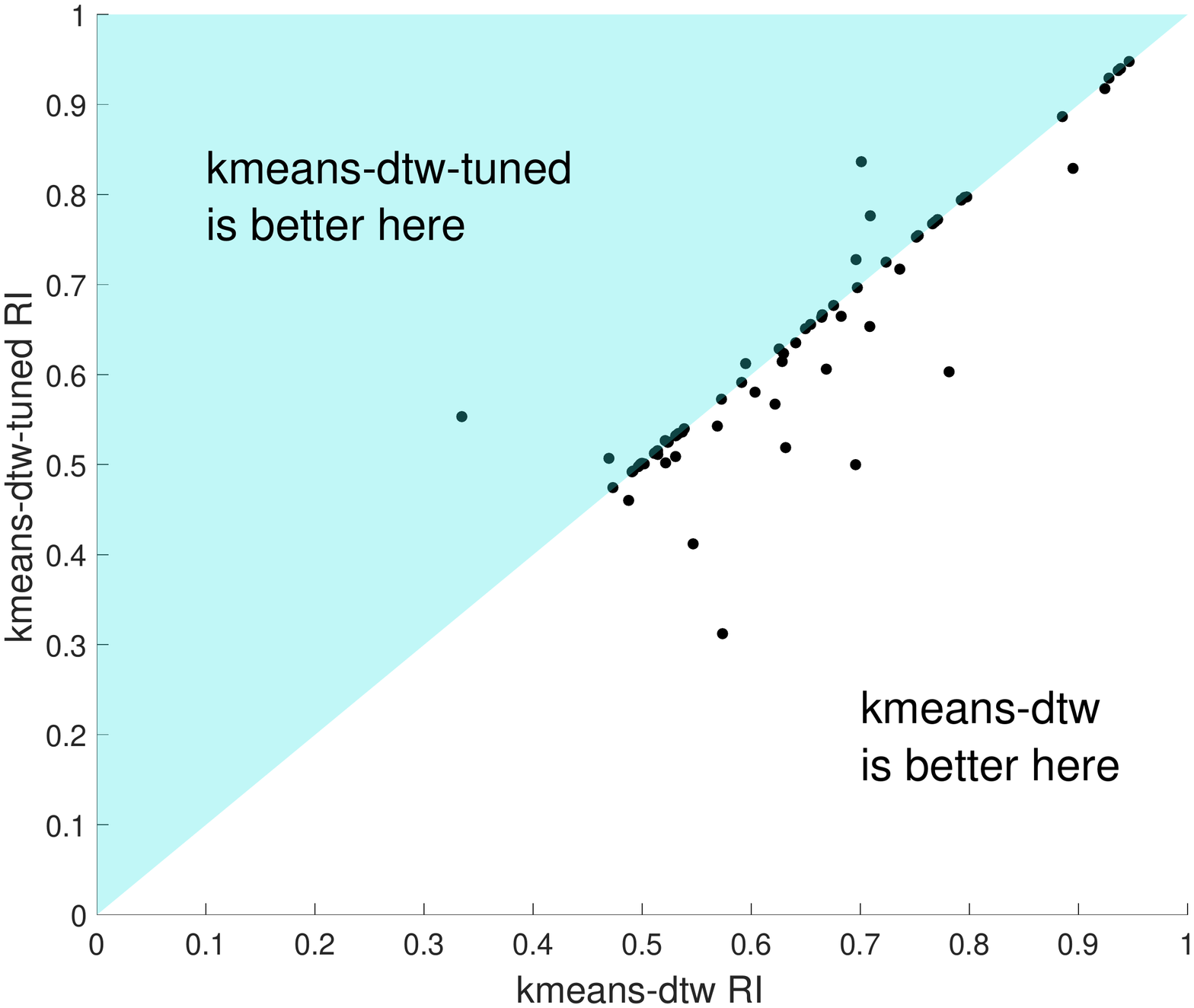}
\end{tabular}
        \caption{Scatter plots of accuracy and rand index for DTW vs Tuned DTW.}
        \label{fig:tune-dtw}
    \end{figure}

The first set of experiments involve comparing alternative distance functions with $k$-means clustering using the arithmetic mean of series. Our primary aim is to investigate whether there are any significant differences between the measures when used on the UCR univariate classification datasets.
For each experiment, we ran $k$-means clustering with the default aeon parameters (random initialisation, maximum 300 iterations, 10 restarts, averaging method is the mean) with Euclidean distance and the nine elastic distance measures, set up with the default parameters listed in Table~\ref{tab:parameters}.
    \begin{figure}[htb]
        \begin{tabular}{cc}
        \includegraphics[width=0.5\linewidth,trim={4.5cm 4cm 4cm 4cm},clip] {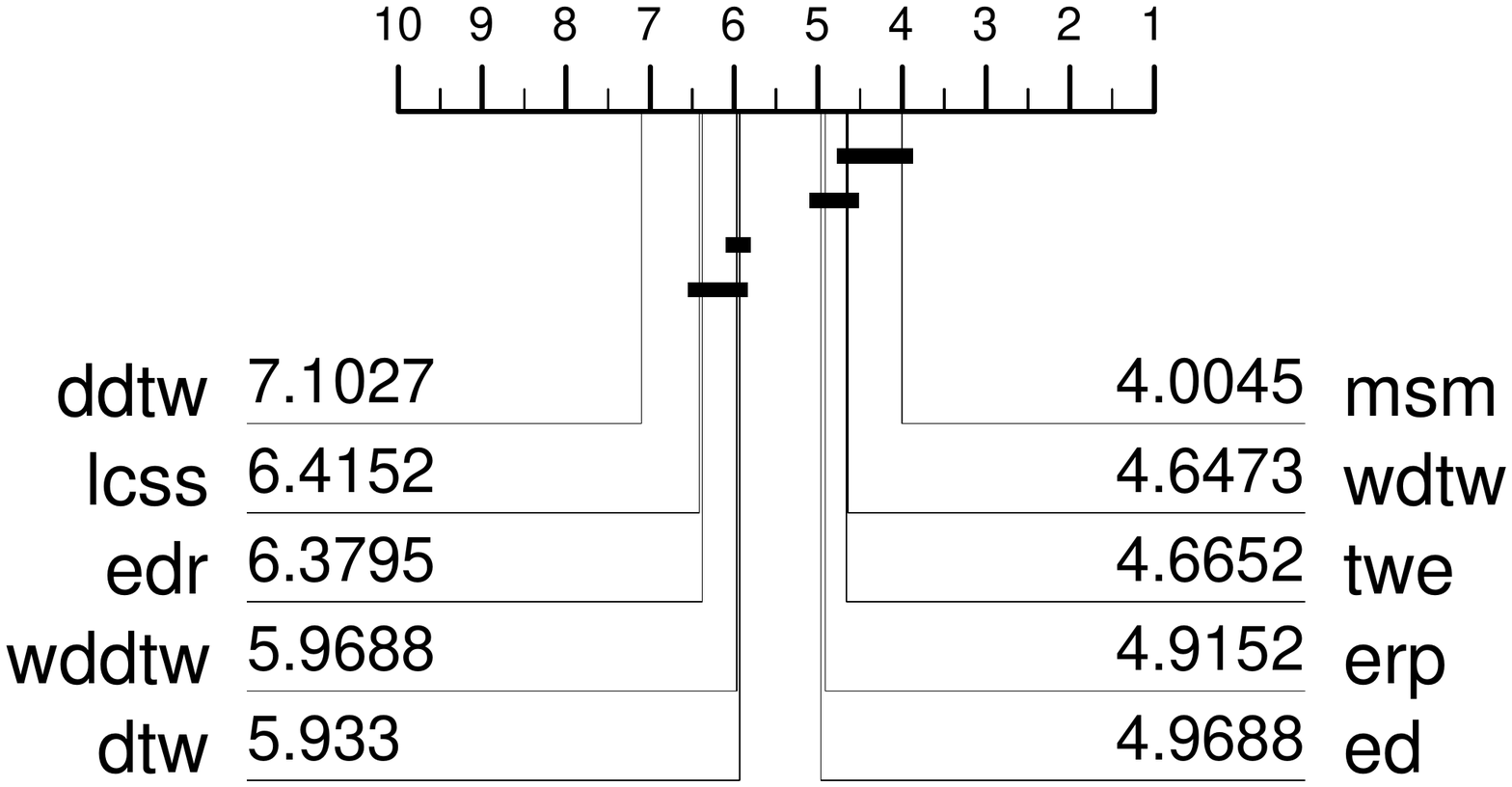} &
        \includegraphics[width=0.5\linewidth,trim={4.5cm 4cm 4cm 4cm},clip] {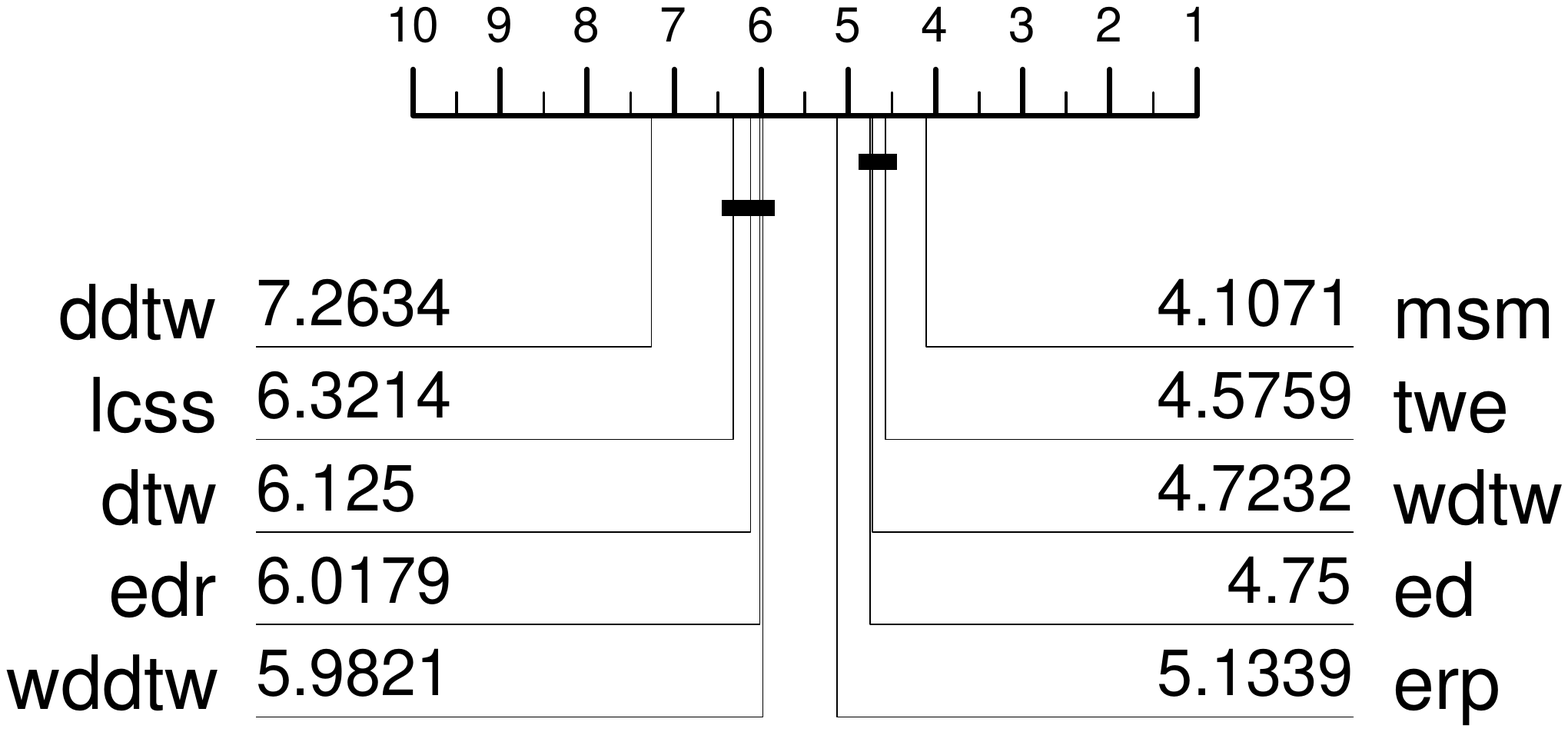}         \\
        (a) Accuracy& (b) Rand Index (RI)\\
        \includegraphics[width=0.5\linewidth,trim={4.5cm 4cm 4cm 4cm},clip] {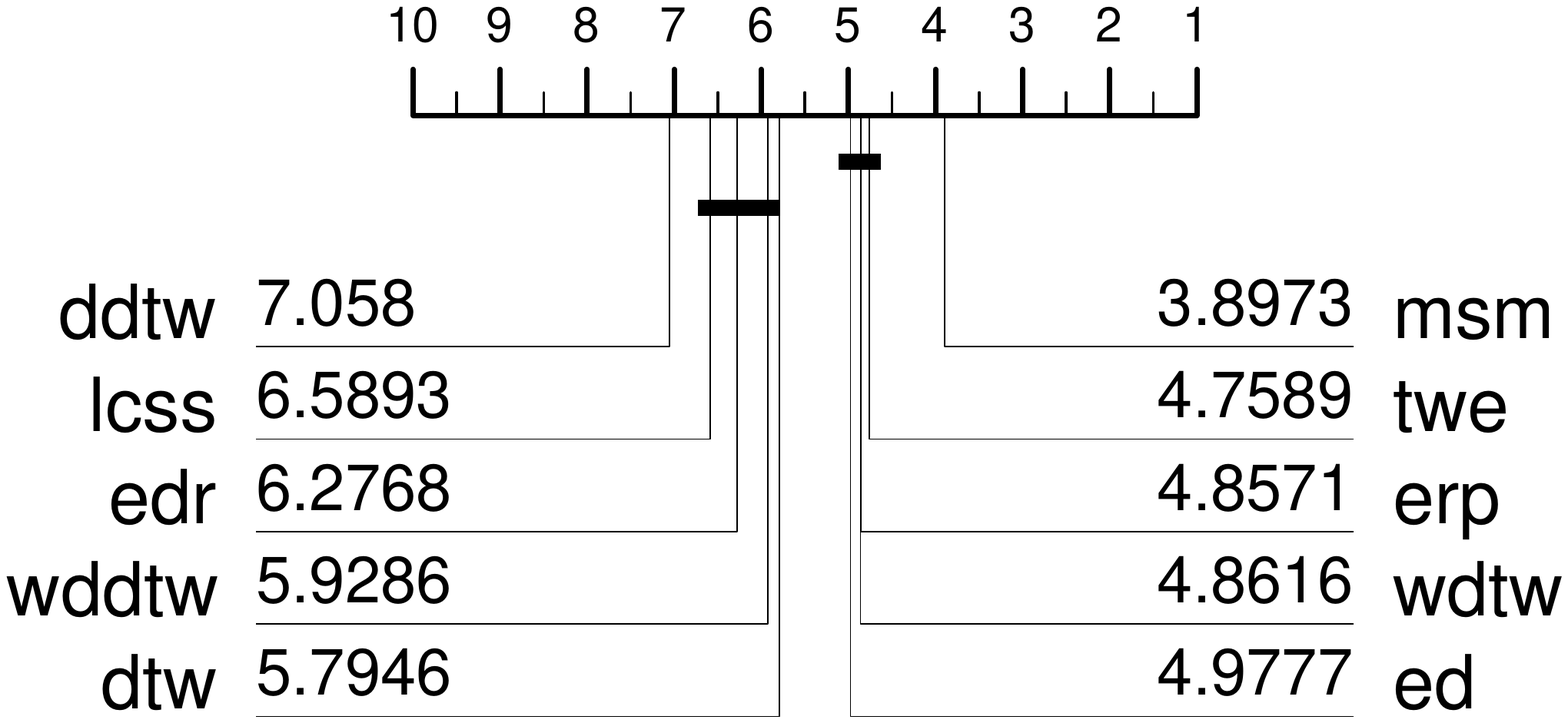} &
        \includegraphics[width=0.5\linewidth,trim={4.5cm 4cm 4cm 4cm},clip] {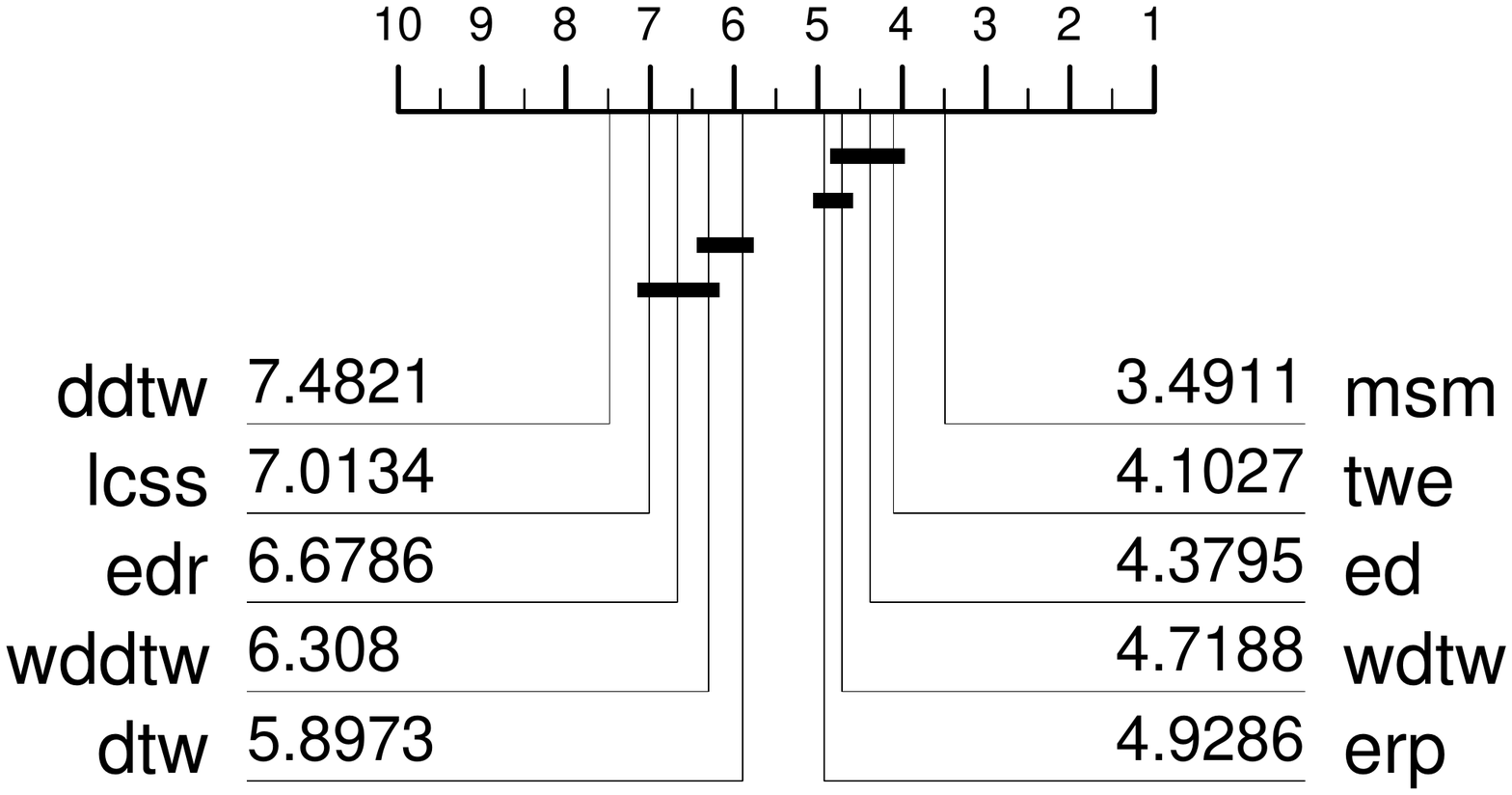}   \\
        (c) Adjusted RI & (d) Mutual Information (MI) \\
        \includegraphics[width=0.5\linewidth,trim={4.5cm 4cm 4cm 4cm},clip] {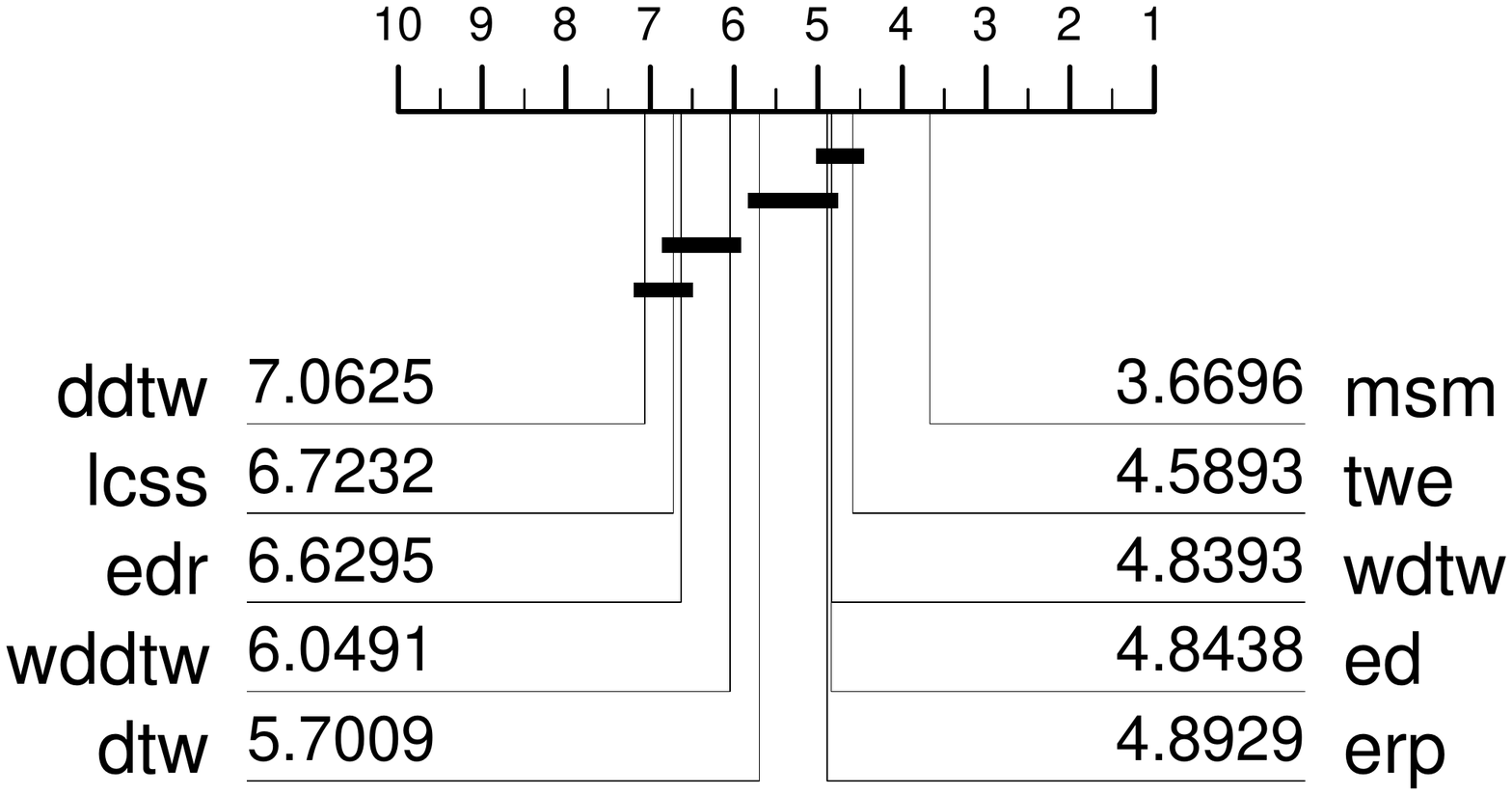} &
        \includegraphics[width=0.5\linewidth,trim={4.5cm 4cm 4cm 4cm},clip]{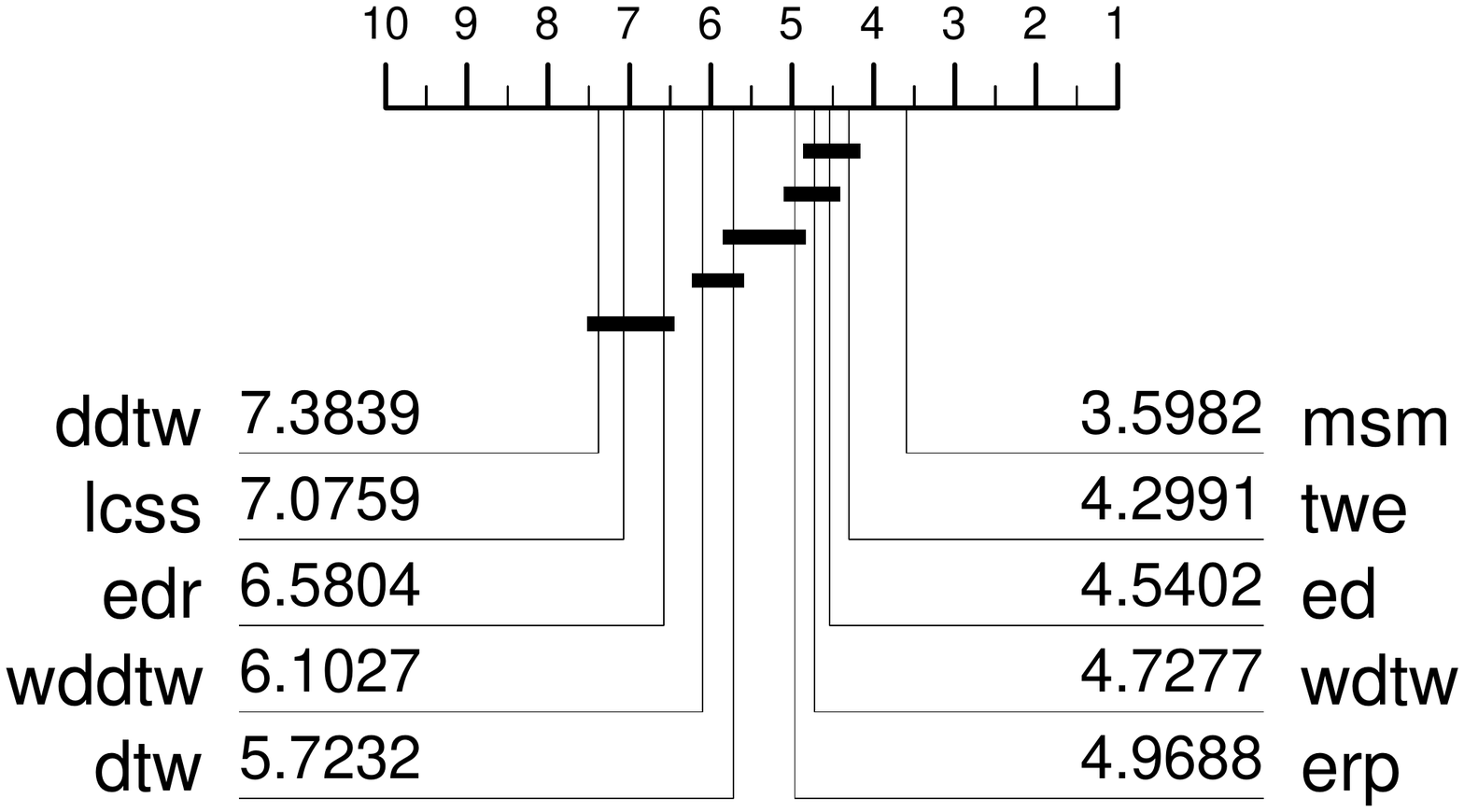}
        \\
        (e) Adjusted MI & (f) Normalised MI \\
\end{tabular}
        \caption{Critical difference diagrams for $k$-means clustering using nine different elastic distance functions on 112 UCR problems using six performance measures. Results are on derived from the test files.}
        \label{fig:test-kmeans}
    \end{figure}
Figure~\ref{fig:test-kmeans} shows the summarised results on the test data. Full results for the test data are given in the Appendix in Table~\ref{tab:kmeans-full1}. These results and code to reproduce them are in the associated GitHub repository. The first obvious conclusion from these results is that MSM is the best performing distance measure: it is significantly better than the other nine distances with all performance measures except accuracy, where it is top ranked. This mirrors similar findings when comparing distance measures with one nearest neighbour classification (see Section 4 of~\cite{bagnall17bakeoff}). Nevertheless, it is still surprising, and we believe not widely known in the time series research community. The second stand out result is that DTW is significantly worse than Euclidean distance. This is particularly surprising, particularly given the prevalence of DTW in the TSCL literature. It is even more notable when we observe that WDTW is significantly better than DTW. To verify this particular result, we reran the clustering experiments using the Java toolkit tsml\footnote{https://github.com/uea-machine-learning/tsml} version of DTW in conjunction with the WEKA $k$-means clusterer, and found no significant difference in the results. Furthermore, we have checked that the aeon DTW distance implementation produces the same distances as other the other implementations listed in Table~\ref{tab:packages}. Further experimentation has shown that full window DTW performs worse than DTW constrained to 20\%. It appears that windowing is significantly less effective than weighting for clustering. This is not true of classification, and reflects the greater uncertainty with unsupervised learning.
Tuning significantly improves the performance of DTW 1-NN classification, so it is possible that it may also improve DTW based clustering. To tune a clusterer we need an unsupervised cluster assessment algorithm. We use the Davies-Bouldin score since it is popular and available in scikit-learn. We evaluate DTW for ten possible windows on even intervals in the range $[0,1)$ and use the window with the highest Davies-Bouldin score for our final clustering. Figure~\ref{fig:tune-dtw} shows the scatter plot of DTW vs tuned DTW. There is no significant difference between tuned and untuned DTW. The inherent uncertainty of clustering and the lack of sensitivity of the performance measure means we do not see the improvement through tuning that we observe in classification with elastic distances.



    \begin{figure}[htb]
        \centering
        \begin{tabular}{c c}
        \includegraphics[width=0.5\linewidth,trim={4.5cm 11cm 4.5cm 11cm},clip] {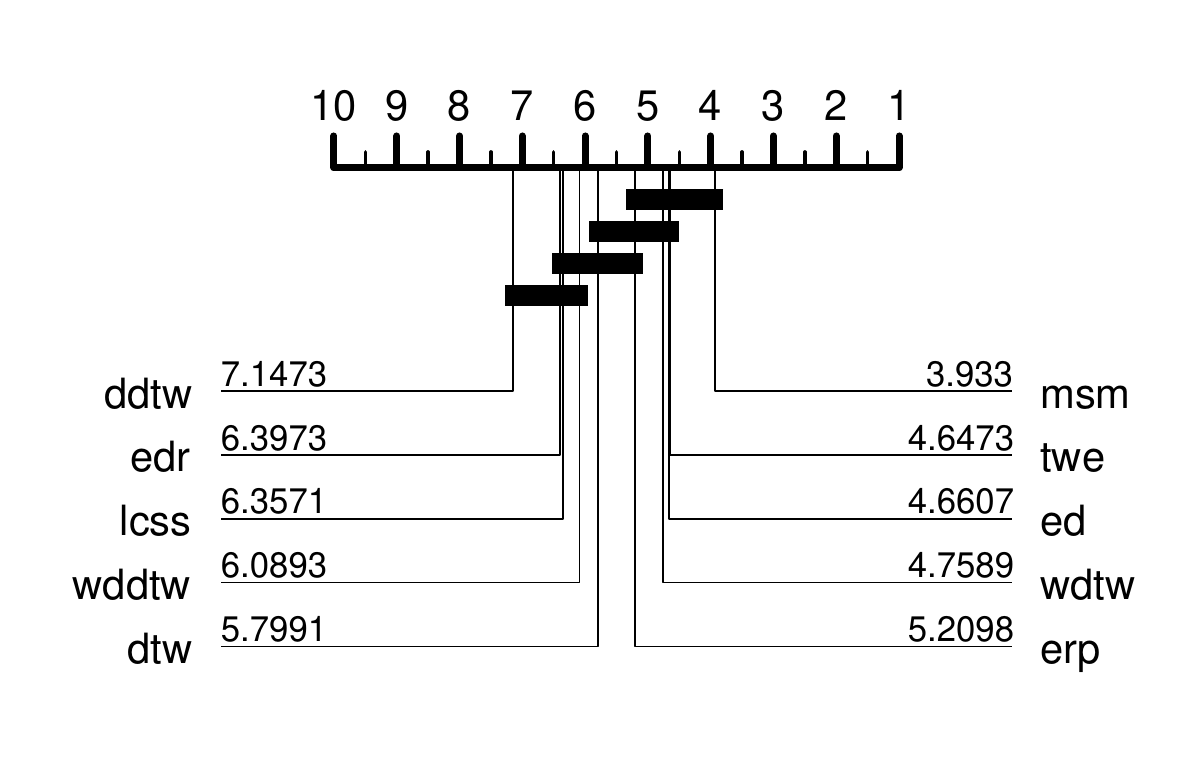} &
        \includegraphics[width=0.5\linewidth,trim={4.5cm 11cm 4.5cm 11cm},clip] {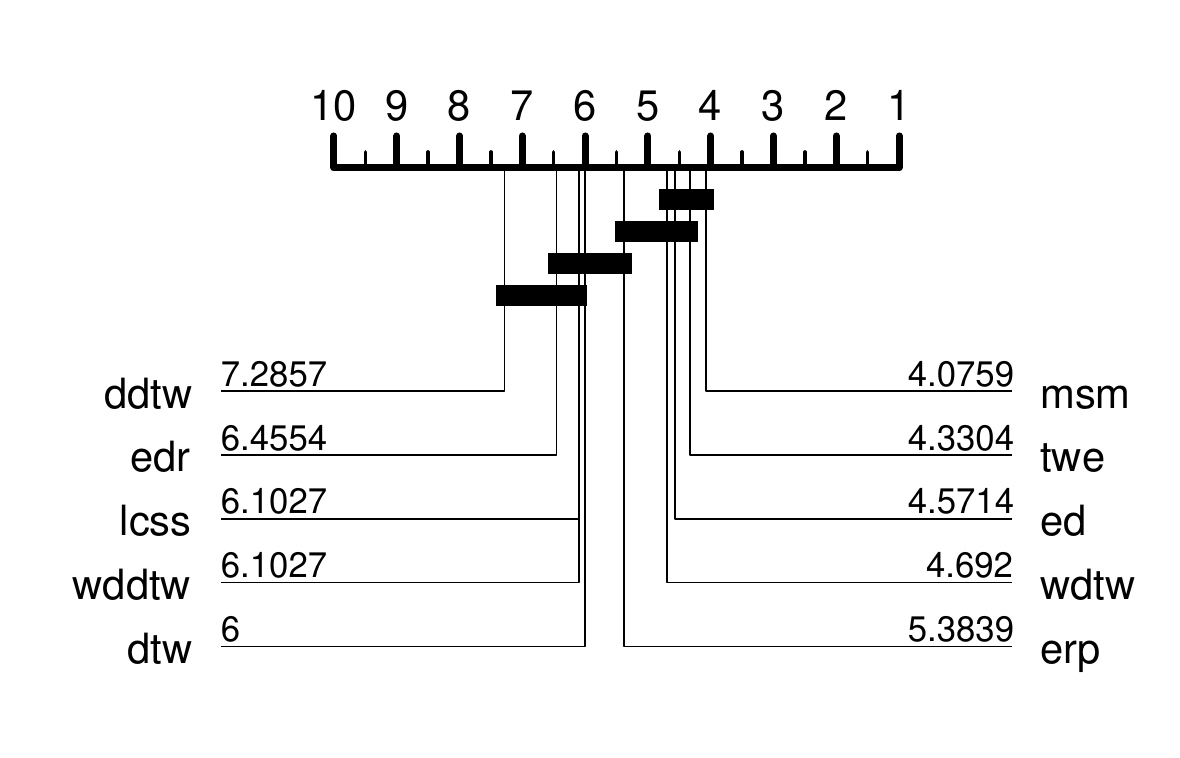}
        \\
        (a) Accuracy
        & (b) Rand Index \\
        \includegraphics[width=0.5\linewidth,trim={4.5cm 11cm 4.5cm 11cm},clip] {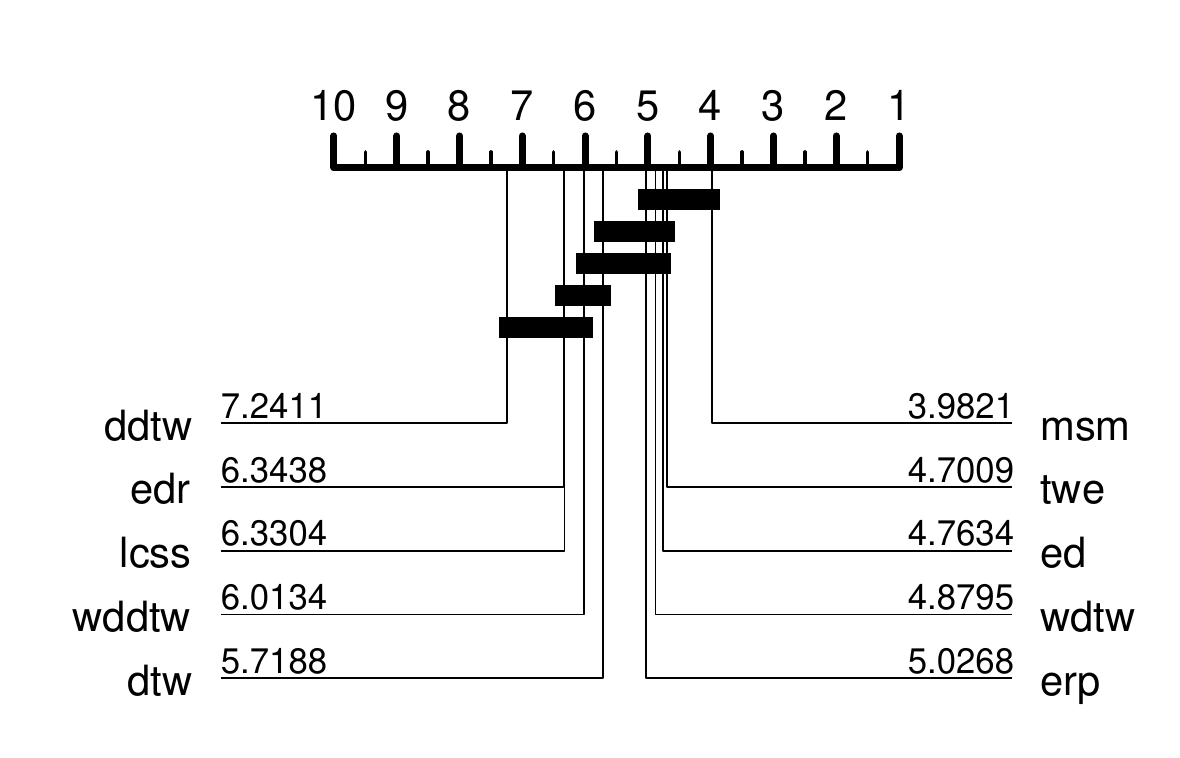} &
        \includegraphics[width=0.5\linewidth,trim={4.5cm 11cm 4.5cm 11cm},clip] {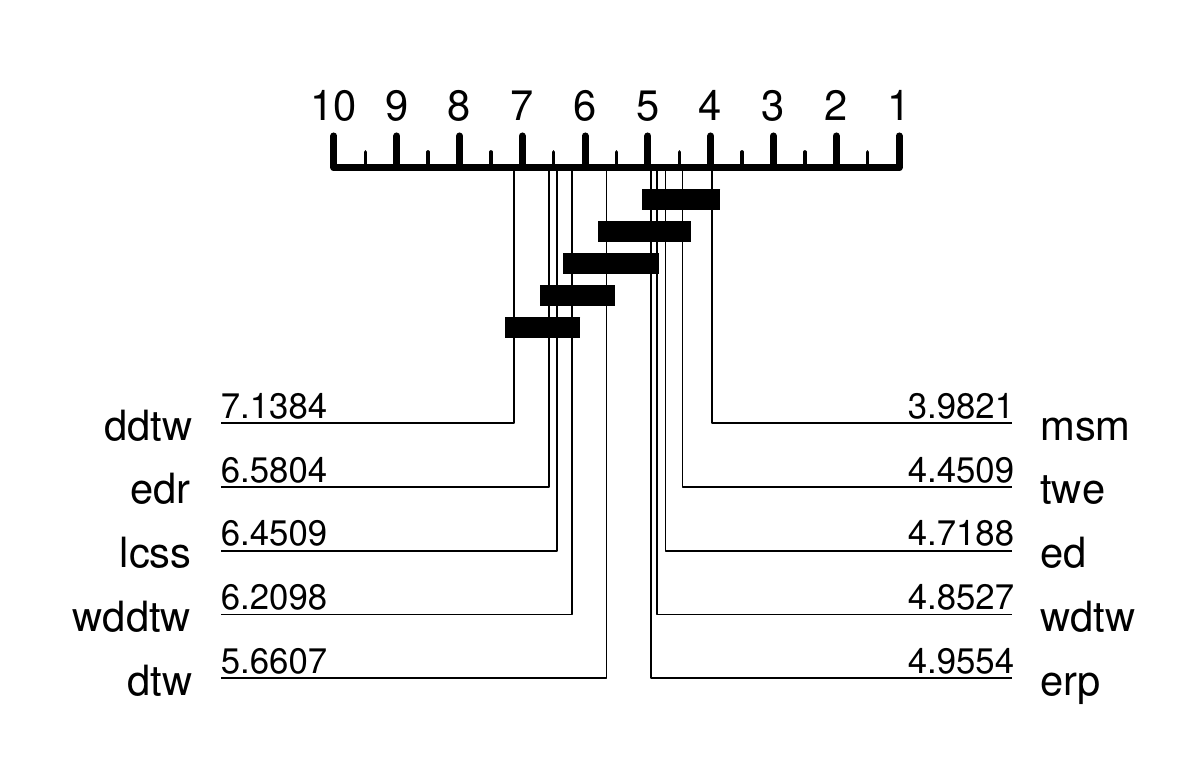}
        \\
        (c) ARIS
        & (d) AMIS \\
        \includegraphics[width=0.5\linewidth,trim={4.5cm 11cm 4.5cm 11cm},clip] {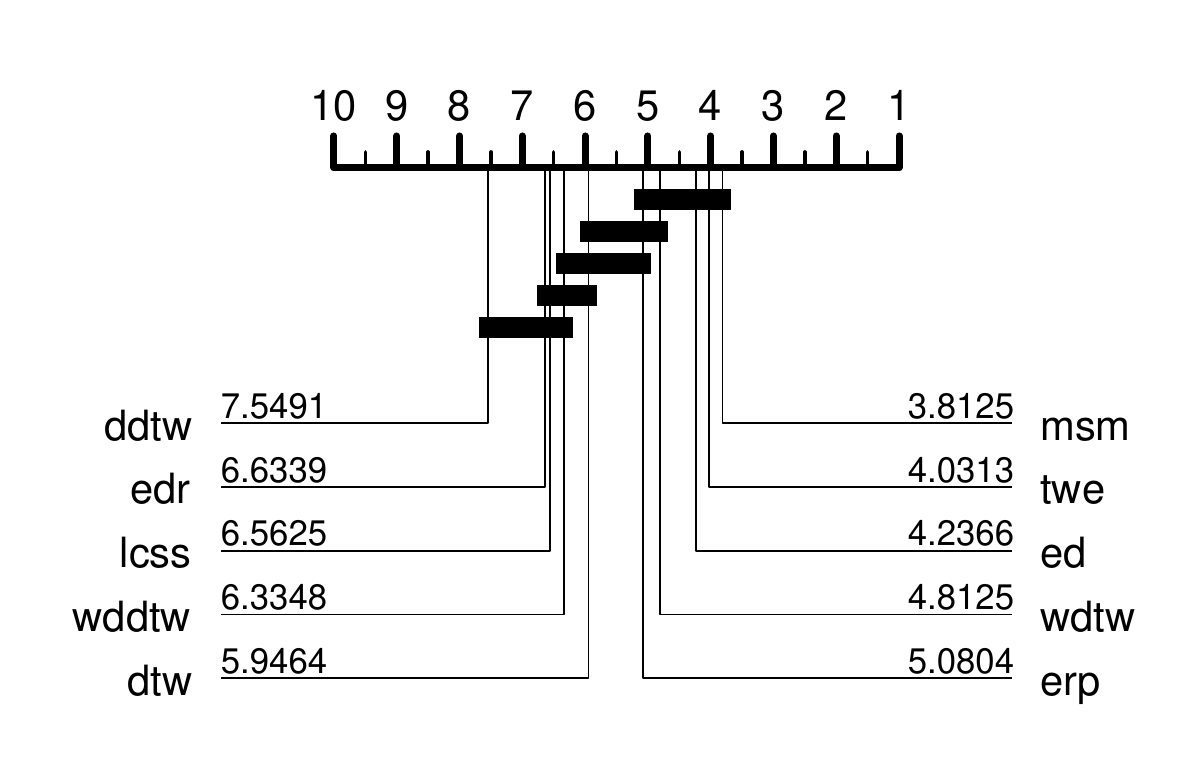} &
        \includegraphics[width=0.5\linewidth,trim={4.5cm 11cm 4.5cm 11cm},clip]{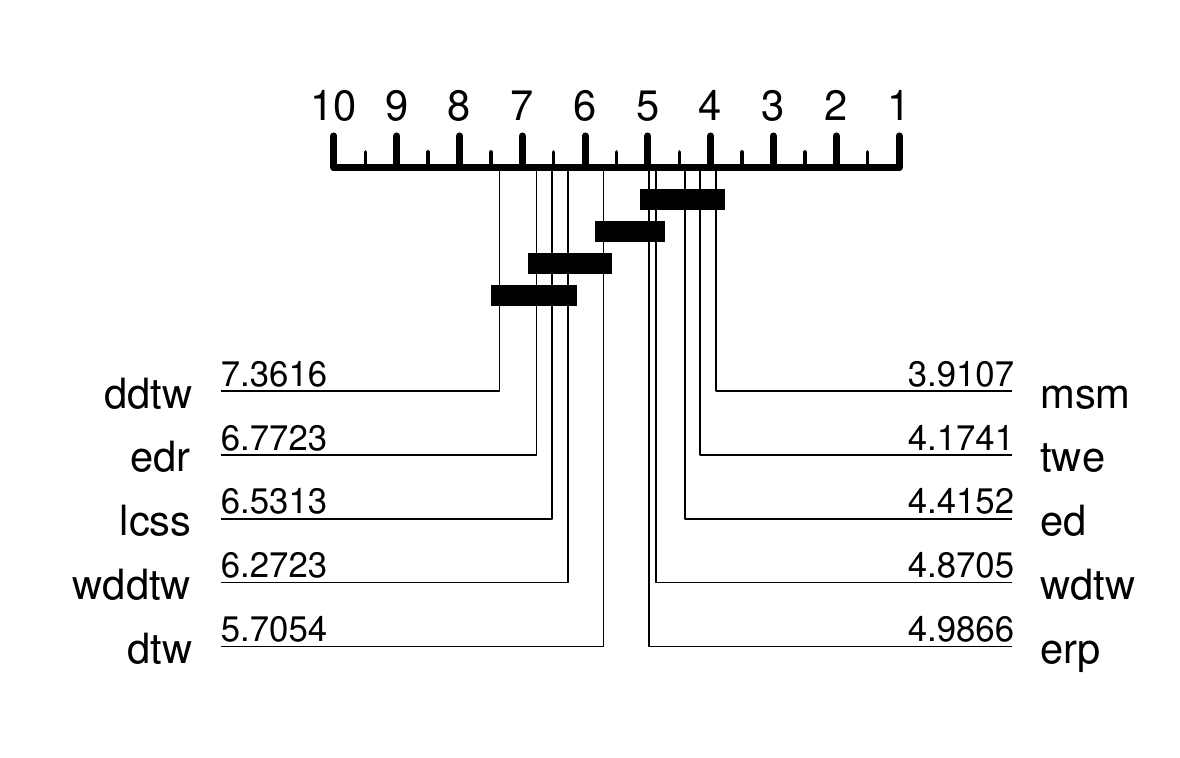}
        \\
        (a) Mutual Information
        & (b) NMIS \\
\end{tabular}
        \caption{Train results for $k$-means clustering with 10 different distance measures.}
        \label{fig:train-kmeans}
    \end{figure}
The two derivative approaches are both significant worse than their alternatives using the raw data, suggesting that clusters in the time domain better reflect the true classes. LCSS and EDR also perform poorly on all tests. This implies the simple edit thresholding is not sensitive enough to find clustering that represent the class labels. The best overall performing measures are MSM and TWE. These measures are similar, in that they combine elements of both warping and editing. Overall, the clear conclusion is that MSM is the best approach for $k$-means clustering with standard averaging to find centroids. WDTW, MSM, TWE and ERP all give an explicit penalty for warping. The DTW penalty for warping is implicit (warping means a longer path), and these results indicate that this allows for more warping than is desirable for clustering.

The obvious first follow up question is: does it make sense to cluster time series with $k$-means and cluster averaging at all? Before directly addressing that question in Section~\ref{sec:kmedoids} and~\ref{sec:dba}, we set the context more thoroughly by considering how much information is in the clusterings in relation to class values.

\begin{figure}[htb]
        \centering
        \begin{tabular}{c c}
        \includegraphics[width=0.5\linewidth,trim={3cm 0cm 3cm 0cm},clip] {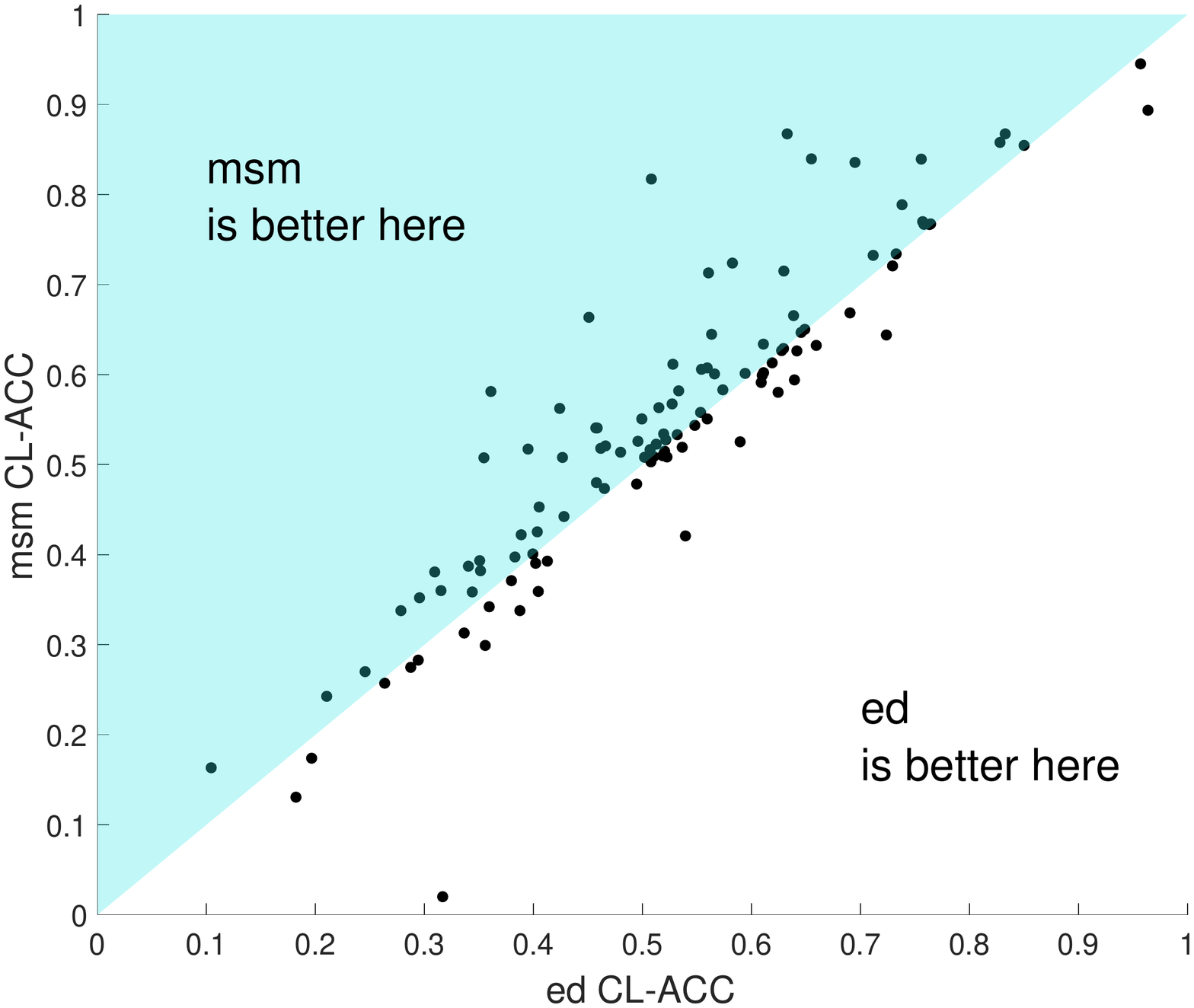} &
        \includegraphics[width=0.5\linewidth,trim={3cm 0cm 3cm 0cm},clip] {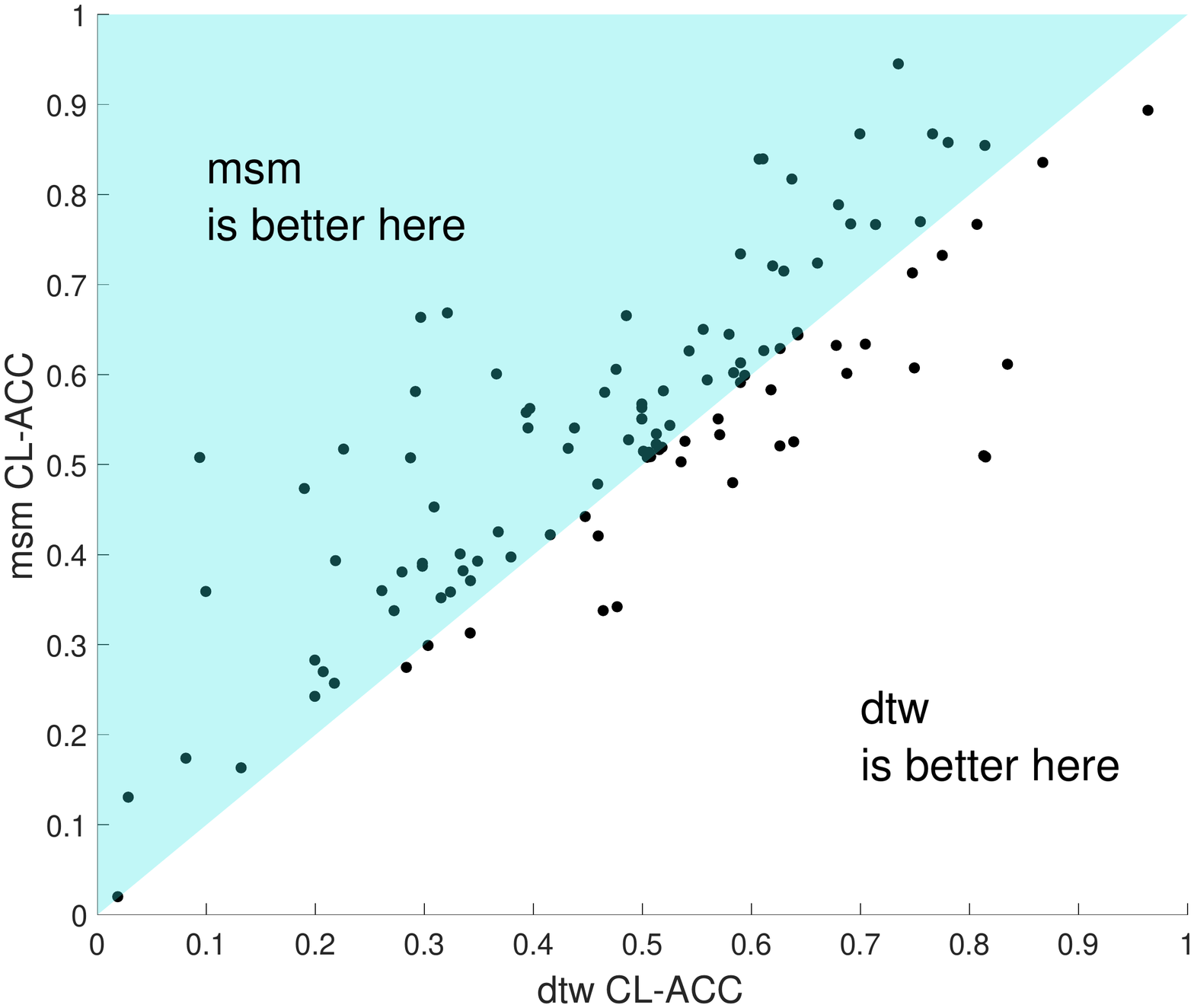}
\end{tabular}
        \caption{Scatter plots of $k$-means MSM against $k$-means Euclidean distance and $k$-means DTW.}
        \label{fig:msm}
    \end{figure}

Although MSM is significantly better than Euclidean distance, it could be that this effect is simply making a bad approach marginally better. Table~\ref{tab:kmeans-full2} presents that the average accuracy over all problems for Euclidean is 51.78\%, DTW 49.08\% and MSM 54.16\%. For reference, predicting just the majority class gives an overall accuracy of 34.26\%. If we take Euclidean distance as our control, then on average DTW is 2.7\% worse, and MSM is 2.38\% better. Clustering clearly provides some insight into class membership. Figure~\ref{fig:msm} shows the accuracy scatter plots of MSM vs ED and DTW, and clearly demonstrates the overall superiority of MSM.
To estimate how much information our clusterers capture, we can upper bound performance with the accuracy of supervised approaches. A 1-NN classifier using Euclidean distance averages 72.36\% on these problems, 1-NN DTW achieves an average accuracy of 76.74\% and the state of the art, HIVE-COTEv2.0~\citep{middlehurst21hc2}, has an average accuracy 89.12\%. This difference is not surprising, an unsupervised method cannot hope to achieve equivalence with a supervised algorithm, but it is worth observing the massive difference in performance. Clustering improves on random guessing on average, but, as discussed in Section~\ref{sec:methods}, many of these data are probably inappropriate for clustering. If we take the highest accuracy of the ten clusterers as our reference and compare accuracy to default class predictions, we find that 18 datasets are less than 5\% better than the predicting a single cluster,
and two (MedicalImages and ChlorineConcentration) are more than 10\% worse than using a single class. Table~\ref{tab:diffs} lists the best performing clusterer, the accuracy obtained from predicting a single cluster and the deviation of the two.

\begin{figure}[htb]
        \centering
        \begin{tabular}{c c}
        \includegraphics[width=0.5\linewidth,trim={4.5cm 4cm 4cm 4cm},clip] {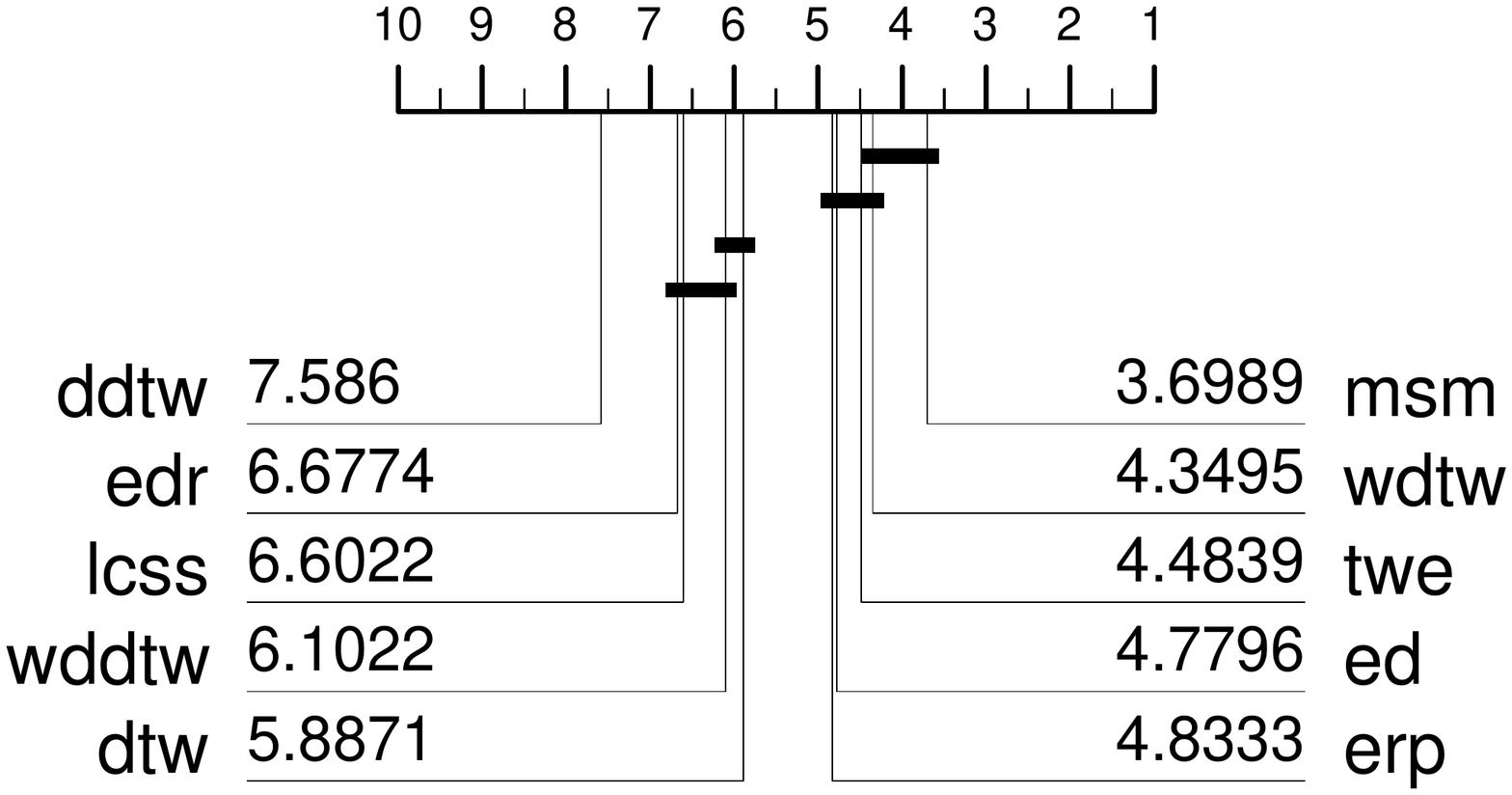} &
        \includegraphics[width=0.5\linewidth,trim={4.5cm 4cm 4cm 4cm},clip] {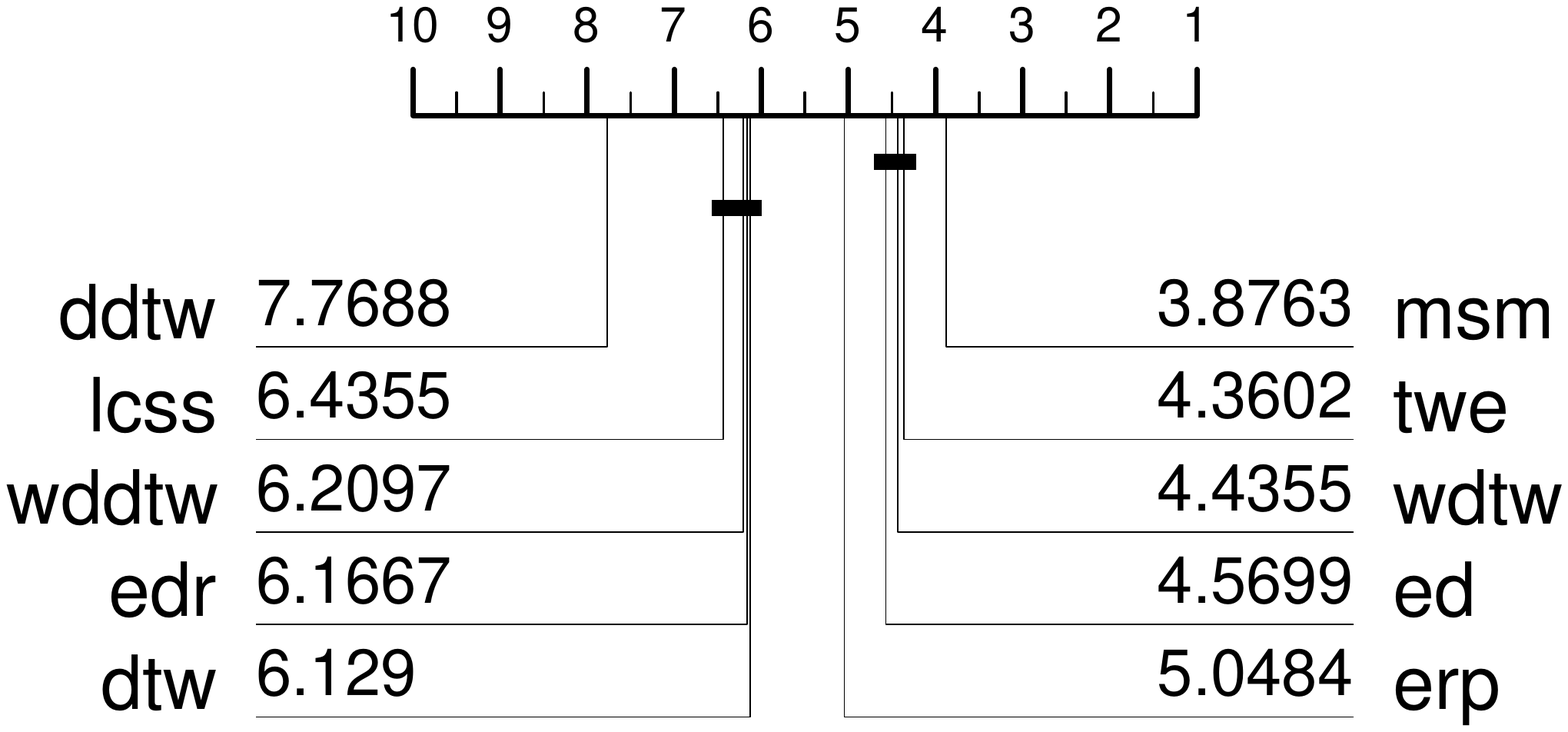}
        \\
        (a) Accuracy
        & (b) Rand Index \\
\end{tabular}
        \caption{Test results for $k$-means clustering with the 93 datasets where the best clusterer is greater than 5\% more accurate than predicting a single cluster.}
        \label{fig:reduced}
    \end{figure}

Figure~\ref{fig:reduced} shows the critical difference diagrams for accuracy and rand index when we rerun the comparison without these 18 datasets. The pattern of results is very similar to those shown in Figure~\ref{fig:test-kmeans}. For completeness sake, we continue with all 112 datasets, but note that it may be worthwhile forming a subset of the UCR problems most suitable for assessing clusterers.
For consistency with some of the related research, Figure~\ref{fig:train-kmeans} show the same results on train data. The pattern of results is the same as on the test data, although unsurprisingly there is less significance in the results.

\subsection{$k$-medoids}
\label{sec:kmedoids}
\begin{figure}[htb]
        \centering
        \begin{tabular}{c c}
        \includegraphics[width=0.5\linewidth,trim={4.5cm 4cm 4cm 4cm},clip] {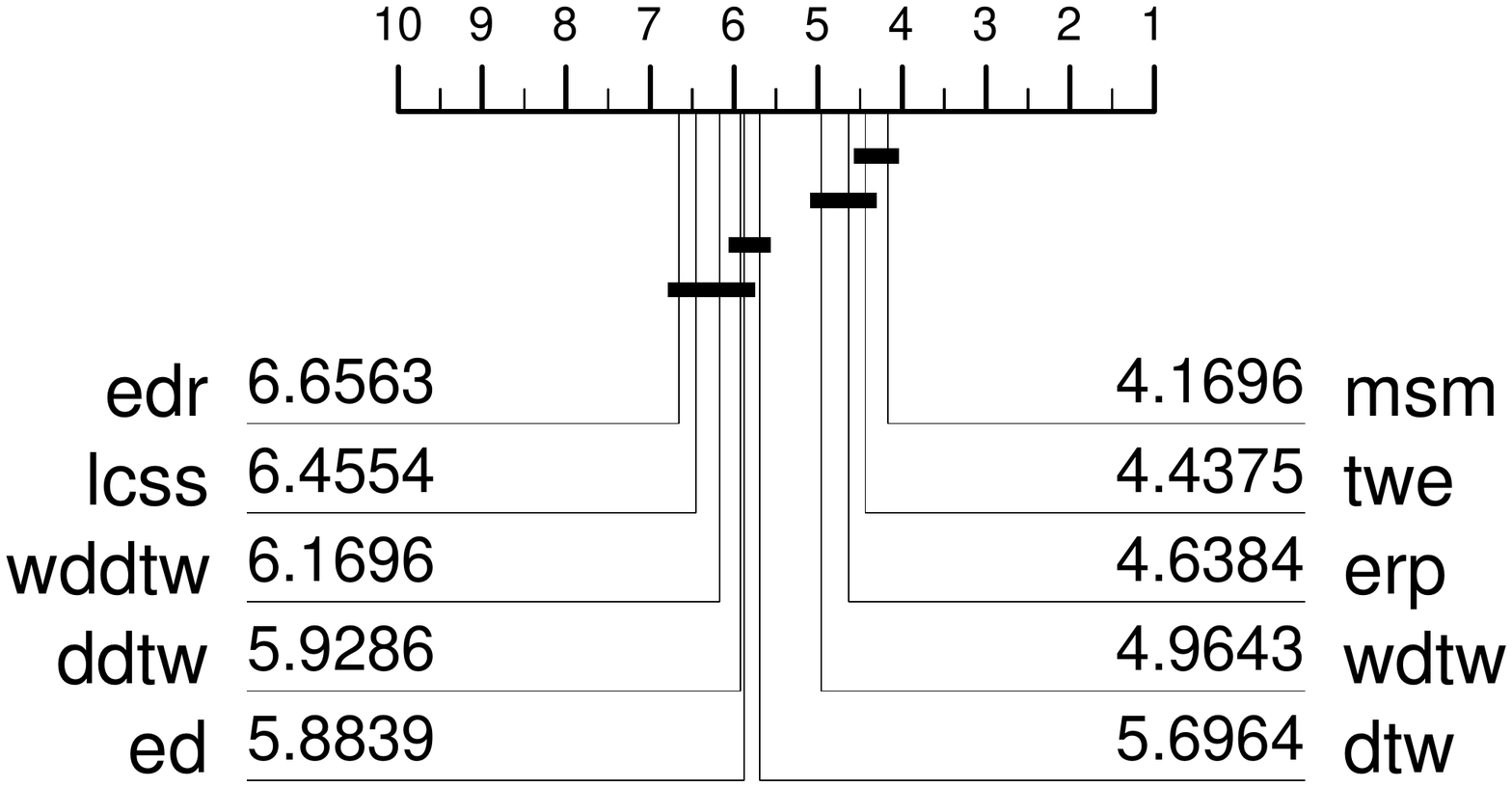} &
        \includegraphics[width=0.5\linewidth,trim={4.5cm 4cm 4cm 4cm},clip] {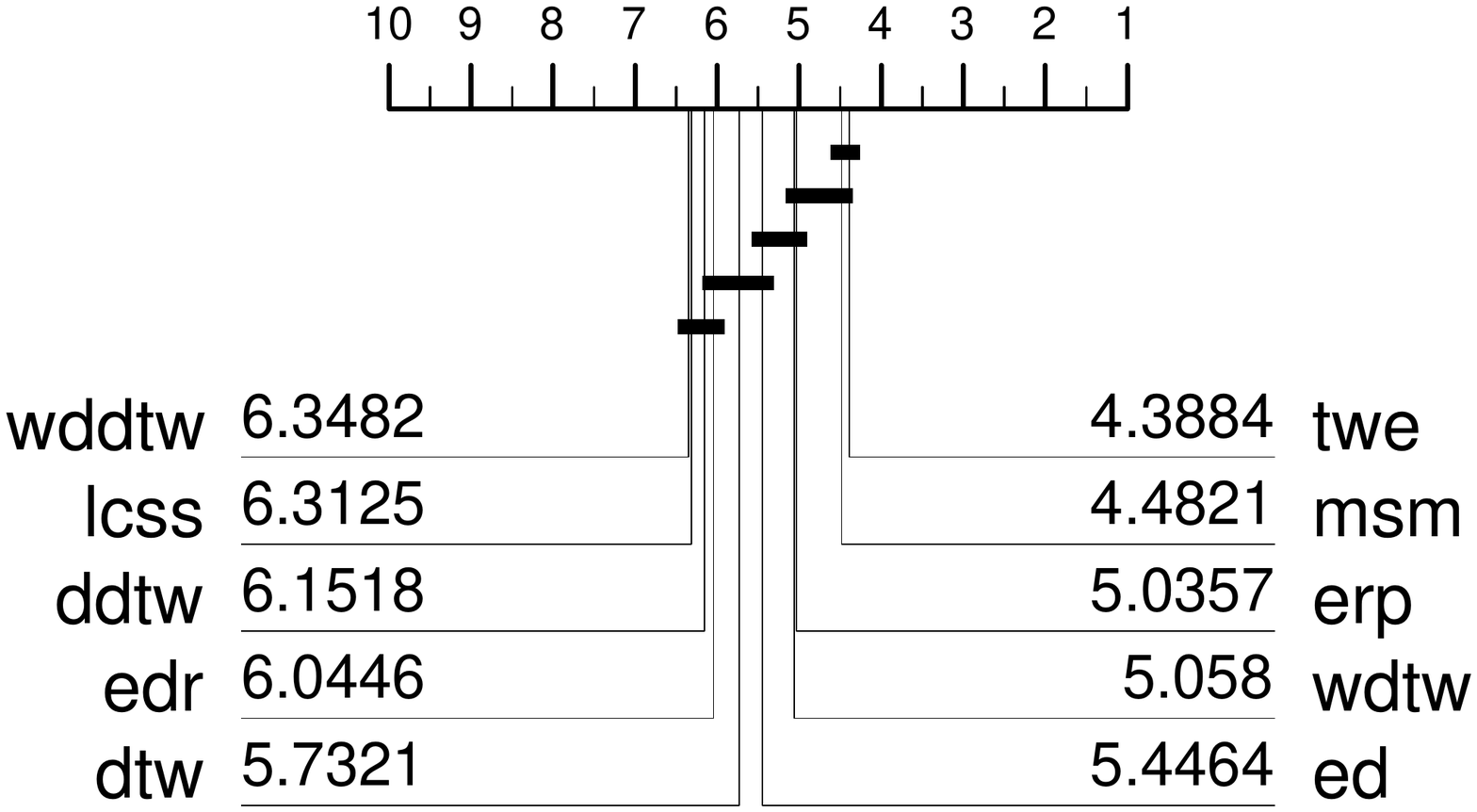}
        \\
        (a) Accuracy
        & (b) Rand Index \\
        \includegraphics[width=0.5\linewidth,trim={4.5cm 4cm 4cm 4cm},clip] {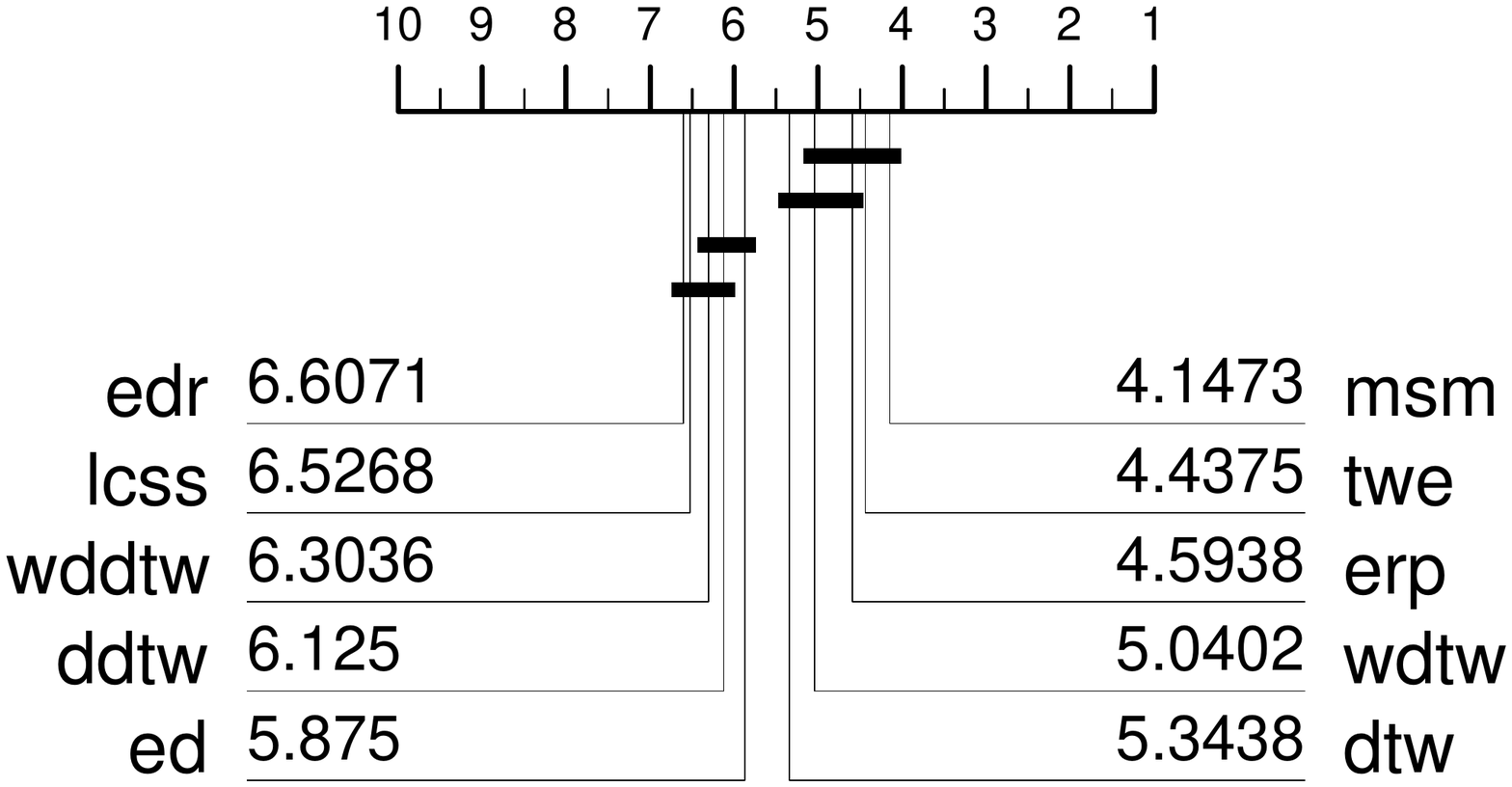} &
        \includegraphics[width=0.5\linewidth,trim={4.5cm 4cm 4cm 4cm},clip] {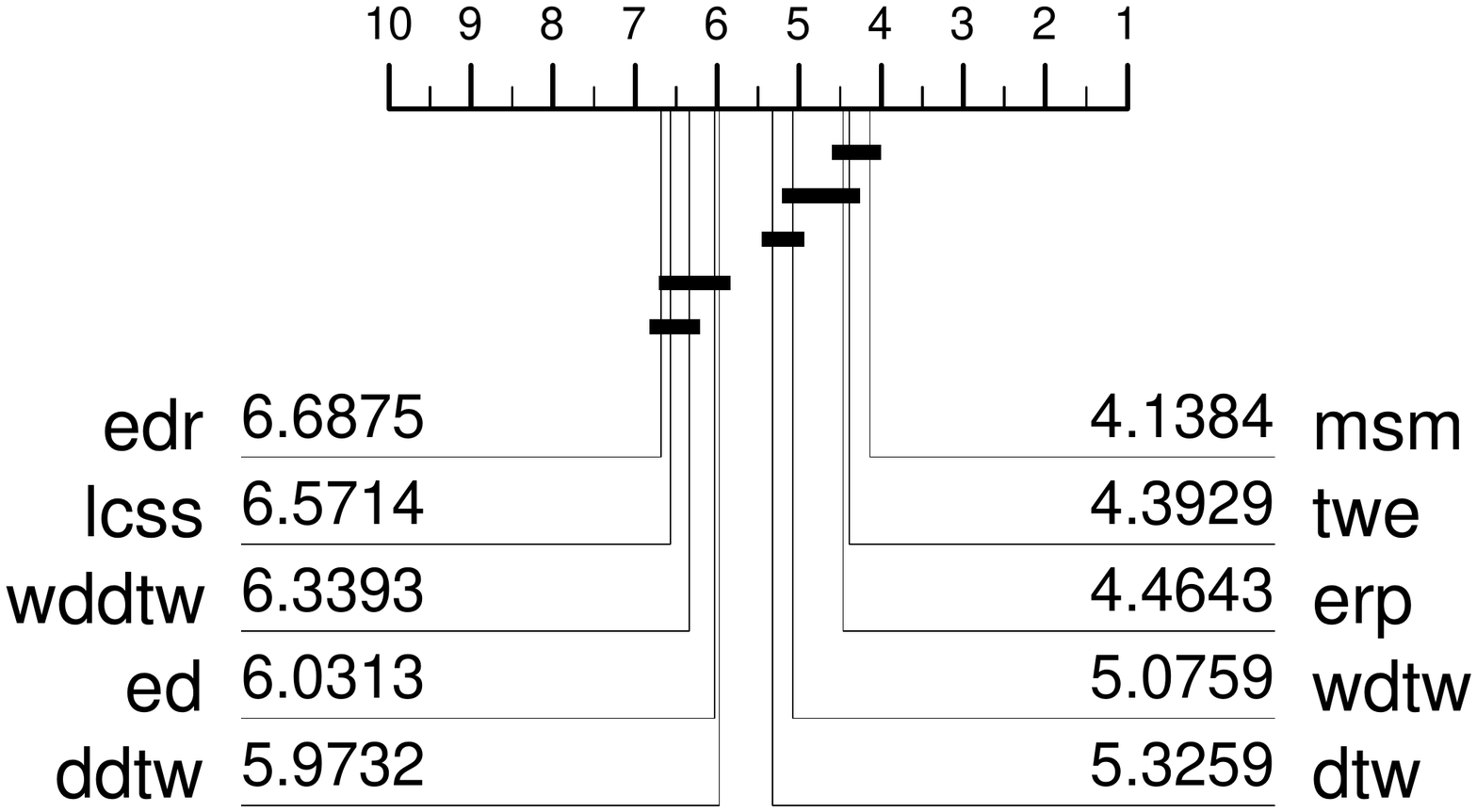}
        \\
        (c) ARIS
        & (d) AMIS \\
        \includegraphics[width=0.5\linewidth,trim={4.5cm 4cm 4cm 4cm},clip] {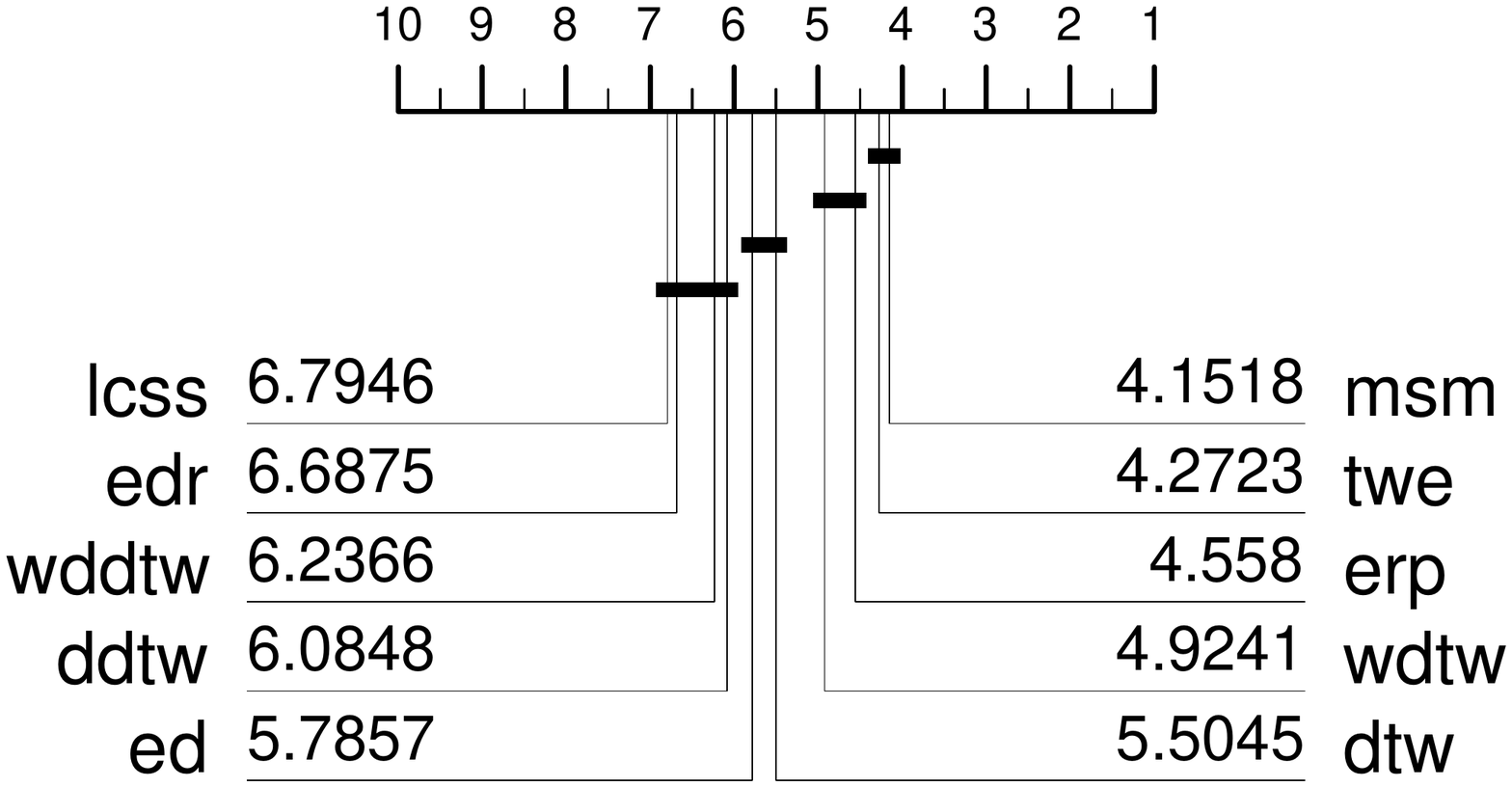} &
        \includegraphics[width=0.5\linewidth,trim={4.5cm 4cm 4cm 4cm},clip]{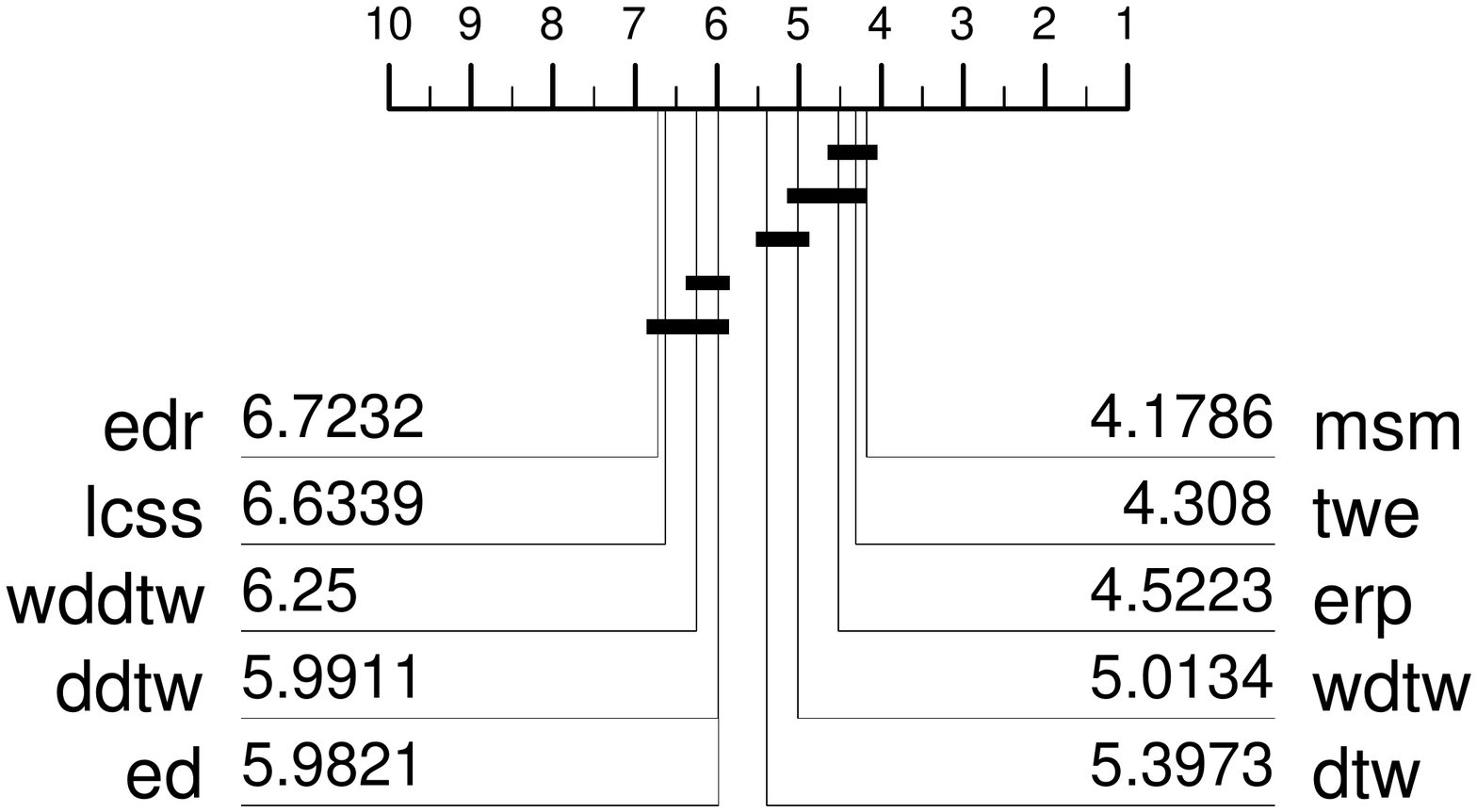}
        \\
        (a) Mutual Information
        & (b) NMIS \\
\end{tabular}
        \caption{Critical difference diagrams for $k$-medoids clustering with ten distance measures assessed on the test data.}
        \label{fig:test-kmedoids}
    \end{figure}

Using standard centroids means that the averaging method is not related to the distance measure used, unless employing Euclidean distance. This disconnect between the clustering stages may account for the poor performance of many of the distance measures, in particular relative to $k$-means with Euclidean distance. We repeated our experiments using the aeon $k$-medoids clusterer. Figure~\ref{fig:test-kmedoids} shows the ranked performance summary for ten distance measures.
The pattern of performance is broadly the same as with $k$-means (Figure~\ref{fig:test-kmeans}) with some notable differences: DTW is now no longer worse than Euclidean; ERP performs much better and there is no overall difference between ERP, TWE and MSM as the top performing algorithms. The top performing distance functions all involve explicit penalty for warping. As with $k$-means, this indicates that regularisation on path length produces better clusters on average.
\begin{table}[ht]
    \centering
        \begin{tabular}{llll}
distance & $k$-means & $k$-medoids & difference \\ \hline
msm  &   54.16\%	& 55.69\%	& 1.54\%   \\
twe  &   52.85\%	& 55.63\%	& 2.78\%   \\
erp  &   50.89\%	& 54.83\%	& 3.94\%   \\
wdtw &   52.25\%	& 53.58\%	& 1.33\%   \\
dtw  &   49.08\%	& 52.96\%	& 3.88\%   \\
ed   &   51.78\%	& 51.40\%	& -0.38\%  \\
ddtw &   42.57\%	& 50.22\%	& 7.65\%   \\
wddtw&   46.99\%	& 49.55\%	& 2.57\%   \\
lcss &   45.76\%	& 49.88\%	& 4.13\%   \\
edr  &   45.20\%	& 49.70\%	& 4.50\%   \\
    \end{tabular}
    \caption{Accuracy averaged over 112 problems for $k$-means and $k$-medoids clustering.}
    \label{tab:av_acc}
\end{table}
Also of interest is the relative performance between $k$-means and $k$-medoids. Table~\ref{tab:av_acc} lists the average accuracy over all problems of $k$-means, $k$-medoids and single cluster predictions for the ten distance measures. Accuracy increases for all distances except Euclidean. This indicates that $k$-medoids is a much stronger benchmark for TSCL algorithms than $k$-means.

\subsection{kmeans-DBA}
\label{sec:dba}
\begin{figure}[htb]
        \centering
        \begin{tabular}{c c}
        \includegraphics[width=0.5\linewidth,trim={3cm 0cm 3cm 0cm},clip] {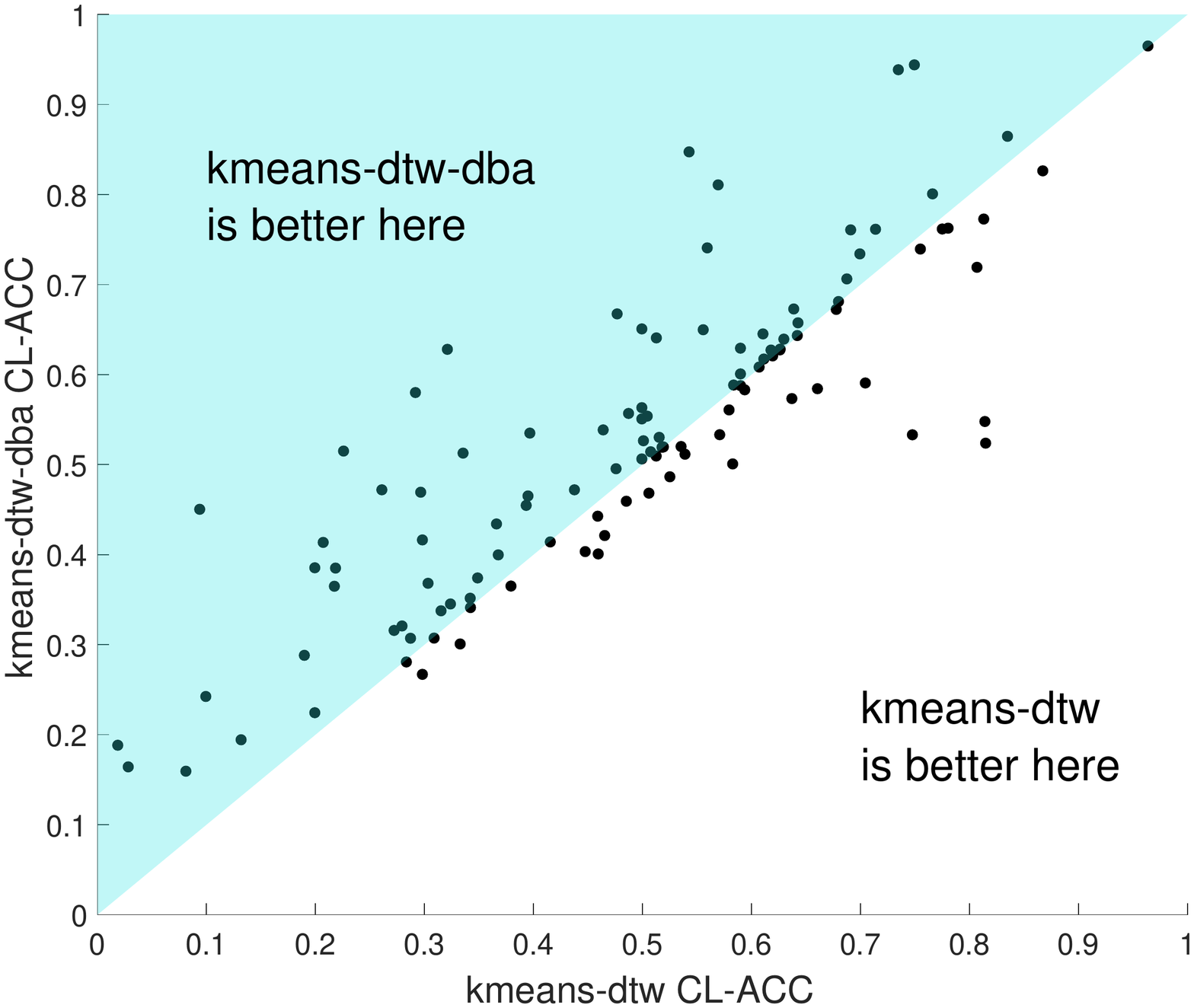} &
        \includegraphics[width=0.5\linewidth,trim={3cm 0cm 3cm 0cm},clip] {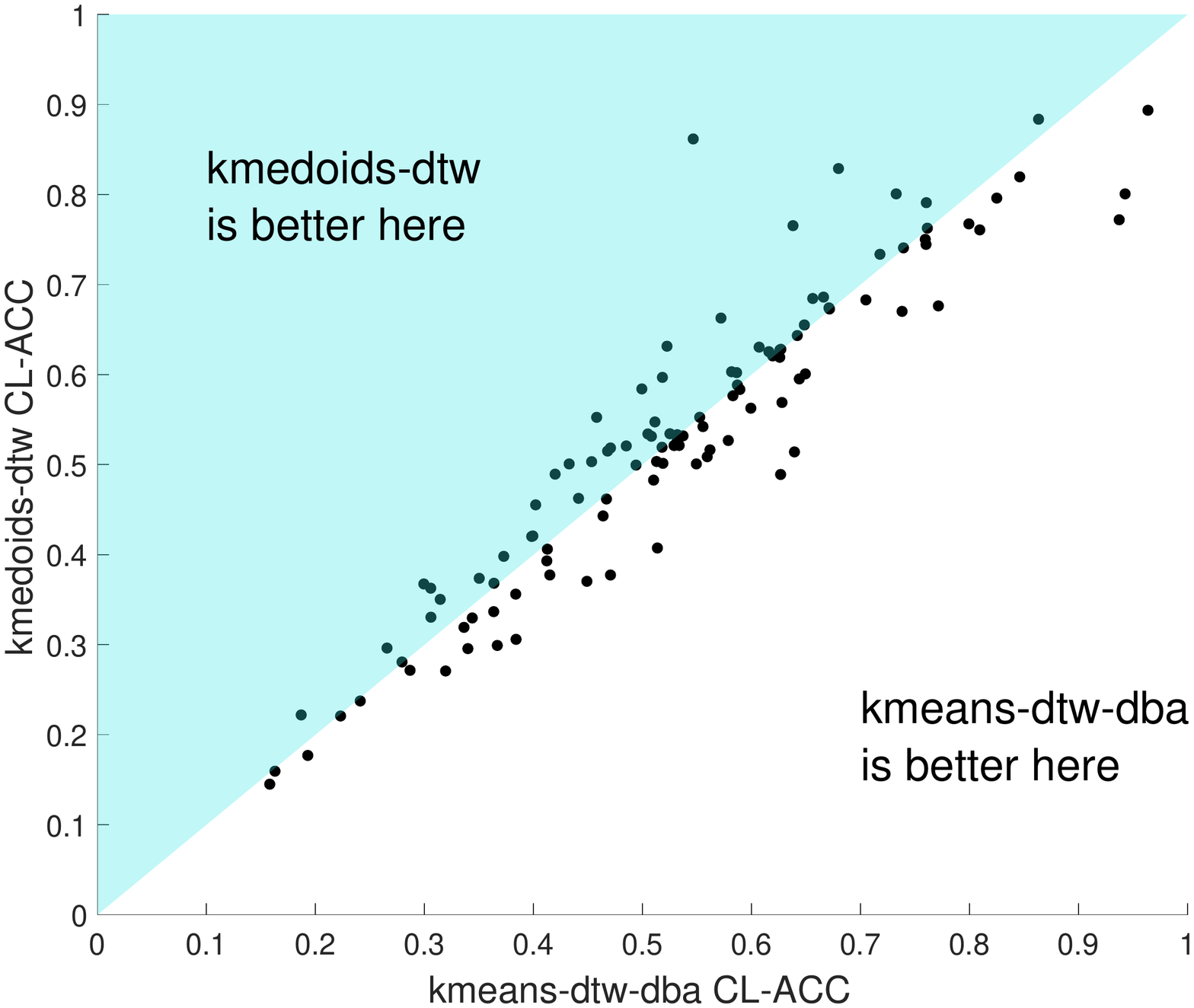}\\
        (a) DBA wins 68 and loses 36 (8 ties) & (b) DBA wins 57 and loses 44 (11 ties)\\
\end{tabular}
        \caption{Scatter plots of accuracy forDTW based $k$-means, with DBA against standard averaging (a) and against $k$-medoids (b).}
        \label{fig:dba-scatter}
    \end{figure}

Using medoids mitigates the problem of averaging centres with $k$-means. DBA, described in Section~\ref{sec:averaging}, has also been proposed as a means of improving centroid finding for DTW. We have repeated our experiments with the same $k$-means set up, but centroids found with the original DBA described in Algorithm~\ref{algo:ba}. Figure~\ref{fig:dba-scatter}(a) shows that DBA does indeed significantly improve DTW, a result that reproduces the findings in~\citep{petitjean11dba}.

\begin{figure}[htb]
        \centering
        \begin{tabular}{c c}
        \includegraphics[width=0.5\linewidth,trim={1cm 4cm 0cm 4cm},clip] {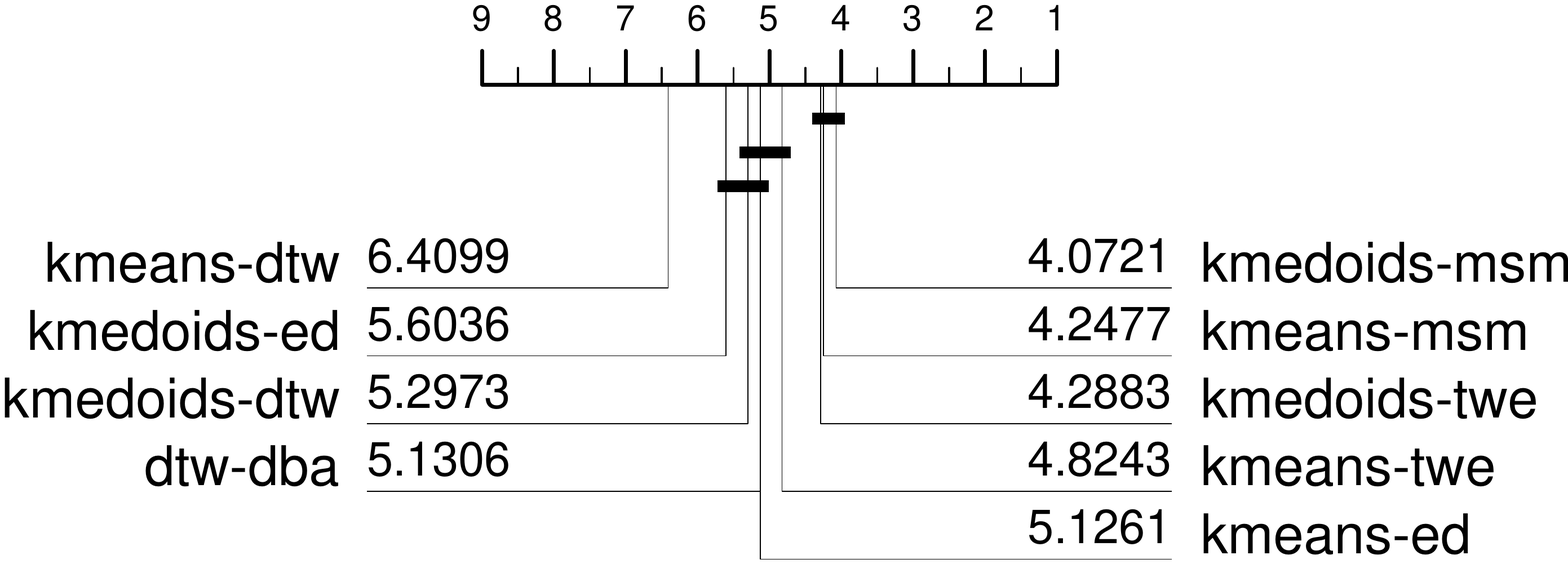} &
        \includegraphics[width=0.5\linewidth,trim={1cm 4cm 0cm 4cm},clip] {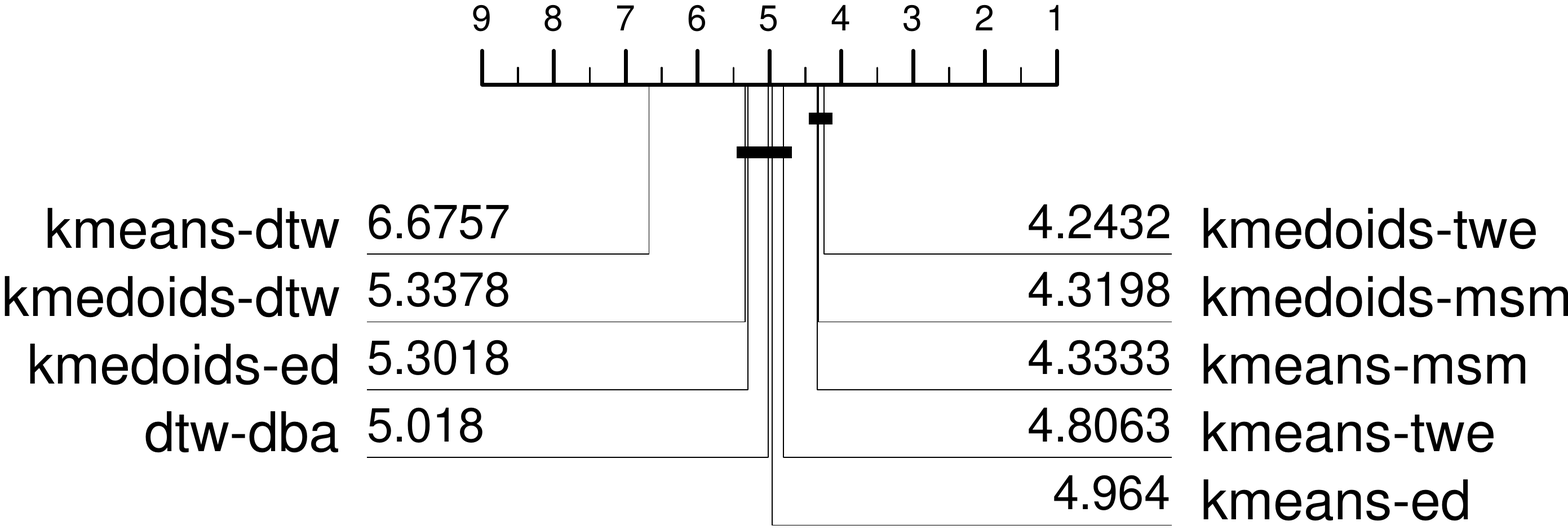}
        \\
        (a) Accuracy
        & (b) Rand Index \\
        \includegraphics[width=0.5\linewidth,trim={1cm 4cm 0cm 4cm},clip] {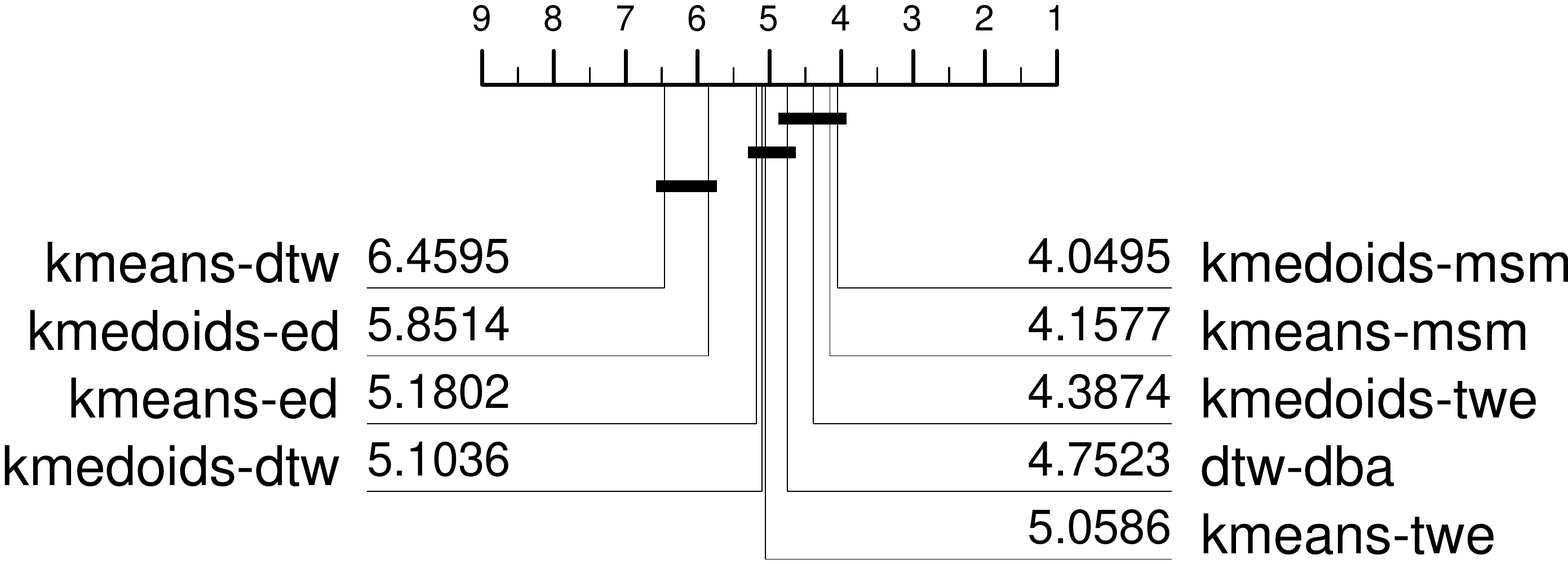} &
        \includegraphics[width=0.5\linewidth,trim={1cm 4cm 0cm 4cm},clip] {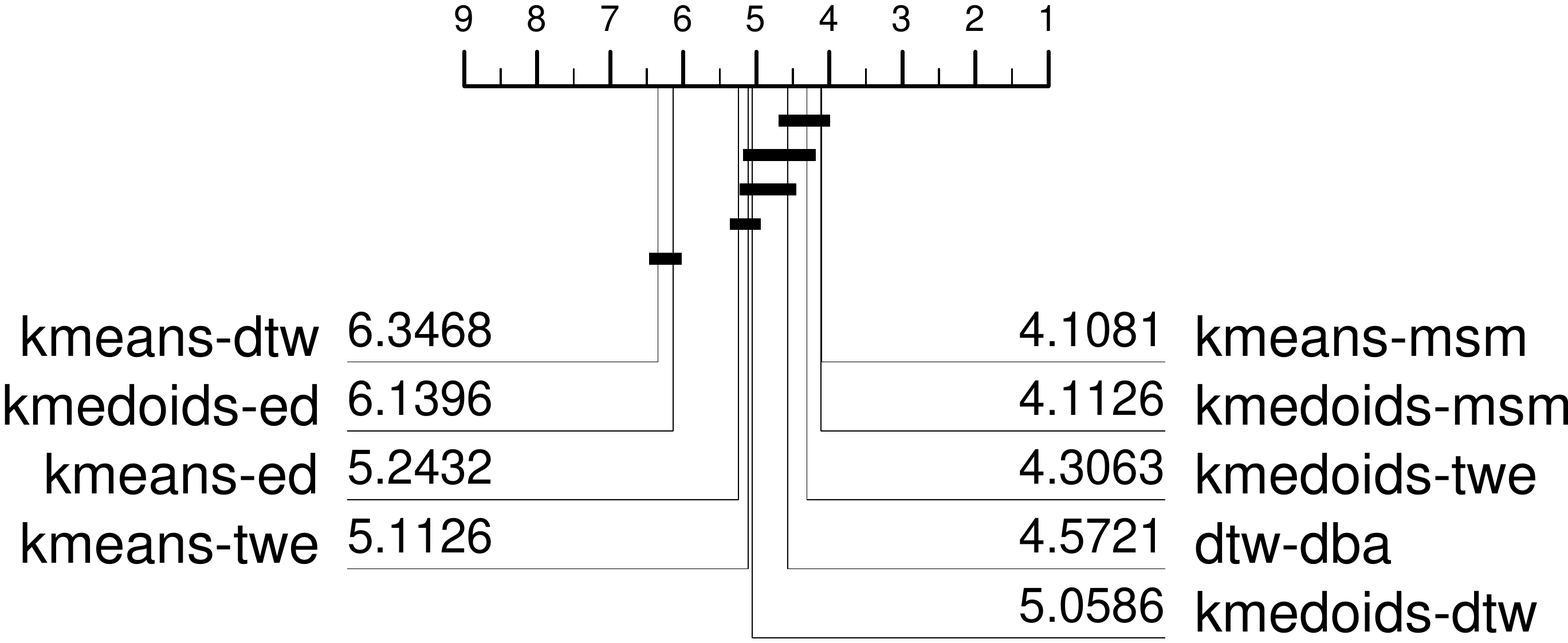}
        \\
        (c) ARIS
        & (d) AMIS \\
        \includegraphics[width=0.5\linewidth,trim={1cm 4cm 0cm 4cm},clip] {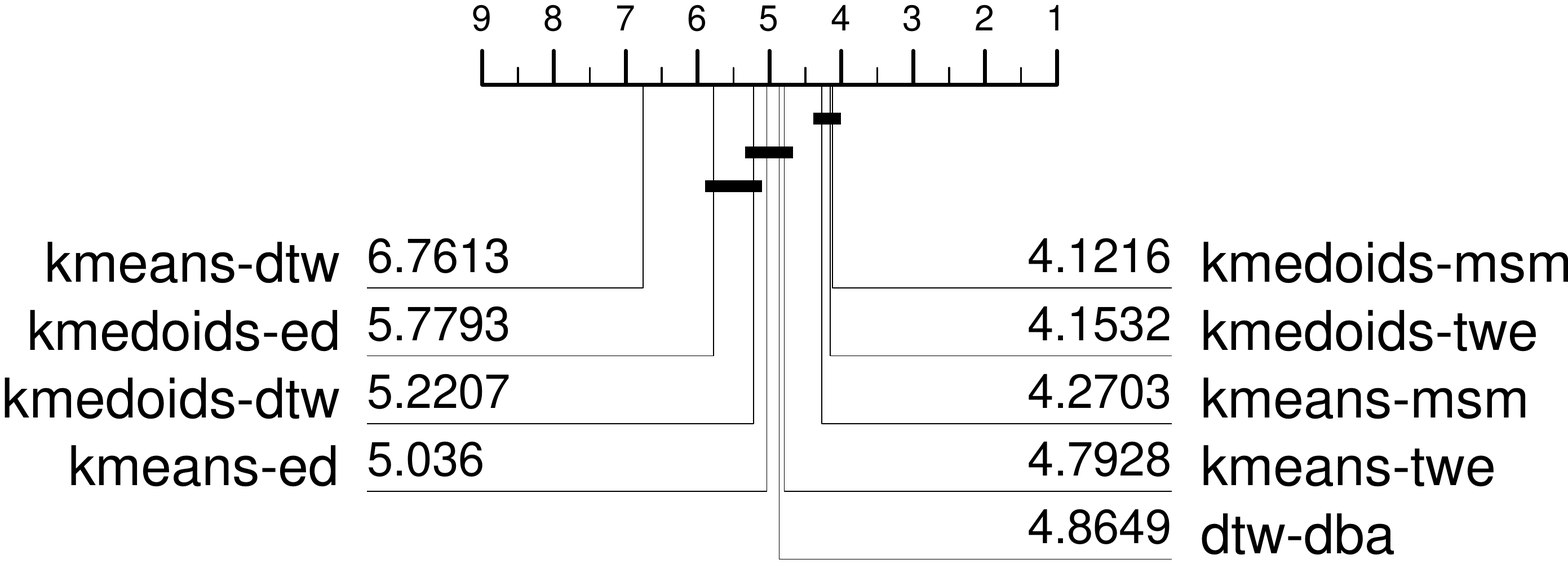} &
        \includegraphics[width=0.5\linewidth,trim={1cm 4cm 0cm 4cm},clip]{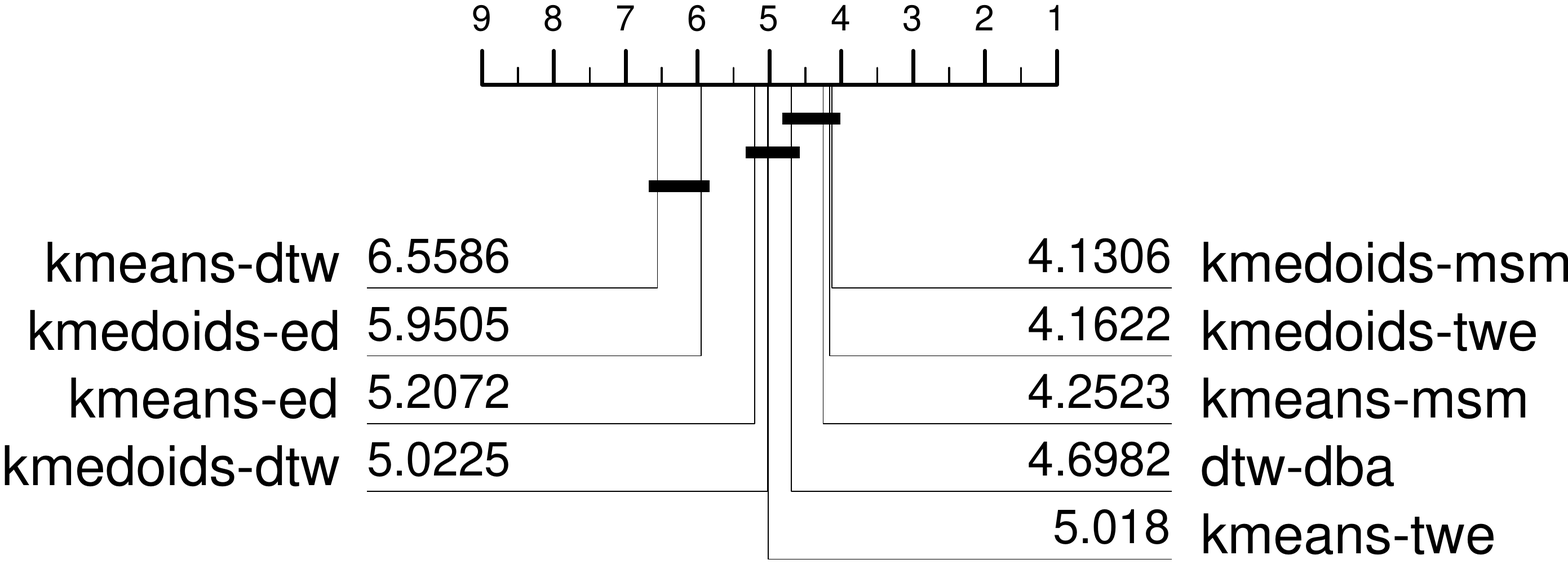}
        \\
        (a) Mutual Information
        & (b) NMIS \\
\end{tabular}
        \caption{Critical difference diagrams for the best performing $k$-means and $k$-medoids distances, and $k$-means with DBA (dtw-dba).}
        \label{fig:dbs_cd}
    \end{figure}

However, Figure~\ref{fig:dba-scatter}(a) illustrates that it is not significantly different to $k$-medoids DTW. Furthemore, Figure~\ref{fig:dbs_cd} shows that $k$-means with DBA is significantly worse than the top clique, composed of $k$-medoids MSM, $k$-means MSM and $k$-medoids TWE.

Our implementation of DBA is faithful to the original. An alternative version of DBA was described in~\citep{schultz17subgradientdba}. This version is implemented in the tslearn toolkit. We have wrapped the tslearn $k$-means DBA algorithm into the aeon toolkit and reran this version.

\begin{figure}[htb]
        \centering
        \begin{tabular}{c c}
        \includegraphics[width=0.5\linewidth,trim={1cm 2cm 1cm 2cm},clip] {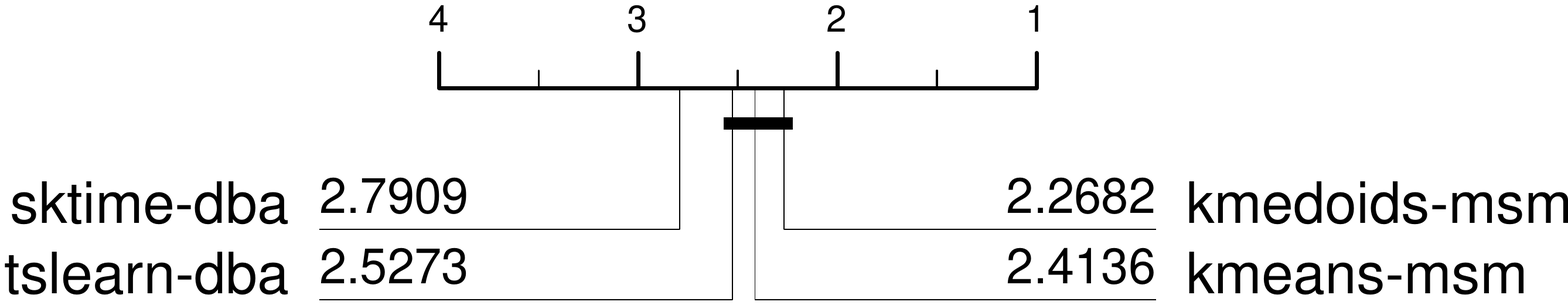} &
        \includegraphics[width=0.5\linewidth,trim={1cm 2cm 1cm 2cm},clip] {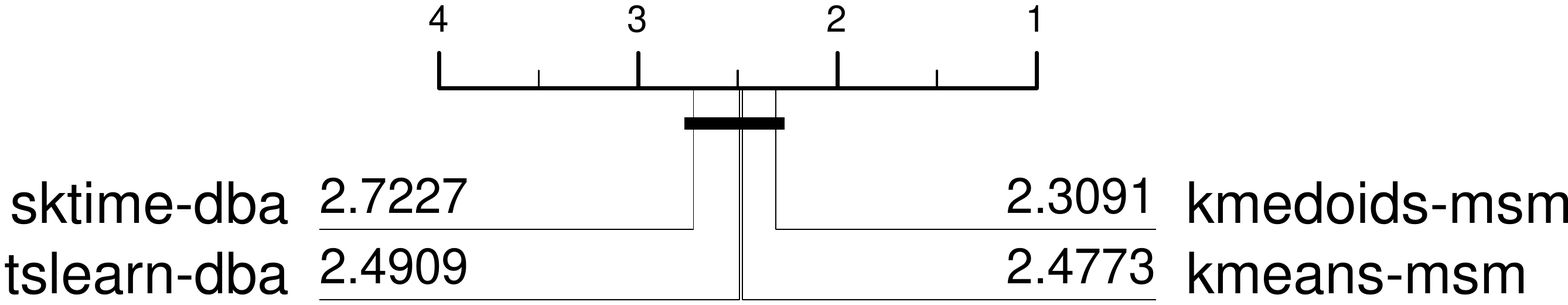}\\
        (a) Accuracy & (b) Average Rand Index\\
\end{tabular}
        \caption{Comparison of performance of two DBA implementations and $k$-medoids and $k$-means with MSM distance.}
        \label{fig:dba-tslearn}
    \end{figure}

Figure~\ref{fig:dba-tslearn} compares the performance of two DBA versions with MSM $k$-medoids and $k$-means.

Table~\ref{tab:timings} summarises the time taken to run experiments. We ran experiments on our HPC cluster, so timing results are indicative only. The differences are fairly small. However, we note that of the two best performing algorithms, $k$-medoids MSM and TWE, $k$-medoids MSM is the faster and hence is our recommended benchmark for elastic distance time series clustering.
\begin{table}[htb]
\label{tab:timings}
\centering
\caption{Run time (in hours) average, maximum and total over 112 problems.}
\begin{tabular}{l|lll}
Distance & Mean  & Max     & Total      \\ \hline
ed          & 0.003 & 0.046   & 0.341    \\
dtw         & 3.272 & 53.579  & 366.514  \\
wdtw        & 4.811 & 95.916  & 538.815  \\
msm         & 4.536 & 70.406  & 508.059  \\
edr         & 5.044 & 81.628  & 564.926  \\
lcss        & 5.063 & 82.803  & 567.047  \\
ddtw        & 5.393 & 85.351  & 603.973  \\
erp         & 5.480 & 90.075  & 613.765  \\
wddtw       & 5.196 & 86.864  & 581.994  \\
twe         & 6.417 & 117.22 & 718.650  \\
aeon-dba  & 8.112 & 116.62  & 900.514  \\
tslearn-dba & 9.724 & 159.79  & 1080.43 \\
\end{tabular}
\end{table}
\section{Conclusions}
\label{sec:conclusions}
We have described 10 time series distance measures and compared them when used to cluster the time series of the UCR archive. There are a wealth of other time series clustering algorithms that we have not evaluated. We have opted to provide an in depth description, with examples and associated code and a tightly constrained bake off, rather than attempt to include all variants of, for example, deep learning and transformation based time series clusterers. Distance based approaches are still very popular, and we believe our experimental observations will help practitioners choose distance based TSCL more effectively. Our first conclusion is that $k$-means using DTW and centroid averaging is not a useful benchmark for new clustering algorithms. For $k$-means, MSM is the best distance measure to use. However, $k$-means with averaging of centroids is itself not an effective approach for TSCL. $k$-medoids is significantly better for all nine elastic distances, and has the added benefit of being able to use a precomputed distance matrix. A popular solution to the problem  of averaging centroids for $k$-means is to use DBA, which uses the warping mechanism to align centroids. Whilst this does indeed improve DTW $k$-means, it is computationally intensive and does not improve DTW enough to make it competitive with the best $k$-medoids approach. The improved version of DBA~\citep{schultz17subgradientdba}  implemented in tslearn is significantly better than the original, but is not different to either MSM clusterers, and it takes approximately twice as long to run.

Our overall conclusion is that, without any data specific prior knowledge as to the best approach, $k$-medoids with either MSM or TWE are the best approaches for distance based clustering of time series, with MSM preferred because of the lower run time. They should form the basis for assessing new clustering algorithms (although the tslearn version of DBA is also a good benchmark). There is little difference in run time between the algorithms, $k$-medoids is a little faster but uses more memory, and TWE is slightly slower than the other distances. However, run time and memory were not a major constraint for these experiments.

We have released all our results and code in a dedicated repository. We would welcome contributors of new distance functions to aeon, and would be happy to extend the evaluation to include them. We encourage any would be contributors to engage with the aeon community and look for or raise issues on the aeon GitHub repository related to the distances module. Our next stage is to extend the bake off to consider clustering algorithms not based on elastic distances, to assess the impact of alternative clustering algorithms parameters (e.g. the initialisation technique) and to investigate alternative medoids based approaches such as partitioning around the medoids~\citep{kaufman90pam}. Furthermore, we have not yet investigated the effect of having to set the number of clusters, $k$. A future experiment will involve testing whether the relative performance remains the same when using standard techniques for setting $k$. There are also several possible directions for algorithmic advancement highlighted by this research: we will try combining the clusterings through an ensemble in a manner similar to the elastic ensemble for classification~\citep{lines15elastic}. Clustering ensembles are more complex than classification ensembles, requiring some form of alignment of labelling. Nevertheless, it seems reasonably likely that there may be some improvements from doing so. These are just some of the numerous open issues in TSCL research. Our study aims to put future research on a sound basis to facilitate the accurate assessment of algorithmic improvements in a fully reproducible manner.

\section*{Acknowledgements}{
     This work is supported by the UK Engineering and Physical Sciences Research Council (EPSRC) iCASE award sponsored by British Telecom. The experiments were carried out on the High Performance Computing Cluster supported by the Research and Specialist Computing Support service at the University of East Anglia.
}

\bibliographystyle{spbasic}

\newpage
\appendix{Results Tables}

\begin{table}[htb]
\caption{Test set accuracy for $k$-means and 10 different distance measures, part 1}
\begin{tabular}{lllllllllll}
TESTCL-ACC                     & msm    & wdtw   & twe    & erp    & ed     & dtw    & wddtw  & edr    & lcss   & ddtw   \\
ACSF1                          & 0.3800 & 0.3800 & 0.4100 & 0.3700 & 0.3100 & 0.2800 & 0.2700 & 0.3500 & 0.3300 & 0.2400 \\
Adiac                          & 0.4246 & 0.3760 & 0.3760 & 0.3990 & 0.4041 & 0.3683 & 0.3223 & 0.3402 & 0.3529 & 0.2020 \\
ArrowHead                      & 0.5429 & 0.5029 & 0.5600 & 0.5543 & 0.5486 & 0.5257 & 0.3943 & 0.5486 & 0.6000 & 0.3943 \\
Beef                           & 0.4000 & 0.3333 & 0.3667 & 0.3000 & 0.4000 & 0.3333 & 0.4000 & 0.4333 & 0.4667 & 0.4000 \\
BeetleFly                      & 0.5500 & 0.5500 & 0.6000 & 0.7000 & 0.5000 & 0.5000 & 0.5000 & 0.5000 & 0.5000 & 0.5000 \\
BirdChicken                    & 0.5500 & 0.6000 & 0.5000 & 0.5500 & 0.5000 & 0.5000 & 0.5500 & 0.5000 & 0.5000 & 0.5000 \\
BME                            & 0.5200 & 0.6400 & 0.5067 & 0.5000 & 0.4667 & 0.6267 & 0.8867 & 0.5733 & 0.5733 & 0.3333 \\
Car                            & 0.6000 & 0.5667 & 0.5500 & 0.6167 & 0.5667 & 0.3667 & 0.5000 & 0.5667 & 0.6000 & 0.3167 \\
CBF                            & 0.6256 & 0.6933 & 0.6267 & 0.5811 & 0.6422 & 0.5433 & 0.3356 & 0.4489 & 0.7189 & 0.3356 \\
Chinatown                      & 0.6006 & 0.6880 & 0.5889 & 0.6093 & 0.5948 & 0.6880 & 0.8892 & 0.7959 & 0.8017 & 0.8892 \\
ChlorineConcentration          & 0.3966 & 0.3781 & 0.3969 & 0.3911 & 0.3836 & 0.3799 & 0.3701 & 0.3841 & 0.4211 & 0.3698 \\
CinCECGTorso                   & 0.4522 & 0.4623 & 0.4428 & 0.3130 & 0.4058 & 0.3094 & 0.6601 & 0.3167 & 0.3109 & 0.2536 \\
Coffee                         & 0.8929 & 0.9286 & 0.8929 & 0.9643 & 0.9643 & 0.9643 & 0.9643 & 0.9286 & 0.5714 & 0.8929 \\
Computers                      & 0.6440 & 0.5680 & 0.6000 & 0.6160 & 0.5640 & 0.5800 & 0.6120 & 0.5880 & 0.5920 & 0.6400 \\
CricketX                       & 0.2692 & 0.3077 & 0.2590 & 0.1821 & 0.2462 & 0.2077 & 0.2000 & 0.1821 & 0.1872 & 0.1026 \\
CricketY                       & 0.2821 & 0.3128 & 0.3026 & 0.2538 & 0.2949 & 0.2000 & 0.1897 & 0.2000 & 0.2205 & 0.0949 \\
CricketZ                       & 0.2564 & 0.2949 & 0.2436 & 0.1974 & 0.2641 & 0.2179 & 0.1974 & 0.1769 & 0.1974 & 0.1026 \\
Crop                           & 0.3703 & 0.3482 & 0.3783 & 0.3010 & 0.3804 & 0.3428 & 0.2546 & 0.0776 & 0.1203 & 0.2214 \\
DiatomSizeReduction            & 0.9444 & 0.7320 & 0.8693 & 0.8105 & 0.9575 & 0.7353 & 0.6961 & 0.7255 & 0.7157 & 0.6993 \\
DistalPhalanxOutlineAgeGroup   & 0.6331 & 0.7050 & 0.6187 & 0.6906 & 0.6115 & 0.7050 & 0.6547 & 0.5324 & 0.5612 & 0.6547 \\
DistalPhalanxOutlineCorrect    & 0.6123 & 0.5906 & 0.6123 & 0.5906 & 0.6196 & 0.5906 & 0.5870 & 0.5688 & 0.5616 & 0.5833 \\
DistalPhalanxTW                & 0.5252 & 0.5252 & 0.4604 & 0.5252 & 0.4964 & 0.5396 & 0.6259 & 0.4245 & 0.4029 & 0.4964 \\
Earthquakes                    & 0.7122 & 0.7482 & 0.5827 & 0.7482 & 0.5612 & 0.7482 & 0.7482 & 0.6115 & 0.6115 & 0.7482 \\
ECG200                         & 0.7200 & 0.6200 & 0.7300 & 0.6900 & 0.7300 & 0.6200 & 0.7500 & 0.7000 & 0.6600 & 0.7600 \\
ECG5000                        & 0.7231 & 0.6593 & 0.7109 & 0.7282 & 0.5831 & 0.6611 & 0.6151 & 0.7233 & 0.7496 & 0.5947 \\
ECGFiveDays                    & 0.8165 & 0.6144 & 0.5819 & 0.6899 & 0.5087 & 0.6376 & 0.5029 & 0.6144 & 0.5041 & 0.5029 \\
ElectricDevices                & 0.3371 & 0.3977 & 0.2996 & 0.4299 & 0.3881 & 0.4645 & 0.2805 & 0.3240 & 0.3944 & 0.2646 \\
EOGHorizontalSignal            & 0.2983 & 0.3591 & 0.3978 & 0.3039 & 0.3564 & 0.3039 & 0.2265 & 0.2072 & 0.2541 & 0.0939 \\
EOGVerticalSignal              & 0.3122 & 0.3895 & 0.3564 & 0.2956 & 0.3370 & 0.3425 & 0.2624 & 0.2017 & 0.2017 & 0.0856 \\
EthanolLevel                   & 0.2740 & 0.2940 & 0.2880 & 0.2720 & 0.2880 & 0.2840 & 0.2840 & 0.2920 & 0.2800 & 0.2680 \\
FaceAll                        & 0.5805 & 0.3444 & 0.4846 & 0.1799 & 0.3615 & 0.2923 & 0.2580 & 0.1751 & 0.1728 & 0.2213 \\
FaceFour                       & 0.5795 & 0.6477 & 0.7386 & 0.6932 & 0.6250 & 0.4659 & 0.4773 & 0.5114 & 0.4659 & 0.4545 \\
FacesUCR                       & 0.5068 & 0.2454 & 0.4888 & 0.2210 & 0.3551 & 0.2878 & 0.1820 & 0.1439 & 0.1995 & 0.1639 \\
FiftyWords                     & 0.5165 & 0.3143 & 0.4440 & 0.1516 & 0.3956 & 0.2264 & 0.2549 & 0.1714 & 0.1319 & 0.1648 \\
Fish                           & 0.6629 & 0.4514 & 0.4343 & 0.5029 & 0.4514 & 0.2971 & 0.2971 & 0.4457 & 0.4114 & 0.1657 \\
FordA                          & 0.5159 & 0.5159 & 0.5129 & 0.5159 & 0.5076 & 0.5159 & 0.5159 & 0.5159 & 0.5159 & 0.5159 \\
FordB                          & 0.5074 & 0.5049 & 0.5185 & 0.5049 & 0.5025 & 0.5049 & 0.5395 & 0.5049 & 0.5049 & 0.5049 \\
FreezerRegularTrain            & 0.7667 & 0.7667 & 0.7667 & 0.7656 & 0.7649 & 0.6916 & 0.6961 & 0.7319 & 0.7372 & 0.6418 \\
FreezerSmallTrain              & 0.7660 & 0.7526 & 0.7660 & 0.7649 & 0.7639 & 0.7144 & 0.6842 & 0.7407 & 0.7456 & 0.6768 \\
GunPoint                       & 0.5333 & 0.5133 & 0.5200 & 0.5267 & 0.5200 & 0.5133 & 0.5333 & 0.6000 & 0.5733 & 0.6667 \\
GunPointAgeSpan                & 0.8386 & 0.5949 & 0.7532 & 0.5190 & 0.7563 & 0.6076 & 0.5538 & 0.5918 & 0.6108 & 0.6361 \\
GunPointMaleVersusFemale       & 0.6108 & 0.6994 & 0.5158 & 0.5285 & 0.5285 & 0.8354 & 0.7373 & 0.6835 & 0.7975 & 0.6424 \\
GunPointOldVersusYoung         & 0.7333 & 0.5365 & 0.7270 & 0.7429 & 0.7333 & 0.5905 & 0.6413 & 0.5048 & 0.6317 & 0.6032 \\
Ham                            & 0.5905 & 0.6095 & 0.6000 & 0.6095 & 0.6095 & 0.5905 & 0.5143 & 0.6190 & 0.6095 & 0.5143 \\
HandOutlines                   & 0.6432 & 0.6649 & 0.7216 & 0.6514 & 0.7243 & 0.6432 & 0.6405 & 0.6459 & 0.6459 & 0.6405 \\
\end{tabular}
\label{tab:kmeans-full1}
\end{table}
\begin{table}[htb]
\caption{Test set accuracy for $k$-means and 10 different distance measures, part 2}
\begin{tabular}{lllllllllll}

Haptics                        & 0.3864 & 0.3571 & 0.3442 & 0.3117 & 0.3409 & 0.2987 & 0.3279 & 0.2240 & 0.2403 & 0.3539 \\
Herring                        & 0.5625 & 0.5156 & 0.5781 & 0.5781 & 0.5156 & 0.5000 & 0.5938 & 0.6094 & 0.5313 & 0.5938 \\
HouseTwenty                    & 0.7143 & 0.6387 & 0.6218 & 0.6387 & 0.6303 & 0.6303 & 0.5462 & 0.5294 & 0.5966 & 0.5714 \\
InlineSkate                    & 0.2418 & 0.2091 & 0.2145 & 0.2218 & 0.2109 & 0.2000 & 0.2327 & 0.2218 & 0.1891 & 0.1836 \\
InsectEPGRegularTrain          & 0.5984 & 0.6185 & 0.6064 & 0.5663 & 0.6104 & 0.5944 & 0.4779 & 0.5984 & 0.6345 & 0.4739 \\
InsectEPGSmallTrain            & 0.5823 & 0.5863 & 0.5743 & 0.5341 & 0.5743 & 0.6185 & 0.4940 & 0.6024 & 0.5462 & 0.6145 \\
InsectWingbeatSound            & 0.4727 & 0.2737 & 0.4975 & 0.1970 & 0.4657 & 0.1904 & 0.3540 & 0.1258 & 0.1429 & 0.1025 \\
ItalyPowerDemand               & 0.5141 & 0.5044 & 0.5121 & 0.5277 & 0.5209 & 0.5015 & 0.8319 & 0.5792 & 0.6463 & 0.8192 \\
LargeKitchenAppliances         & 0.4213 & 0.4160 & 0.3867 & 0.4187 & 0.3893 & 0.4160 & 0.3493 & 0.4587 & 0.4160 & 0.3333 \\
Lightning2                     & 0.5246 & 0.6230 & 0.5246 & 0.6393 & 0.5902 & 0.6393 & 0.5082 & 0.6721 & 0.6393 & 0.5410 \\
Lightning7                     & 0.5616 & 0.4658 & 0.5068 & 0.6027 & 0.4247 & 0.3973 & 0.4795 & 0.5068 & 0.4247 & 0.2603 \\
Mallat                         & 0.8350 & 0.8009 & 0.8657 & 0.8640 & 0.6955 & 0.8678 & 0.5697 & 0.5228 & 0.7552 & 0.4994 \\
Meat                           & 0.8667 & 0.7333 & 0.8667 & 0.8667 & 0.6333 & 0.7000 & 0.5333 & 0.6667 & 0.6667 & 0.4833 \\
MedicalImages                  & 0.3513 & 0.3158 & 0.3329 & 0.3250 & 0.2961 & 0.3158 & 0.2618 & 0.2750 & 0.2355 & 0.3171 \\
MiddlePhalanxOutlineAgeGroup   & 0.5130 & 0.5130 & 0.5000 & 0.5130 & 0.4805 & 0.5065 & 0.5390 & 0.5260 & 0.4675 & 0.5390 \\
MiddlePhalanxOutlineCorrect    & 0.6014 & 0.5842 & 0.5979 & 0.5979 & 0.6117 & 0.5842 & 0.5842 & 0.5842 & 0.5773 & 0.5842 \\
MiddlePhalanxTW                & 0.4416 & 0.4416 & 0.4156 & 0.4351 & 0.4286 & 0.4481 & 0.5519 & 0.4026 & 0.4675 & 0.5519 \\
MixedShapesRegularTrain        & 0.6495 & 0.6210 & 0.6536 & 0.5827 & 0.6495 & 0.5563 & 0.3880 & 0.3113 & 0.3608 & 0.2697 \\
MixedShapesSmallTrain          & 0.6647 & 0.5979 & 0.6165 & 0.6540 & 0.6392 & 0.4858 & 0.4915 & 0.3254 & 0.2841 & 0.2697 \\
MoteStrain                     & 0.8538 & 0.8107 & 0.8339 & 0.8842 & 0.8506 & 0.8147 & 0.5823 & 0.5863 & 0.7580 & 0.5256 \\
NonInvasiveFetalECGThorax1     & 0.5399 & 0.4972 & 0.4901 & 0.5038 & 0.4590 & 0.4382 & 0.2041 & 0.4656 & 0.4031 & 0.0575 \\
NonInvasiveFetalECGThorax2     & 0.6051 & 0.4865 & 0.5196 & 0.5746 & 0.5547 & 0.4763 & 0.2168 & 0.4972 & 0.4621 & 0.0855 \\
OliveOil                       & 0.8667 & 0.7667 & 0.8667 & 0.8667 & 0.8333 & 0.7667 & 0.3667 & 0.6000 & 0.7333 & 0.3667 \\
OSULeaf                        & 0.3926 & 0.3554 & 0.3760 & 0.2314 & 0.3512 & 0.2190 & 0.2273 & 0.2231 & 0.2273 & 0.2273 \\
PhalangesOutlinesCorrect       & 0.6259 & 0.6119 & 0.6270 & 0.6259 & 0.6282 & 0.6119 & 0.5991 & 0.6026 & 0.6014 & 0.5979 \\
Phoneme                        & 0.1624 & 0.1377 & 0.1250 & 0.1218 & 0.1050 & 0.1324 & 0.1930 & 0.1735 & 0.1566 & 0.1113 \\
PigAirwayPressure              & 0.1731 & 0.1587 & 0.1923 & 0.1683 & 0.1971 & 0.0817 & 0.2548 & 0.0240 & 0.0192 & 0.0385 \\
PigArtPressure                 & 0.0192 & 0.2212 & 0.3173 & 0.0192 & 0.3173 & 0.0192 & 0.2837 & 0.0192 & 0.0192 & 0.0529 \\
PigCVP                         & 0.1298 & 0.2452 & 0.1923 & 0.0865 & 0.1827 & 0.0288 & 0.2260 & 0.0240 & 0.0240 & 0.0385 \\
Plane                          & 0.8571 & 0.7619 & 0.9905 & 0.8571 & 0.8286 & 0.7810 & 0.8857 & 0.3619 & 0.2476 & 0.2000 \\
PowerCons                      & 0.8389 & 0.5889 & 0.6778 & 0.7333 & 0.6556 & 0.6111 & 0.5111 & 0.5722 & 0.6556 & 0.5778 \\
ProximalPhalanxOutlineAgeGroup & 0.7317 & 0.7756 & 0.7317 & 0.7415 & 0.7122 & 0.7756 & 0.7854 & 0.7707 & 0.7220 & 0.7854 \\
ProximalPhalanxOutlineCorrect  & 0.6460 & 0.6426 & 0.6460 & 0.6426 & 0.6460 & 0.6426 & 0.6357 & 0.6392 & 0.6392 & 0.6323 \\
ProximalPhalanxTW              & 0.5268 & 0.4878 & 0.5268 & 0.5317 & 0.5220 & 0.4878 & 0.5610 & 0.5220 & 0.4683 & 0.6439 \\
RefrigerationDevices           & 0.3813 & 0.3520 & 0.3520 & 0.3627 & 0.3520 & 0.3360 & 0.3733 & 0.3387 & 0.4320 & 0.3333 \\
Rock                           & 0.4200 & 0.5600 & 0.5000 & 0.4000 & 0.5400 & 0.4600 & 0.5200 & 0.5000 & 0.4800 & 0.5400 \\
ScreenType                     & 0.3920 & 0.4347 & 0.4373 & 0.3707 & 0.4133 & 0.3493 & 0.4587 & 0.3840 & 0.3760 & 0.4987 \\
SemgHandGenderCh2              & 0.6317 & 0.6067 & 0.6367 & 0.6367 & 0.6600 & 0.6783 & 0.6783 & 0.6633 & 0.6633 & 0.6700 \\
SemgHandMovementCh2            & 0.3578 & 0.2511 & 0.3600 & 0.3022 & 0.3444 & 0.3244 & 0.2844 & 0.1911 & 0.1822 & 0.2311 \\
SemgHandSubjectCh2             & 0.5400 & 0.4067 & 0.5511 & 0.4200 & 0.4578 & 0.3956 & 0.2822 & 0.2267 & 0.2156 & 0.3489 \\
ShapeletSim                    & 0.5667 & 0.5000 & 0.5000 & 0.5000 & 0.5278 & 0.5000 & 0.5000 & 0.5000 & 0.5000 & 0.5000 \\
ShapesAll                      & 0.3583 & 0.4017 & 0.3933 & 0.1233 & 0.4050 & 0.1000 & 0.1783 & 0.0283 & 0.0267 & 0.0167 \\
SmallKitchenAppliances         & 0.3413 & 0.4027 & 0.3413 & 0.4960 & 0.3600 & 0.4773 & 0.3600 & 0.4747 & 0.5707 & 0.5280 \\
SmoothSubspace                 & 0.5933 & 0.5533 & 0.5067 & 0.4867 & 0.6400 & 0.5600 & 0.5000 & 0.4133 & 0.3867 & 0.4600 \\
SonyAIBORobotSurface1          & 0.5092 & 0.8120 & 0.5158 & 0.8403 & 0.5191 & 0.8136 & 0.9085 & 0.5524 & 0.5324 & 0.9151 \\
SonyAIBORobotSurface2          & 0.7692 & 0.7566 & 0.7712 & 0.7387 & 0.7576 & 0.7555 & 0.7712 & 0.5635 & 0.5582 & 0.7377 \\
StarLightCurves                & 0.7661 & 0.7659 & 0.7634 & 0.7699 & 0.7589 & 0.8074 & 0.7067 & 0.7489 & 0.7890 & 0.5548 \\
Strawberry                     & 0.5081 & 0.5216 & 0.5081 & 0.5081 & 0.5108 & 0.5081 & 0.6405 & 0.5270 & 0.5324 & 0.6270 \\
SwedishLeaf                    & 0.5072 & 0.2448 & 0.4288 & 0.4752 & 0.4272 & 0.0944 & 0.1760 & 0.2480 & 0.1552 & 0.1536 \\
Symbols                        & 0.7879 & 0.7809 & 0.7367 & 0.7930 & 0.7387 & 0.6804 & 0.3970 & 0.5648 & 0.8010 & 0.5407 \\
SyntheticControl               & 0.6067 & 0.7633 & 0.4833 & 0.4900 & 0.5600 & 0.7500 & 0.1667 & 0.3367 & 0.3400 & 0.1667 \\
ToeSegmentation1               & 0.5219 & 0.5526 & 0.5175 & 0.5000 & 0.5132 & 0.5132 & 0.5175 & 0.5263 & 0.5263 & 0.5526 \\
ToeSegmentation2               & 0.5077 & 0.5308 & 0.5231 & 0.8154 & 0.5231 & 0.8154 & 0.5538 & 0.7923 & 0.8154 & 0.8385 \\
Trace                          & 0.5500 & 0.6200 & 0.5500 & 0.5600 & 0.5600 & 0.5700 & 0.4400 & 0.7300 & 0.5200 & 0.2900 \\
TwoLeadECG                     & 0.5812 & 0.5733 & 0.5505 & 0.6137 & 0.5338 & 0.5198 & 0.5066 & 0.7199 & 0.6839 & 0.5022 \\
TwoPatterns                    & 0.3593 & 0.3228 & 0.3108 & 0.2588 & 0.3158 & 0.2615 & 0.2588 & 0.2598 & 0.2588 & 0.2588 \\
UMD                            & 0.4792 & 0.5833 & 0.4583 & 0.5000 & 0.4583 & 0.5833 & 0.5347 & 0.4514 & 0.4444 & 0.3333 \\
UWaveGestureLibraryAll         & 0.6678 & 0.7473 & 0.6896 & 0.1284 & 0.6910 & 0.3216 & 0.3590 & 0.1393 & 0.1293 & 0.2702 \\
UWaveGestureLibraryX           & 0.5572 & 0.5623 & 0.5533 & 0.4458 & 0.5539 & 0.3939 & 0.2764 & 0.1770 & 0.2574 & 0.1457 \\
UWaveGestureLibraryY           & 0.4777 & 0.4704 & 0.5075 & 0.4162 & 0.4953 & 0.4595 & 0.2605 & 0.1803 & 0.2605 & 0.1284 \\
UWaveGestureLibraryZ           & 0.5173 & 0.5070 & 0.4659 & 0.5209 & 0.4623 & 0.4324 & 0.1963 & 0.1714 & 0.2007 & 0.2077 \\
Wafer                          & 0.6282 & 0.6274 & 0.6282 & 0.6287 & 0.6299 & 0.6269 & 0.6325 & 0.7148 & 0.6832 & 0.8921 \\
Wine                           & 0.5185 & 0.5185 & 0.5185 & 0.5185 & 0.5370 & 0.5185 & 0.5926 & 0.5370 & 0.5185 & 0.5926 \\
WordSynonyms                   & 0.3370 & 0.3448 & 0.3213 & 0.2288 & 0.2790 & 0.2727 & 0.2335 & 0.1928 & 0.1991 & 0.2100 \\
Worms                          & 0.3896 & 0.3377 & 0.3506 & 0.4026 & 0.4026 & 0.2987 & 0.3766 & 0.4286 & 0.4286 & 0.4805 \\
WormsTwoClass                  & 0.5325 & 0.5065 & 0.5195 & 0.5065 & 0.5325 & 0.5714 & 0.5714 & 0.5584 & 0.5714 & 0.5584 \\
Yoga                           & 0.5023 & 0.5107 & 0.5073 & 0.5270 & 0.5083 & 0.5360 & 0.5083 & 0.5357 & 0.5357 & 0.5357 \\ hline
Average                        & 0.5416	& 0.5225 & 0.5285 &	0.5089 & 0.5178 & 0.4908 & 0.4698 &	0.4520 & 0.4576	& 0.4257
\end{tabular}
\label{tab:kmeans-full2}
\end{table}

\begin{table}[htb]
\caption{Difference between the highest accuracy of the ten clusterers evaluated and predicting the majority class. Problems ranked from worst performing relative to majority to best.}
\begin{tabular}{llll|llll}
Problem                       & Max Acc & Majority & Difference & Problem                        & Max Acc & Majority & Difference  \\ \hline
MedicalImages                 & 0.3513  & 0.5145         & -0.1632    & CricketX                       & 0.3077  & 0.0667         & 0.2410     \\
ChlorineConcentration         & 0.4211  & 0.5326         & -0.1115    & UMD                            & 0.5833  & 0.3333         & 0.2500     \\
ProximalPhalanxOutline & 0.6460  & 0.6838         & -0.0378    & FreezerSmallTrain              & 0.7660  & 0.5000         & 0.2660     \\
Strawberry                    & 0.6405  & 0.6432         & -0.0027    & Beef                           & 0.4667  & 0.2000         & 0.2667     \\
WormsTwoClass                 & 0.5714  & 0.5714         & 0.0000     & FreezerRegularTrain            & 0.7667  & 0.5000         & 0.2667     \\
Wafer                         & 0.8921  & 0.8921         & 0.0000     & MiddlePhalanxTW                & 0.5519  & 0.2727         & 0.2792     \\
FordA                         & 0.5159  & 0.5159         & 0.0000     & ProximalPhalanxTW              & 0.6439  & 0.3512         & 0.2927     \\
Earthquakes                   & 0.7482  & 0.7482         & 0.0000     & ProximalPhalanxOutline         & 0.7854  & 0.4878         & 0.2976     \\
Yoga                          & 0.5360  & 0.5357         & 0.0003     & PigArtPressure                 & 0.3173  & 0.0192         & 0.2981     \\
PhalangesOutlines      & 0.6282  & 0.6131         & 0.0151     & EOGVerticalSignal              & 0.3895  & 0.0829         & 0.3066     \\
Herring                       & 0.6094  & 0.5938         & 0.0157     & SmoothSubspace                 & 0.6400  & 0.3333         & 0.3067     \\
ToeSegmentation2              & 0.8385  & 0.8154         & 0.0231     & ACSF1                          & 0.4100  & 0.1000         & 0.3100     \\
ToeSegmentation1              & 0.5526  & 0.5263         & 0.0263     & GunPointMaleVersusFemale       & 0.8354  & 0.5253         & 0.3101     \\
SemgHandGenderCh2             & 0.6783  & 0.6500         & 0.0283     & EOGHorizontalSignal            & 0.3978  & 0.0829         & 0.3149     \\
DistalPhalanxOutline   & 0.6196  & 0.5833         & 0.0363     & ECGFiveDays                    & 0.8165  & 0.4971         & 0.3194     \\
MiddlePhalanxOutline   & 0.6117  & 0.5704         & 0.0413     & DistalPhalanxTW                & 0.6259  & 0.3022         & 0.3237     \\
EthanolLevel                  & 0.2940  & 0.2520         & 0.0420     & GunPointAgeSpan                & 0.8386  & 0.5063         & 0.3323     \\
FordB                         & 0.5395  & 0.4951         & 0.0444     & ItalyPowerDemand               & 0.8319  & 0.4985         & 0.3334     \\
Worms                         & 0.4805  & 0.4286         & 0.0519     & Crop                           & 0.3804  & 0.0417         & 0.3387     \\
ShapeletSim                   & 0.5667  & 0.5000         & 0.0667     & PowerCons                      & 0.8389  & 0.5000         & 0.3389     \\
Phoneme                       & 0.1930  & 0.1129         & 0.0801     & Lightning7                     & 0.6027  & 0.2603         & 0.3424     \\
HandOutlines                  & 0.7243  & 0.6405         & 0.0838     & MoteStrain                     & 0.8842  & 0.5391         & 0.3451     \\
InlineSkate                   & 0.2418  & 0.1545         & 0.0873     & MiddlePhalanxOutline   & 0.5390  & 0.1883         & 0.3507     \\
Wine                          & 0.5926  & 0.5000         & 0.0926     & SemgHandSubjectCh2             & 0.5511  & 0.2000         & 0.3511     \\
RefrigerationDevices          & 0.4320  & 0.3333         & 0.0987     & FacesUCR                       & 0.5068  & 0.1434         & 0.3634     \\
BirdChicken                   & 0.6000  & 0.5000         & 0.1000     & UWaveGestureLibraryY           & 0.5075  & 0.1209         & 0.3866     \\
TwoPatterns                   & 0.3593  & 0.2588         & 0.1006     & CBF                            & 0.7189  & 0.3311         & 0.3878     \\
Ham                           & 0.6190  & 0.5143         & 0.1047     & ShapesAll                      & 0.4050  & 0.0167         & 0.3883     \\
ECG200                        & 0.7600  & 0.6400         & 0.1200     & FiftyWords                     & 0.5165  & 0.1253         & 0.3912     \\
WordSynonyms                  & 0.3448  & 0.2194         & 0.1254     & UWaveGestureLibraryZ           & 0.5209  & 0.1209         & 0.4000     \\
LargeKitchenAppliances        & 0.4587  & 0.3333         & 0.1254     & Car                            & 0.6167  & 0.2167         & 0.4000     \\
Lightning2                    & 0.6721  & 0.5410         & 0.1311     & Adiac                          & 0.4246  & 0.0205         & 0.4041     \\
HouseTwenty                   & 0.7143  & 0.5798         & 0.1345     & InsectWingbeatSound            & 0.4975  & 0.0909         & 0.4066     \\
Rock                          & 0.5600  & 0.4200         & 0.1400     & CinCECGTorso                   & 0.6601  & 0.2478         & 0.4123     \\
Computers                     & 0.6440  & 0.5000         & 0.1440     & Coffee                         & 0.9643  & 0.5357         & 0.4286     \\
InsectEPGSmallTrain           & 0.6185  & 0.4739         & 0.1446     & UWaveGestureLibraryX           & 0.5623  & 0.1209         & 0.4414     \\
SonyAIBORobotSurface2         & 0.7712  & 0.6170         & 0.1542     & SwedishLeaf                    & 0.5072  & 0.0528         & 0.4544     \\
InsectEPGRegularTrain         & 0.6345  & 0.4739         & 0.1606     & MixedShapesRegularTrain        & 0.6536  & 0.1889         & 0.4647     \\
ScreenType                    & 0.4987  & 0.3333         & 0.1654     & OliveOil                       & 0.8667  & 0.4000         & 0.4667     \\
ECG5000                       & 0.7496  & 0.5838         & 0.1658     & MixedShapesSmallTrain          & 0.6647  & 0.1889         & 0.4758     \\
GunPoint                      & 0.6667  & 0.4933         & 0.1734     & SonyAIBORobotSurface1          & 0.9151  & 0.4293         & 0.4858     \\
Haptics                       & 0.3864  & 0.2078         & 0.1786     & NonInvasiveFetalECGThorax1     & 0.5399  & 0.0183         & 0.5216     \\
SemgHandMovementCh2           & 0.3600  & 0.1667         & 0.1933     & Meat                           & 0.8667  & 0.3333         & 0.5334     \\
BeetleFly                     & 0.7000  & 0.5000         & 0.2000     & Fish                           & 0.6629  & 0.1257         & 0.5372     \\
ArrowHead                     & 0.6000  & 0.3943         & 0.2057     & FaceAll                        & 0.5805  & 0.0426         & 0.5379     \\
OSULeaf                       & 0.3926  & 0.1818         & 0.2108     & Trace                          & 0.7300  & 0.1900         & 0.5400     \\
GunPointOldVersusYoung        & 0.7429  & 0.5238         & 0.2191     & BME                            & 0.8867  & 0.3333         & 0.5534     \\
TwoLeadECG                    & 0.7199  & 0.4996         & 0.2203     & FaceFour                       & 0.7386  & 0.1591         & 0.5795     \\
ElectricDevices               & 0.4645  & 0.2424         & 0.2221     & NonInvasiveFetalECGThorax2     & 0.6051  & 0.0183         & 0.5868     \\
PigCVP                        & 0.2452  & 0.0192         & 0.2260     & SyntheticControl               & 0.7633  & 0.1667         & 0.5966     \\
StarLightCurves               & 0.8074  & 0.5772         & 0.2302     & Chinatown                      & 0.8892  & 0.2741         & 0.6151     \\
CricketZ                      & 0.2949  & 0.0615         & 0.2334     & UWaveGestureLibraryAll         & 0.7473  & 0.1209         & 0.6264     \\
PigAirwayPressure             & 0.2548  & 0.0192         & 0.2356     & Symbols                        & 0.8010  & 0.1739         & 0.6271     \\
SmallKitchenAppliances        & 0.5707  & 0.3333         & 0.2374     & DiatomSizeReduction            & 0.9575  & 0.3007         & 0.6568     \\
DistalPhalanxOutlineAge & 0.7050  & 0.4676         & 0.2374     & Mallat                         & 0.8678  & 0.1232         & 0.7446     \\
CricketY                      & 0.3128  & 0.0744         & 0.2384     & Plane                          & 0.9905  & 0.0952         & 0.8953
\end{tabular}
\label{tab:diffs}
\end{table}



\end{document}